\documentclass[screen,acmsmall,authorversion]{acmart}

\AtBeginDocument{%
  }

\setcopyright{acmlicensed}
\copyrightyear{2025}
\acmYear{2025}
\acmDOI{10.1145/3770077}

\acmJournal{TIST}
\acmVolume{0}
\acmNumber{0}
\acmArticle{0}
\acmMonth{0}

\usepackage{longtable}
\usepackage{enumitem}
\usepackage{tablefootnote}
\usepackage{multirow}
\usepackage{footmisc}

\usepackage{tikz}
\usepackage[edges]{forest}
\usetikzlibrary{trees, positioning, shapes, shadows, arrows.meta}

\definecolor{light-blue}{RGB}{217, 232, 252}
\definecolor{light-pink}{RGB}{255, 232, 252}
\definecolor{draw}{RGB}{106,142,189}

\tikzstyle{my-box}=[
    rectangle,
    draw=draw,
    rounded corners,
    text opacity=1,
    minimum height=1.5em,
    minimum width=5em,
    inner sep=2pt,
    align=center,
    fill opacity=.5,
]

\tikzstyle{leaf-single}=[my-box, minimum height=1.5em,
    fill=light-blue!60, text=black, align=left,font=\scriptsize,
    inner xsep=2pt,
    inner ysep=4pt,
]

\tikzstyle{leaf-multi}=[my-box, minimum height=1.5em,
    fill=light-pink!60, text=black, align=left,font=\scriptsize,
    inner xsep=2pt,
    inner ysep=4pt,
]

\begin{document}

\newcommand{\identificationACM}{417}
\newcommand{\identificationACL}{12,132}
\newcommand{\identificationIEEE}{311}
\newcommand{\identificationScienceDirect}{383}
\newcommand{\identificationTotal}{13,243}
\newcommand{\identificationExtendedSearch}{52}
\newcommand{\screeningRecordsScreened}{13,243}
\newcommand{\screeningRecordsExcluded}{13,010}
\newcommand{\screeningRecordsSought}{233}
\newcommand{\screeningRecordsNotRetrieved}{9}
\newcommand{\screeningRecordsAssessed}{224}
\newcommand{\screeningRecordsExcludedNonNLP}{38}
\newcommand{\screeningRecordsExcludedNonRelevant}{52}
\newcommand{\screeningOtherReportsAssessed}{52}
\newcommand{\screeningOtherReportsExcludedNonNLP}{6}
\newcommand{\screeningOtherReportsExcludedNonRelevant}{5}
\newcommand{\includedTotal}{175}

\title[Survey on Automatic Credibility Assessment Using Textual Credibility Signals]{A Survey on Automatic Credibility Assessment Using Textual Credibility Signals in the Era of Large Language Models}


\author{Ivan Srba}
\email{ivan.srba@kinit.sk}
\orcid{0000-0003-3511-5337}
\affiliation{
  \institution{Kempelen Institute of Intelligent Technologies}
  \city{Bratislava}
  \country{Slovakia}
}

\author{Olesya Razuvayevskaya}
\email{o.razuvayevskaya@sheffield.ac.uk}
\orcid{0000-0002-7922-7982}
\affiliation{
  \institution{The University of Sheffield}
  \city{Sheffield}
  \country{UK}
}

\author{João A. Leite}
\email{jaleite1@sheffield.ac.uk}
\orcid{0000-0002-3587-853X}
\affiliation{
  \institution{The University of Sheffield}
  \city{Sheffield}
  \country{UK}
}

\author{Robert Moro}
\email{robert.moro@kinit.sk}
\orcid{0000-0002-3052-8290}
\affiliation{
  \institution{Kempelen Institute of Intelligent Technologies}
  \city{Bratislava}
  \country{Slovakia}
}

\author{Ipek Baris Schlicht}
\email{ipek.baris-schlicht@dw.com}
\orcid{0000-0002-5037-2203}
\affiliation{
  \institution{Deutsche Welle}
  \city{Bonn/Berlin}
  \country{Germany}
}
\affiliation{
  \institution{Universitat Politècnica de València}
  \city{Valencia}
  \country{Spain}
}

\author{Sara Tonelli}
\email{satonelli@fbk.eu}
\orcid{0000-0001-8010-6689}
\affiliation{
  \institution{Fondazione Bruno Kessler}
  \city{Trento}
  \country{Italy}
}

\author{Francisco Moreno García}
\email{fran.moreno@upm.es}
\orcid{0000-0001-7883-4812}
\affiliation{
  \institution{Universidad Politécnica de Madrid}
  \city{Madrid}
  \country{Spain}
}

\author{Santiago Barrio Lottmann}
\email{s.barrio@upm.es}
\orcid{0000-0003-1881-7871}
\affiliation{
  \institution{Universidad Politécnica de Madrid}
  \city{Madrid}
  \country{Spain}
}

\author{Denis Teyssou}
\email{Denis.TEYSSOU@afp.com}
\orcid{0000-0003-0505-6356}
\affiliation{
  \institution{Agence France-Presse}
  \city{Paris}
  \country{France}
}

\author{Valentin Porcellini}
\email{Valentin.PORCELLINI@afp.com}
\orcid{0000-0001-7909-8853}
\affiliation{
  \institution{Agence France-Presse}
  \city{Paris}
  \country{France}
}

\author{Carolina Scarton}
\email{c.scarton@sheffield.ac.uk}
\orcid{0000-0002-0103-4072}
\affiliation{
  \institution{The University of Sheffield}
  \city{Sheffield}
  \country{UK}
}

\author{Kalina Bontcheva}
\email{k.bontcheva@sheffield.ac.uk}
\orcid{0000-0001-6152-9600}
\affiliation{
  \institution{The University of Sheffield}
  \city{Sheffield}
  \country{UK}
}

\author{Maria Bielikova}
\email{maria.bielikova@kinit.sk}
\orcid{0000-0003-4105-3494}
\affiliation{
  \institution{Kempelen Institute of Intelligent Technologies}
  \city{Bratislava}
  \country{Slovakia}
}

\renewcommand{\shortauthors}{Srba et al.}

\begin{abstract}
  In the age of social media and generative AI, the ability to automatically assess the credibility of online content has become increasingly critical, complementing traditional approaches to false information detection. Credibility assessment relies on aggregating diverse credibility signals -- small units of information, such as content subjectivity, bias, or a presence of persuasion techniques -- into a final credibility label/score. However, current research in automatic credibility assessment and credibility signals detection remains highly fragmented, with many signals studied in isolation and lacking integration. Notably, there is a scarcity of approaches that detect and aggregate multiple credibility signals simultaneously. These challenges are further exacerbated by the absence of a comprehensive and up-to-date overview of research works that connects these research efforts under a common framework and identifies shared trends, challenges, and open problems. In this survey, we address this gap by presenting a systematic and comprehensive literature review of {\includedTotal} research papers, focusing on textual credibility signals within the field of Natural Language Processing (NLP), which undergoes a rapid transformation due to advancements in Large Language Models (LLMs). While positioning the NLP research into the the broader multidisciplinary landscape, we examine both automatic credibility assessment methods as well as the detection of nine categories of credibility signals.  We provide an in-depth analysis of three key categories: 1) factuality, subjectivity and bias, 2) persuasion techniques and logical fallacies, and 3) check-worthy and fact-checked claims. In addition to summarising existing methods, datasets, and tools, we outline future research direction and emerging opportunities, with particular attention to evolving challenges posed by generative AI.
\end{abstract}





\maketitle

\section{Introduction}

Tackling false information (i.e., disinformation and misinformation) has attracted a substantial attention in recent years from researchers, media professionals, AI and social media practitioners as well as the general public. Researchers have approached false information detection through a variety of classification tasks, leveraging AI technologies such as machine learning, natural language processing, computer vision, or social network analysis~\cite{Sharma2019}. The primary stream of research works on \textit{false information detection}, commonly also denoted as a \textit{fake news detection}~\cite{Shu2017,Zhou2020,huOverviewFakeNews2024}, assesses the \textit{veracity} (i.e., factual accuracy) of information by means of a single-label prediction, typically a binary one (i.e., false/true content)~\cite{Shu2017}. Some works go beyond a binary classification and predict multiple classes, such as veracity levels, commonly used by fact-checking organizations (e.g., true, mostly true, mixture, mostly false, false).

However, veracity alone offers only a limited perspective on the potential harm of online content. In practice, it should be considered alongside credibility. By proceeding from existing (quite heterogeneous) definitions~\cite{credibility_measures,9543698,8572695,riehCredibilityMultidisciplinaryFramework2007,selfCredibility2019}, under the \textit{credibility} we understand the perceived trustworthiness, accuracy, and reliability of information or its source. It reflects the extent to which content is believed to be factual, unbiased, and free from manipulation or deception. False content that is perceived as highly credible by its audience can lead to significantly greater harm than false content that is clearly perceived as lacking credibility.

\textit{Credibility assessment} therefore provides an interesting potential to extend and complement the existing stream of works on false information detection. It typically follows a two-step process. First, granular \textit{credibility signals} (also referred to as \textit{credibility indicators}) are detected. Second, these signals (serving as evidence) are aggregated into a single ordinal \textit{credibility label} or a numerical \textit{credibility score}. Common examples of credibility signals include subjectivity and bias present in the content, persuasion techniques, logical fallacies or presence of machine-generated content, which has become increasingly relevant with the rise of generative AI. While credibility assessment can be conducted manually by the end users, its full potential lies in automated detection of credibility signals and their consequent aggregation by a \textit{credibility assessment algorithm}. These automated approaches may range from simple (e.g., rule-based or heuristic) methods to more advanced systems powered by AI, capable of nuanced and context-aware content analysis. Such automation not only improves scalability but also enables real-time evaluation of large volumes of online content.

The benefit of credibility signals is their flexibility and wide-applicability. Besides their primary usage, to be aggregated into a credibility label/score, they can also be utilized and interpreted by a user directly (e.g., as an ``information nutrition label'' \cite{10.1145/3190580.3190588}), provide valuable inputs to information retrieval engines or recommender systems in order to prefer content/sources associated with a high level of credibility \cite{weerkamp2008credibility}, or even serve as features in subsequent classification tasks (including false information detection itself \cite{Sitaula2020}). Credibility assessment and credibility signals are, furthermore, especially useful in cases in which we cannot easily ascertain that something is true or false (on a single veracity dimension) -- concept of credibility provides a necessary level of granularity to represent a potentially complex information from multiple perspectives. 

Credibility signals are also more easily interpretable and naturally support semi-automatic human-centered AI approaches that attempt supporting instead of replacing a human expert or a lay person (individual credibility signals can be automatically detected and their interpretation can be subsequently done by a human). Study by \citet{lu2022effects} showed that even if people are influenced by others when judging the veracity of online content, providing accurate AI-based credibility indicators can effectively improve people’s ability in detecting false information. Last but not least, they are also aligned with journalistic/fact-checking workflows for identifying trustworthy information (to be cited) or potential false information (to be fact-checked). 

In contrast to other related research areas (including false information detection), research works on automatic credibility assessment are mostly carried out in \textit{isolation}. As we show in this survey, there is a lack of works addressing multiple categories of credibility signals at once and, moreover, many research works do not explicitly state that the outcome of their solution can actually serve as a credibility signal. Getting familiar with the existing works on automatic credibility assessment and detection of credibility signals is challenging due to an \textit{ambiguous and inconsistent terminology} used by researchers, a lack of \textit{clear definition of credibility and credibility signals} as well as due to \textit{a missing standardized taxonomy} of various credibility signals' categories. At the same time, automatic credibility assessment can significantly benefit from a \textit{holistic approach}. For example, due to common underlying similarities, there is a big potential of transfer or multi-task learning.

While there is a wide range of possible credibility signals, we specifically focus on such signals that relate to textual content and can be automatically detected by \textit{Natural Language Processing (NLP)} methods. This area is currently at the spotlight due to the recent advancements of \textit{Large Language Models (LLMs)}. Their rapid development has a revolutionizing effect on many text classification tasks, automatic credibility assessment and detection of credibility signals not being an exception. 

Despite multiple surveys addressing false information detection (e.g., \cite{huOverviewFakeNews2024,tufchiComprehensiveSurveyMultimodal2023,a.b.SystematicSurveyExplainable2023,Zhou2020,Shu2017}), to the best of our knowledge, there is \textit{no survey on automatic credibility assessment and detection of credibility signals from the NLP perspective}. In this paper, we therefore provide the first comprehensive and systematic literature overview of credibility signal detection with the focus on NLP and LLMs. The main contributions are as follows:
\begin{enumerate}
    \item We address the fragmentation of existing works on automatic credibility assessment and credibility signals detection. We connect these research efforts under a common framework and identify shared trends, challenges, and open problems. In this way, we motivate future research in connecting and integrating individual outcomes (i.e., to move from currently isolated individual signals to credibility assessment effectively combining them).
    \item We provide a necessary background overview of definitions and dimensions of credibility assessment (including a unified categorization of credibility signals based on their various taxonomies and list used in the existing literature) to setup the common understanding that is currently missing in the existing works.
    \item We analyse and describe a total of {\includedTotal} NLP works tackling the automatic credibility assessment and detection of credibility signals. We provide an in-depth description of 3 key categories of credibility signals, which have been selected due to a considerable NLP research interest and their recognition as important by end users, namely: (i) factuality, subjectivity and bias; (ii) persuasion techniques and logical fallacies; and (iii) check-worthy and fact-checked claims. This analysis is complemented with an overview of additional 6 categories of credibility signals. We also thoroughly position such NLP research in the context of other modalities as well as multidisciplinary works.
    \item We identify and critically analyse key gaps in the existing literature, highlighting underexplored areas and methodological limitations in the current approaches to credibility assessment. Furthermore, we outline future research challenges and opportunities, with particular emphasis on the rapidly evolving landscape shaped by generative AI, especially Large Language Models (LLMs). We discuss how these advancements both pose new threats, such as the large-scale generation of highly credible yet false content, and offer new capabilities for enhancing the detection and aggregation of credibility signals. Our analysis aims to guide future research toward more comprehensive, robust and explainable credibility assessment approaches.
\end{enumerate}

Besides the review of published research works, this survey builds on extensive past research activities and acquired knowledge of its authors in the target area. A unique composition of authors consisting of experts not only on the primary NLP area, but also on other related research domains (such as computer vision, social network analysis) as well as media professionals, provides a novel holistic perspective on the addressed topic. Such diversity contributes, besides others, to identification of open problems and future challenges, in which an application of research in practice plays a prominent role. 

This survey is structured into 10 sections. Section 2 provides definitions of core concepts related to credibility, describes various dimensions of credibility assessment as well as defines the scope of this survey while situating it into the context of other existing surveys. In Section 3, the methodology employed in the survey process is thoroughly described. Section 4 focuses on research papers that address automatic credibility assessment (i.e., approaches addressing both steps --- detection of credibility signals and their aggregation into a credibility label/score). Subsequently, Sections 5-7 address an automatic detection of 3 selected key categories of credibility signals. Section 8 supplements this analysis with a brief overview of additional 6 categories of credibility signals, including the complementary perspective of credibility signals addressed in the non-NLP research. In Section 9, we provide an orthogonal analysis of challenges and open problems that characterize the state of the art in this area. Finally, conclusions are drawn in Section 10.
\section{Background}
\label{section:background}

\subsection{Definitions}
\label{section:background-definitions}

The construct of \textit{credibility} has received more scholarly attention than most other communication variables, with foundational research dating back to 1951, when \citet{10.1086/266350} demonstrated that the effectiveness of communication is heavily influenced by the perceived credibility of its source. A systematic meta-analysis of leading international communication journals spanning 1951 to 2011~\cite{credibility_measures} examined how source, message, and media credibility have been conceptualized and measured over time. The findings revealed significant inconsistencies across credibility scales, a lack of operational precision, and limited replication and validation of the credibility construct. These discrepancies largely stem from the fact that credibility is a complex, multidimensional concept that lacks a single, unified, and widely accepted definition.

Existing definitions vary especially across various disciplines and fields, which study credibility from different perspectives, such as information science, psychology, marketing or human-computer interaction (HCI)~\cite{riehCredibilityMultidisciplinaryFramework2007}. It is commonly defined as a high-level construct with a help of related concepts, such as believability (which is considered to be roughly a synonym with credibility), trust/trusthworthiness, reliability or their various combinations~\cite{9543698,8572695,riehCredibilityMultidisciplinaryFramework2007,selfCredibility2019}. Credibility is always tied to a \textit{target entity} (an object of assessment), that can be either a piece of content or a source spreading such content \cite{kiousisPublicTrustMistrust2001}. Due to ambiguity of terminology, in this survey we opt for community-contributed definitions \cite{hawkeTechnologicalApproachesImproving} gathered by the \textit{Credible Web Community Group}\footnote{\url{https://credweb.org/}}. It defines credibility as a degree to which information is credible (believable) and to which information appears non-misleading and useful (for the given audience).

\textit{Credibility assessment} is a process of ascertaining some level/degree of credibility to a target entity. Practically, credibility assessment can be formulated as~\cite{9543698}: (i) an ordinal classification problem (i.e., assigning a label from pre-defined categories, such as a low/high level of credibility); or as (ii) a regression/scoring problem (i.e., assigning a numerical score). In both cases, there are no standard scales adopted yet. Credible Web Community Group distinguishes two kinds of credibility assessment acts~\cite{hawkeTechnologicalApproachesImproving}:
\begin{enumerate}
    \item \textit{Making credibility decisions} is an act of \textit{adhoc} deciding for oneself what to believe, which is often informal, immediate, and unconscious. Doing this incorrectly often leads to being misled, although it may not be practical to do it correctly at all times.
    \item \textit{Credibility analysis} is an act of \textit{systematic} gathering, organizing, and analysing evidence to help people make credibility decisions about a particular information item. A credibility analysis process (performed manually by people or automatically by machines) might produce some kind of report which might itself be called a \textit{credibility assessment} and might include a \textit{credibility score}. In the following text, we will focus especially on this act.
\end{enumerate}

Consequently, a \textit{credibility signal}~\cite{hawkeTechnologicalApproachesImproving} is a small unit of information used in making a credibility assessment as an evidence. This can be a measurable feature of the information being assessed for credibility, or information about it (metadata), or information about entities which relate to it in various ways, such as the entity who provided it.

\textit{Credibility indicator}~\cite{hawkeTechnologicalApproachesImproving} is commonly used interchangeably with  credibility signal. In some communities, \textit{credibility signal} is used for inputs to \textit{credibility assessment algorithms} and \textit{credibility indicator} is used for the display features added to the output, to communicate (explain) results to an end user. In this survey, we prefer to use \textit{credibility signal}, while we would like to emphasize that in practice many signals can be used directly also as indicators communicated to end users.

Finally, \textit{credibility assessment tools}~\cite{hawkeTechnologicalApproachesImproving} are software features or applications which perform credibility assessment or help people do so. Such tools can either derive credibility signals, or implement also a \textit{credibility assessment algorithm} to aggregate such signals into a credibility label/score.

Credibility and credibility assessment is inherently very close to \textit{false information} and the task of \textit{false information detection}. Nevertheless, credibility (as also defined above) is rather a parallel/complementary concept. While they tend to correlate (low credibility may indicate false content), non-credible content does not necessarily need to be false only and vice versa. 

We would like to emphasize that credibility signals have been already recognized in false information research as a possible solution towards more accurate and explainable methods. The work by \citet{Grieve_Woodfield_2023} delve into the complexities of identifying and analysing disinformation deceptive news. The authors study the way language adapts to different contexts and purposes, arguing that the linguistic choices in deceptive news differ systematically from those in genuine news. The difference in intent -- to deceive versus to inform -- leads to detectable variations in language use. The authors illustrate this with the famous case of fake news involving Jayson Blair of The New York Times, where false news were less informationally dense and less confident than real ones. The pressure to invent news and the lack of factual grounding influenced writing style, making it less precise and authoritative. The authors propose that this type of register variation could be a broader indicator of fake news. They call for continued research to understand the linguistic patterns of fake news and to develop strategies to combat its spread. Although the above framework is only based on the analysis of one reporter's articles, this approach may be extended to corpora of non-credible/fake news and credible/real news on various topics to try to identify credibility signals/register variation  between the two.

To complement this recent case, we can also recall that the idea of analysing the language of disinformation and propaganda has its source in the pioneering work of the German linguist and philologist Viktor Klemperer \cite{klemperer2006language}. Klemperer showed how hate speech and conspiracy theories were used by the Nazis to justify the persecution of Jews. He also showed how these discourses were repeated and amplified by the media and institutions, until they became the norm.

\subsection{Dimensions of credibility assessment}
\label{section:background-dimensions}

In the past research works, a wide range of approaches studying the concept of credibility and credibility assessment was developed. In the following overview (see also Figure~\ref{fig:credibility_dimensions}), we provide various dimensions adopted to study these concepts.

\begin{figure}
    \centering
    \resizebox{\textwidth}{!}{

    \begin{forest}
        forked edges,
        for tree={
            grow=east,
            reversed=true,
            anchor=base west,
            parent anchor=east,
            child anchor=west,
            base=left,
            font=\small,
            rectangle,
            draw=draw,
            rounded corners,
            align=left,
            minimum width=5em,
            edge+={darkgray, line width=1pt},
            s sep=3pt,
            inner xsep=5pt,
            inner ysep=4pt,
            ver/.style={rotate=90, child anchor=north, parent anchor=south, anchor=center},
        },
        where level=1{text width=11em,font=\scriptsize}{},
        where level=2{text width=6em,font=\scriptsize}{},
        where level=3{text width=6.8em,font=\scriptsize}{},
        [Dimensions of credibility assessment, ver
            [Approaches to credibility assessment
                [\textbf{Automation-based approaches}, 
                 leaf-single, text width=10em]
                [Human-based approaches, 
                 leaf-single, text width=10em]
                [Hybrid approaches, 
                 leaf-single, text width=10em]
            ]
            [Levels of credibility assessment
                [\textbf{Content level}, 
                 leaf-single, text width=10em]
                [User/source level, 
                 leaf-single, text width=10em]
                [Topic/event level, 
                 leaf-single, text width=10em]
                [Media level, 
                 leaf-single, text width=10em]
                [Hybrid level, 
                 leaf-single, text width=10em]
            ]
            [Taxonomies of credibility signals
                [Information Source
                    [\textbf{Content-based signals}, 
                     leaf-single, text width=20em]
                    [\textbf{Context-based signals}, 
                     leaf-single, text width=20em]
                ]
                [Polarity
                    [\textbf{Predictors of high credibility}, 
                     leaf-single, text width=20em]
                    [\textbf{Predictors of low credibility}, 
                     leaf-single, text width=20em]
                ]
                [Subject Type
                    [\textbf{Text-based subjects: Claim, Text (in general), Article, Title}, 
                     leaf-single, text width=20em]
                    [{Multimedia-based subjects: Image, Audio, Video}, 
                     leaf-single, text width=20em]
                    [{Metadata subjects: Web page, Website, Aggregation, Venue, \\ Provider, Creator, Person, Organization}, 
                     leaf-single, text width=20em]
                ]
            ]
        ]
    \end{forest}
    }

    \caption{Dimensions of credibility assessment. Bold font highlights such aspects which are within the scope of this survey (see Section~\ref{section:background-scope} for more information). Please note that \textit{Content level} under the dimension of \textit{Levels of credibility assessment} and \textit{Content-based signals} under the dimension of \textit{Taxonomies of credibility signals} refer to two different concepts -- the former to the level of analysis, and the latter to the nature of the source data from which the credibility signals are detected.}
    \label{fig:credibility_dimensions}
\end{figure}

\textbf{Approaches to credibility assessment}. First, the credibility assessment methods can be broadly divided into three major groups~\cite{8572695}: 
\begin{enumerate}
    \item A large group of \textit{automation-based approaches} focuses either on automatically detecting the credibility signals or aggregating them into the credibility indicators. The complexity of such automation can be very diverse, ranging from simple rule- and heuristic-based methods, through weighted and information retrieval (IR) approaches to supervised/unsupervised machine learning. This group also covers research works focusing on building datasets appropriate to create such methods/models; as well as systems (web applications, browser extensions, etc.) built on the top of such automated methods.
    \item The second group corresponds to \textit{human-based approaches}, in which various aspects of credibility are studied from the perspective of end-users. The human-based credibility assessment approaches can range from voting methods, cognitive and perception approaches to manual verification approaches. 
    \item Finally, \textit{hybrid approaches} combine and utilize the advantages of both the automation- and human-based approaches. 
\end{enumerate}

\textbf{Levels of credibility assessment}. Following a type of a target entity, which is a subject of credibility assessment, we can distinguish five levels of credibility assessment as follows~\cite{8572695,9543698}:

\begin{enumerate}
    \item \textit{Content level}. At the content level, the task is to analyse the content attributes typically of a social media post or a news article. It is the most fundamental and important type of assessment, which is commonly used to determine the credibility at other levels (e.g., if a post is assessed to be credible, also a corresponding user/source or an associated topic/event is considered to be credible). 
    \item \textit{User/source level}. This level of credibility assessment depends on features extracted from user accounts (e.g., age, education, profile image) and a history of user-generated content. Besides a specific user, it can relate to a broader source (e.g., a news portal, an organization).    
    \item \textit{Topic/event level}. At this level, a credibility is assessed by proceeding from a cluster of posts falling under a specific trending or potentially high-impact topic or event (e.g., elections, societal crisis situations).
    \item \textit{Media level}. At the media level, the medium used to communicate and spread information is a target of credibility analysis (e.g., an online social network). This level typically encompasses credibility analysis of authors, spreaders as well as posts themselves.   
    \item \textit{Hybrid level}. To optimally utilize the advantages of individual levels and take advantage of high correlation between them, researchers commonly apply a hybrid approach in which the credibility is assessed at multiple levels at the same time.
\end{enumerate}

\textbf{Taxonomies of credibility signals}. Credibility signals, as a core unit in the credibility assessment process, can be categorized by various taxonomies, following different views.

First, reflecting the \textit{source of information}, many works broadly distinguish between content- and context-based credibility signals \cite{10.1145/3184558.3188731, 10.1145/2166966.2166998, 10.1145/3415164, 10.1016/j.ins.2019.12.040, el-ballouli-etal-2017-cat, esteves2018belittling, QURESHI2023109028, piccolo2021agents, 10.1145/3613904.3642473, CHANG2021106629, JI2023103210, GAO201521, CHOI2023103321}. Content-based signals are derived from the content itself, such as partisanship, emotional appeal, or persuasion techniques. On the other hand, \textit{context-based} signals, also referred to as provenance- \cite{feng2023examining} or meta-information, take into account contextual (e.g., user, source) cues, such as user statistics, location, domain reputation, or the number of shares. In practice, when a content-level credibility (e.g., a credibility of a social media post) is about to be assessed, a combination of both -- content-based credibility signals (e.g., a presence of persuasion techniques) as well as context-based credibility signals (e.g., the same claim was already fact-checked) -- may be utilized.

Second, the credibility signals differ in their polarity. In the existing works, some signals are formulated as \textit{predictors of high credibility} (e.g., high objectivity), while other signals are formulated as \textit{predictors of low credibility} (e.g., presence of logical flaws). When interpreting or aggregating individual credibility signals (either by an end user or a credibility assessment algorithm), such polarity must be explicitly taken into consideration (e.g., in \cite{podgurskiEvaluatingWebContent2022,leiteDetectingMisinformationLLMPredicted2023}).

Third, Credible Web Community Group~\cite{multipleauthorsCredibilitySignalsUnofficial}  structured credibility signals by a \textit{subject type} (a part of target entity they are derived from): claim, text, image, audio, video, article, title, web page, website, aggregation (e.g., Really Simple Syndication - RSS), venue, provider, creator, person, or organization.

Despite many commendable efforts, there is no standardized unified list and categorization of credibility signals. As a result, research works commonly utilize various and very diverse lists of credibility signals. For example, \citet{molinaFakeNewsNot2021} proposed separate lists of credibility signals (or features as they have been denoted by the authors) to distinguish between eight types of online content, such as real news, fabricated news, or satire. Each list grouped signals into 4 main categories roughly corresponding to levels of credibility assessment: (i) message and linguistic, (ii) sources and intentions, (ii) structural, and (iv) network.

An alternative approach was used by the \textit{Credibility Coalition}\footnote{\url{https://credibilitycoalition.org/}}, which organized weekly remote sessions during which participants drafted about 100 credibility signals specifically aimed at the credibility of web pages~\cite{10.1145/3184558.3188731}. They have been later coalesced into 12 major categories, including reader behaviour, revenue models, publication metadata, and inbound and outbound references. In 2017, the \textit{Credibility Coalition} formed the \textit{World Wide Web Consortium (W3C) Credible Web Community Group}, which continued in this initiative. It crowdsourced from human experts (researchers as well practitioners) an extensive list containing more than 200 credibility signals, commonly referred to as \textit{W3C signals}~\cite{multipleauthorsCredibilitySignalsUnofficial}. This list introduces a number of signals categorized by subject types. While such a list still remains an informal and incomplete draft and has not been standardized yet, it has been adopted and served as a foundation to several research works (e.g.,~\cite{piccolo2021agents,podgurskiEvaluatingWebContent2022}).

We can conclude that no standard or widely used and recognized list or categorization of credibility signals currently exists that could serve as a solid basis for our literature survey. Categorization by means of \textit{subject types} as proposed by W3C signals~\cite{multipleauthorsCredibilitySignalsUnofficial} is probably the most comprehensive and elaborated one, nevertheless, it mixes at the top level different modalities (text, video, audio, etc.) and content types (article, title, claim etc.). More specifically, the signals relevant to our survey are not optimally classified under the \textit{text} modality or across various individual content types. Therefore, we thoroughly analysed the relevant existing works~\cite{molinaFakeNewsNot2021,10.1145/3184558.3188731,multipleauthorsCredibilitySignalsUnofficial,podgurskiEvaluatingWebContent2022,nakov2021survey} and proposed a novel unified categorization of textual credibility signals (see Figure \ref{fig:credibility_signals_taxonomy}). We adopted a bottom-up approach as follows:
\begin{enumerate}
    \item First, out of mentioned credibility signals, we identified specifically \textit{textual credibility signals} (i.e., signals that can be derived from textual content by means of diverse NLP methods).
    \item We grouped the resulting credibility signals according to their semantic similarity and shared datasets and methods applied for their detection. To this end, we employed three grouping strategies. At first, we considered the existing categorizations that have been proposed by the relevant works, especially individual (sub)categories from the above-mentioned W3C signals~\cite{multipleauthorsCredibilitySignalsUnofficial}. In this way, we grouped all signals related to a title into a single category named \textit{Clickbaits and title representativeness}, all signals related to claims into a category of \textit{Check-worthy and fact-checked claims} as well as reused a subcategory of \textit{References and citations}. Second, we grouped together signals which refer to the same or highly-similar concepts and just use a different terminology, e.g., grammar and spelling errors, grammar, spelling or punctuation mistakes, punctuation naturally fall under the category of \textit{Text quality}; or profanity, incivility, impoliteness and hate speech under the category of \textit{Offensive language}. In some cases, the signals refer to the same concept, but in an opposite meaning, like originality and attribution of non-original content belong to the same category of \textit{Originality and content reuse}. In some cases, the signals refer to the same concepts but are defined on different granularity levels, e.g., logical fallacies, logical flaws, ad-hominem attacks all together belong under \textit{Persuasion techniques and logical fallacies}. Third, we grouped together signals where the borders between them are not clearly defined and in practice they correlate with each other, e.g., subjectivity, political bias, confirmation bias, emotional tone were grouped under \textit{Factuality, subjectivity and bias}. We also jointly combined these strategies, especially the category of \textit{Factuality, subjectivity and bias} groups signals that are closely related, commonly refer to the same concepts but from opposite angles, or are at different levels of abstractions.
    \item In this way, we finally proposed 8 categories of credibility signals (see also Figure \ref{fig:credibility_signals_taxonomy}; more detailed description of each category is provided in the later sections): (i) Factuality, subjectivity and bias; (ii) Persuasion techniques and logical fallacies; (iii) Check-worthy and fact-checked claims; (iv) Text quality; (v) References and citations; (vi) Clickbaits and title representativeness; (vii) Originality and content reuse; and (viii) Offensive language.
    \item In addition, we recognized one more category of signals -- (ix) Machine-generated text -- that has become very relevant only recently with the emergence of LLMs and, therefore, it did not appear in the existing lists/taxonomies. At the same time, it is a strong predictor of credibility as fully generated text can contain many factual errors either as a result of unintended hallucinations of LLMs or due to their intended misuse to generate false content (the previous works demonstrated significant vulnerabilities of LLMs to generate disinformation~\cite{vykopal-etal-2024-disinformation}, even personalized one~\cite{zugecova2024evaluationllmvulnerabilitiesmisused}).
\end{enumerate}

\begin{figure}
    \centering
    \resizebox{\textwidth}{!}{

    \begin{forest}
        forked edges,
        for tree={
            grow=east,
            reversed=true,
            anchor=base west,
            parent anchor=east,
            child anchor=west,
            base=left,
            font=\small,
            rectangle,
            draw=draw,
            rounded corners,
            align=left,
            minimum width=5em,
            edge+={darkgray, line width=1pt},
            s sep=3pt,
            inner xsep=5pt,
            inner ysep=4pt,
            ver/.style={rotate=90, child anchor=north, parent anchor=south, anchor=center},
        },
        where level=1{text width=11em,font=\scriptsize}{},
        where level=2{text width=6.8em,font=\scriptsize}{},
        where level=3{text width=6.8em,font=\scriptsize}{},
        [Textual Credibility Signals, ver
            [{\textbf{Factuality, subjectivity and bias}}
                [{
                Factual accuracy~\cite{nakov2021survey,molinaFakeNewsNot2021},
                Event factuality~\cite{9310484, cao-etal-2021-uncertain, rovera-etal-2025-modafact},
                Subjectivity~\cite{podgurskiEvaluatingWebContent2022},
                Selection and\\presentation bias~\cite{nakov2021survey},
                Hyperpartisanship / political bias~\cite{multipleauthorsCredibilitySignalsUnofficial},
                Confirmation bias~\cite{multipleauthorsCredibilitySignalsUnofficial}, \\
                Framing~\cite{nakov2021survey}, 
                Emotional tone~\cite{10.1145/3184558.3188731},
                Formal / Informal tone~\cite{multipleauthorsCredibilitySignalsUnofficial},
                Emotional, sensational or affective \\ language~\cite{podgurskiEvaluatingWebContent2022}, 
                Emotionally charged~\cite{molinaFakeNewsNot2021,multipleauthorsCredibilitySignalsUnofficial},
                Emotional Valence~\cite{multipleauthorsCredibilitySignalsUnofficial}
                Use of hyperboles~\cite{molinaFakeNewsNot2021}, \\
                Impartial reporting~\cite{molinaFakeNewsNot2021}, 
                One-sided reporting~\cite{molinaFakeNewsNot2021},
                Discrepancies or omissions~\cite{molinaFakeNewsNot2021}
                }, leaf-single, text width=30em]
            ]
            [\textbf{Persuasion techniques and} \\ \textbf{logical fallacies}
                [{
                Logical fallacies~\cite{10.1145/3184558.3188731},
                Logical flaws~\cite{molinaFakeNewsNot2021},
                Inference~\cite{10.1145/3184558.3188731},
                Ad-hominem attacks~\cite{molinaFakeNewsNot2021}, \\
                Conspirational reasoning~\cite{molinaFakeNewsNot2021},
                Arguments from authority~\cite{molinaFakeNewsNot2021},
                Common man appeals~\cite{molinaFakeNewsNot2021}, \\
                Call to Action (Political)~\cite{multipleauthorsCredibilitySignalsUnofficial}
                }, leaf-single, text width=30em]
            ]
            [\textbf{Check-worthy and} \\ \textbf{fact-checked claims}
                [{
                Fact-checked~\cite{10.1145/3184558.3188731,molinaFakeNewsNot2021},
                Fact-check status of claim~\cite{multipleauthorsCredibilitySignalsUnofficial}, 
                Article has a central claim~\cite{multipleauthorsCredibilitySignalsUnofficial}
                }, leaf-single, text width=30em]
            ]
            [Text quality 
                [{
                Grammar and spelling errors~\cite{podgurskiEvaluatingWebContent2022},
                Grammar, spelling or punctuation mistakes~\cite{molinaFakeNewsNot2021,multipleauthorsCredibilitySignalsUnofficial}, \\
                Readability~\cite{podgurskiEvaluatingWebContent2022},
                Punctuation~\cite{podgurskiEvaluatingWebContent2022},
                Vocabulary or reading level~\cite{multipleauthorsCredibilitySignalsUnofficial}
                }, leaf-single, text width=30em]
            ]
            [References and citations 
                [{
                Quotes from outside experts~\cite{10.1145/3184558.3188731},
                Citation of organizations and studies~\cite{10.1145/3184558.3188731}, \\
                Representative citations~\cite{10.1145/3184558.3188731},
                External links~\cite{podgurskiEvaluatingWebContent2022},
                No quotes or made-up quotes~\cite{molinaFakeNewsNot2021}, \\
                No source attribution~\cite{molinaFakeNewsNot2021,multipleauthorsCredibilitySignalsUnofficial},
                Uses standardized references or citations~\cite{multipleauthorsCredibilitySignalsUnofficial}, \\ 
                Few to zero references or citations~\cite{multipleauthorsCredibilitySignalsUnofficial} 
                }, leaf-single, text width=30em]
            ]
            [Clickbaits and title representativeness 
                [{
                Clickbait title~\cite{10.1145/3184558.3188731,podgurskiEvaluatingWebContent2022,multipleauthorsCredibilitySignalsUnofficial},
                Misleading and clickbait headlines~\cite{molinaFakeNewsNot2021}, \\
                Title representativeness~\cite{10.1145/3184558.3188731,multipleauthorsCredibilitySignalsUnofficial}
                }, leaf-single, text width=30em]
            ]
            [Originality and content reuse 
                [{
                Originality~\cite{10.1145/3184558.3188731},
                Originality types~\cite{multipleauthorsCredibilitySignalsUnofficial},
                Attribution of Non-Original Content~\cite{multipleauthorsCredibilitySignalsUnofficial}
                }, leaf-single, text width=30em]
            ]
            [Offensive language 
                [{
                Profanity~\cite{podgurskiEvaluatingWebContent2022},
                Incivility and impoliteness~\cite{multipleauthorsCredibilitySignalsUnofficial},
                Hate speech~\cite{multipleauthorsCredibilitySignalsUnofficial}
                }, leaf-single, text width=30em]
            ]
            [Machine-generated text 
                [{No appearances in the existing lists/categorizations yet 
                }, leaf-single, text width=30em]
            ]
        ]
    \end{forest}
    }

    \caption{Unified categorization of textual credibility signals. Light blue nodes consist of credibility signals depicted in the existing taxonomies and categorizations. Bold font highlights categories of credibility signals on which this survey put a focus on (see Section~\ref{section:background-scope} for more information).}
    \label{fig:credibility_signals_taxonomy}
\end{figure}

\textbf{Importance and effects of credibility signals on end users}. Due to a broad definition of credibility, as well as rich taxonomies of credibility signals, a number of human-based studies have analysed the importance of individual credibility signals and their effects on the perceived credibility, in general as well as for specific groups of end users (according to their expertise, age, education etc.).

First, these studies address various content types and platforms, such as social media posts \cite{10.1145/3613905.3650801, 10.1145/1806338.1806455}, news articles \cite{lu2022effects, 10.1145/3342220.3343670,10.1145/3635147}, videos and video sharing platforms \cite{10.1145/3613904.3642490, feng2023examining}, images \cite{feng2023examining}, search engines \cite{10.1145/3576840.3578277, 10.1145/3613904.3642059, 10.1145/3201064.3201095}, or online encyclopedias \cite{10.1145/3555051.3555052}. 

At the same time, they address various types of end users. \citet{piccolo2021agents} perform a human study with two groups of participants, academics/students and random social media users. \citet{RIJO2023107619} looked into how voting patterns and political beliefs influence the way people evaluate information credibility. \citet{10.1145/3555096} focused on liberal and conservative users. Other works (e.g., \cite{10.1145/3613904.3642490}) focus on social media users in general. 

Regarding the scope of credibility signals, some works focus on signals provided directly by (social media) platforms~\cite{10.1145/3576840.3578277,10.1145/3201064.3201095}, some investigate a single credibility signal (e.g., hyper-partisanship in \cite{10.1145/3555096}) or a whole spectrum of signals (e.g., 28 signals analysed in  \cite{piccolo2021agents}). \citet{10.1145/3555096} compared credibility signals produced by three various sources -- an algorithm, a community, and a third-party fact-checker. Besides content- and context-based signals, \citet{CHANG2021106629} considered also an additional group of design signals (consisting of four signals: interaction design, interface design, navigation design, and security settings). While the most of these studies focused on English content, the number of studies comparing the importance of credibility signals on non-English data is rather limited. The work by \citet{GAO201521} performs the human study on the importance of credibility signals for the Weibo Chinese misinformation dataset regarding health and safety information. 

All these humans studies resulted into valuable findings about which credibility signals are the most important and have the greatest effect on the end-users. Out of 28 credibility signals, \citet{piccolo2021agents} found that context-based (presence of links, publisher, author) contribute most towards human judgement. Interestingly, different groups of participants exhibit different patterns towards content-based signals. For example, fallacy and partisanship, as well as clickbait title, were ranked high by academics and students only, while social media users did not find this information important. In the human study on e-Health information, \citet{CHANG2021106629} found that content signals were most important, accounting for 57.3\% of cases, followed by source-related or context-based signals (26.0\%). Finally, they found that design-related signals, such as website layouts and overall design are least important and account for 16.7\% of decisions. \citet{LIN2016264} found that the authority and retweet patterns are most important for human participants when deciding on information credibility. \citet{GAO201521} showed that the impact of source credentials, such as the authenticity of the authors, on users’ credibility assessment was limited. On the other hand, content-based signals, such as objective claims, are considered more insightful for credibility assessment.

\subsection{Existing surveys}

The existing surveys on automatic credibility assessment focus on specialised domains. \citet{8572695} conducted a literature review on the credibility of social media information while considering 4 levels of credibility assessment (i.e., post, member, topic/event, hybrid level). The surveyed methods were categorised into automated-, human-, and hybrid-based approaches. Automated-based approaches are further subcategorised into machine learning approaches, graph-based semisupervised approaches, and weighted and IR algorithms. Their analysis spans until 2018, and therefore, the automated methods are limited to traditional machine learning (e.g., classifications algorithms like decision trees, logistic regression) and do not include more recent transformer or LLM-based approaches. Similarly, \citet{9543698} perform an analysis of the credibility detection methods in the social media and microblog domains. The authors adopt the level-based classification of the task proposed by \citet{8572695}. In addition, they propose a fine-grained framework based on the existing approaches that targets various aspects behind credibility assessment. Their theoretical framework consists of content-, media-, website- and interaction-based assessment of social media information. Within this framework, they identify two main directions of research, namely, {\em post-level} and {\em user-level} constructs. The post-level constructs are further divided into the detection of: (i) deceptive and trustful information, (ii) bias and objectivity detection, and (iii) hate speech and offensive language detection tasks. The user-based constructs, on the other hand, are based on: (i) user deception level, (ii) user competence, and (iii) user authority.

Another group of the surveys focuses on particular categories of credibility signals. For example, \citet{nakov2021survey} surveyed the literature on factual accuracy and biases in social media, \citet{PANCHENDRARAJAN2024100066} analysed claim detection for automated fact checking, or \citet{wu-etal-2025-survey} addressed machine-generated text detection.

In contrast to these existing surveys, we focus on in-depth analysis of \textit{Natural Language Processing (NLP)} research, with a specific focus on \textit{automatic credibility assessment and detection of individual textual credibility signals} in the context of \textit{online social media content}. This focus is motivated and determined by the recent emergence of large language models, that have significantly influenced the NLP field. The research works on credibility assessment are already adopting this new generation of models, nevertheless, their potential is still far from being explored completely. This survey thus stands on the intersection of credibility assessment (which is an application domain of utmost importance for protecting our society and democratic values) and artificial intelligence (which is a research domain rapidly advancing thanks to many recent discoveries).

\subsection{Scope of the survey}
\label{section:background-scope}

At first, this survey aims to systematically cover research that explicitly addresses the entire process of automatic credibility assessment (i.e., both steps of credibility signals detection as well as their aggregation into a single credibility label/score). However, restricting our review solely to such works would overlook a substantial body of research focused on the detection of individual categories of credibility signals. Although these studies may not explicitly frame their contributions within the context of credibility assessment, i.e., they often do not adopt the relevant credibility-related terminology or perform aggregation of multiple signals, they provide valuable insights and methods that are highly relevant to the broader credibility assessment framework.

More specifically, we define the scope of this survey as follows:

\begin{enumerate}
    \item Out of \textit{methods to credibility assessment}, this survey specifically tackles \textit{automation-based approaches}.
    \item Regarding the \textit{levels of credibility assessment}, the research works falling into the scope of this survey typically address the credibility assessment at the \textit{content level}, but indirectly also at user/source, topic/event, media and hybrid levels. 
    \item Out of various \textit{taxonomies of credibility signals}, the survey covers both \textit{content- as well as context-based signals}, predicting \textit{high as well as low credibility} that are text-based (while relation to other modalities is briefly documented in Section~\ref{section:additional-signals-non-nlp}).
    \item Regarding the categorization of credibility signals, the survey systematically covers all 9 categories of textual credibility signals from the unified categorization of textual credibility signals (as shown in Figure \ref{fig:credibility_signals_taxonomy}).
\end{enumerate}

Out of 9 categories of textual credibility signals, we put a specific focus and provide an in-depth analysis for three of them, namely:
    \begin{enumerate}
        \item Factuality, subjectivity and bias; 
        \item Persuasion techniques and logical fallacies; 
        \item Check-worthy and fact-checked claims.
    \end{enumerate}

This selection is motivated by several factors. (i) The selected categories have attracted a considerable amount of interest in the NLP community, resulting in a substantial body of work that is suitable for systematic review. Moreover, the authors of this survey have direct practical experience in training and deploying detection models for these specific signals, which enables a more informed and application-oriented analysis and discussion. (ii) Although some of these categories have been partially addressed in the previous surveys, they have not been examined through the lens of credibility and credibility assessment, nor in terms of their integration with other signal detection approaches. Importantly, these categories offer strong potential for effective methodological integration. For example, subjectivity, bias, and persuasion techniques share underlying linguistic features, which can be exploited through transfer learning or multitask learning applied to shared datasets. (iii) The selected categories have been also identified in human-centered studies as particularly influential in shaping perceptions among social media users and domain experts (e.g., journalists and media professionals) (see, e.g., \citet{piccolo2021agents} and \citet{CHANG2021106629}, and Section~\ref{section:background-dimensions} for further discussion).

Regarding the \textit{time span}, the survey covers the research papers published before May 2024, when the systematic identification of research works was done. The first papers covered by the survey originate in 2004.

\section{Methodology}

\subsection{Research paper collection process}

\begin{figure}[!tbp]
    \centering
    \includegraphics[width=\linewidth]{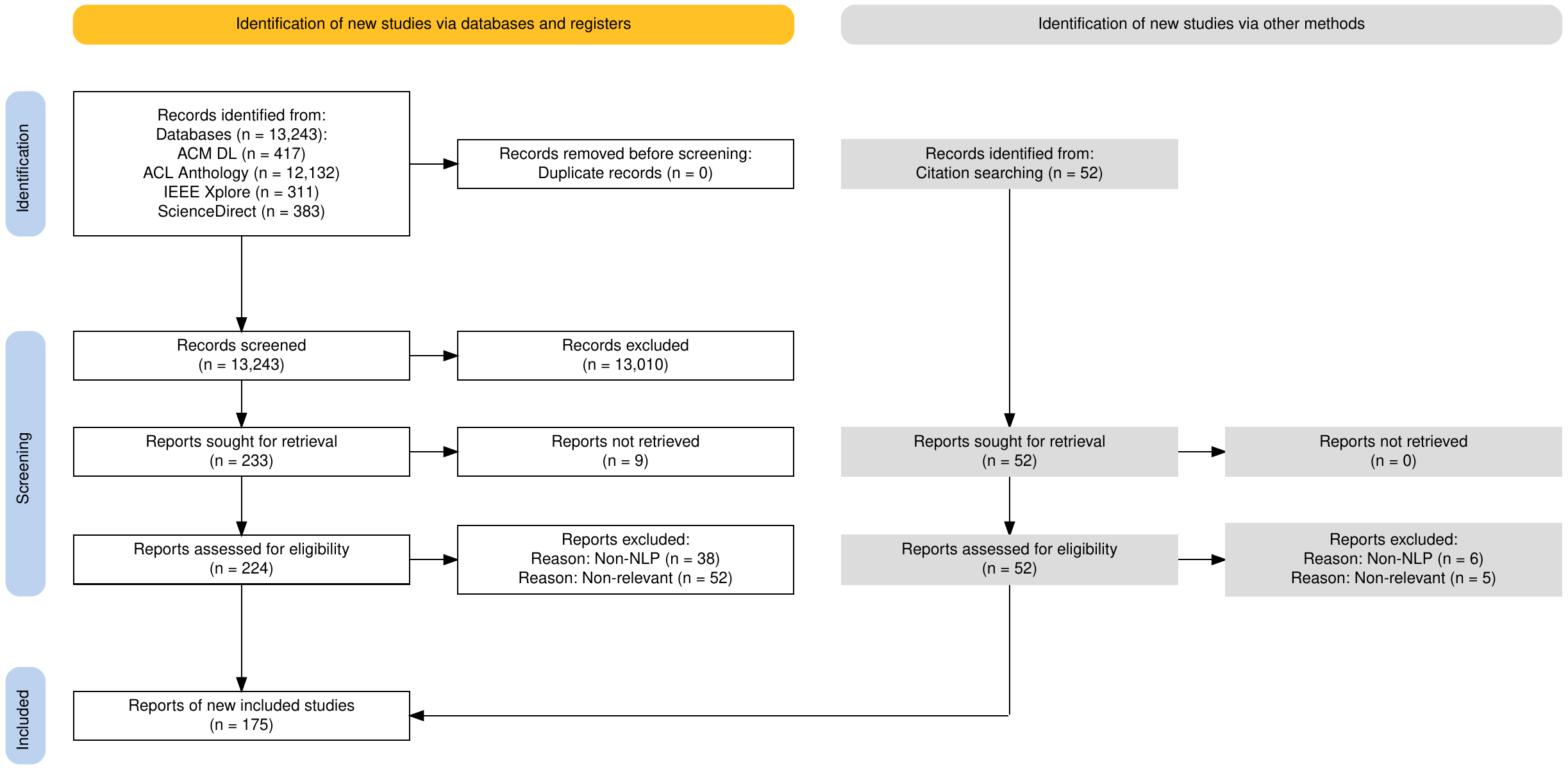}
    \caption{PRISMA 2020 flow diagram \cite{prisma2020diagram} depicting the standardized methodology applied to collect the relevant research papers, together with a number of research papers processed in each step.}
    \label{fig:prisma}
\end{figure}

In order to collect the relevant papers, we adopted the conventional and standardized methodology for literature review called \textit{Preferred Reporting Items for Systematic Reviews and Meta-Analyses (PRISMA)} \cite{takkouche2011prisma} and specifically its updated version PRISMA 2020 \cite{prisma2020}. In addition to being commonly used in literature surveys, PRISMA was also previously used to perform the analysis of the topics related to credibility signals, such as media bias \cite{rodrigo2023systematic}. Figure~\ref{fig:prisma} depicts the standardized PRISMA 2020 flow diagram \cite{prisma2020diagram} illustrating individual steps and the number of research papers processed.

\begin{table}[!tbp]
    \centering
    \footnotesize

    \caption{Search queries used during the systematic literature search}
    \label{tab:query}

\resizebox{\columnwidth}{!}{
\begin{tabular}{|l|l|l|l|}
\hline
\multicolumn{1}{|c|}{\textbf{Survey section}}                                             & \multicolumn{1}{c|}{\textbf{Description}}                                                                                                                                                                                                                     & \multicolumn{1}{c|}{\textbf{Search query}}                                                                                                                                                                                                                                                                                                                                                                                                                                                                                                                           & \multicolumn{1}{c|}{\textbf{Search scope}} \\ \hline
\begin{tabular}[c]{@{}l@{}}Automatic credibility \\ assessment \end{tabular}                                                           & \begin{tabular}[c]{@{}l@{}}Generic credibility assessment approaches \\ based on credibility signal detection \\ and their aggregation \end{tabular}                                                                                                                                                                  & \begin{tabular}[c]{@{}l@{}}"credibility signals" OR \\ "credibility indicators" OR \\ "confidence indicators" OR \\ "non-credibility index" OR \\ "noncredibility index" OR \\ "credibility features"\end{tabular}                                                                                                                                                                                                                                                                                                                                                   & \begin{tabular}[c]{@{}l@{}}Title, abstract, keywords, \\ the whole article body\end{tabular}                     \\ \hline
Factuality, subjectivity and bias                                                                    & \begin{tabular}[c]{@{}l@{}}Credibility signal category comprising \\ factuality, subjectivity, various forms \\ of biases, objectivity-reporting, \\ partisanship, tone, emotional valence, \\ framing\end{tabular}                                                                                                                   & \begin{tabular}[c]{@{}l@{}}(“factuality” OR \\ “text subjectivity” OR \\ “text objectivity” OR \\ "subjective bias" OR \\ "sentiment bias" OR \\ ”media bias” OR \\ “news framing” OR \\ “partisan”) \\ AND \\ (detection OR \\ classification OR \\ identification OR \\ extraction)\end{tabular}                                                                                                                                                                                                                                                                   & Title, abstract, keywords                              \\ \hline
\begin{tabular}[c]{@{}l@{}}Persuasion techniques and \\ logical fallacies\end{tabular} & \begin{tabular}[c]{@{}l@{}}Credibility signal category comprising \\ persuasion techniques, logical fallacies, \\ propaganda \end{tabular}                                                                                                                                                                   & \begin{tabular}[c]{@{}l@{}}("propaganda" OR \\ "persuasion" OR \\ "logical fallacies") \\ AND \\ (detection OR \\ classification OR \\ identification OR \\ extraction)\end{tabular}                                                                                                                                                                                                                                                                                                                                                                                 & Title, abstract, keywords                              \\ \hline
\begin{tabular}[c]{@{}l@{}}Check-worthy and \\ fact-checked claims\end{tabular}                                                                   & \begin{tabular}[c]{@{}l@{}}Credibility signal category comprising \\ check-worthy claims, previously \\ fact-checked claims\end{tabular}                                                                                                                                                                        & \begin{tabular}[c]{@{}l@{}}("check-worthy" OR \\ "fact-checked") \\ AND \\ (claim OR \\ statement OR \\ misinformation) \\ AND \\ (detection OR \\ classification OR \\ identification OR \\ extraction)\end{tabular}                                                                                                                                                                                                                                                                                                                                                & Title, abstract, keywords                              \\ \hline
Additional credibility signals                                                         & \begin{tabular}[c]{@{}l@{}}Machine-generated text, \\ Text Quality, \\ References and citations, \\ Clickbaits and title representativeness, \\ Originality and content reuse, \\ Offensive language\end{tabular} & \begin{tabular}[c]{@{}l@{}} Search is done on \\ad-hoc basis, \\ no systematic keywords \\were used \end{tabular}                                                                                                                                                                                                                                                                                                                                                                                                                                                                                            & Title, abstract, keywords                              \\ \hline
\end{tabular}}
\end{table}

\textbf{Identification -- Search queries}. To conduct a systematic literature search aimed at identifying papers with high potential relevance, we formulated custom search queries (see Table~\ref{tab:query}) targeting two main parts of this survey:
(i) studies that explicitly address the credibility assessment process, including both the detection of credibility signals and their aggregation, and
(ii) studies focused on the automatic detection of the three selected individual categories of credibility signals. Each query combined keywords and logical search operations (\textit{AND}, \textit{OR}) to achieve the highest possible coverage and reduce false positives at the same time. The keywords have been carefully selected to capture inconsistent and ambiguous terminology used in the existing works. The first version of queries was proposed following our past experience and research in the respective area, and iteratively refined to cover additional alternative phrases with the same semantics identified in the research papers. If a search query returned a paper that was more appropriately aligned with a different category of credibility signals, the paper was reassigned accordingly. Thanks to the use of mutually distinctive categories of credibility signals, no duplicates were present across the identified sets of papers. Additional categories of credibility signals were covered through ad hoc searches, conducted without predefined search queries.

\textbf{Identification -- Databases and registers}. The proposed queries were submitted to four large repositories of peer-reviewed scientific publications, namely: (i) ACL Anthology\footnote{\url{https://aclanthology.org/}}; (ii)  ScienceDirect\footnote{\url{https://www.sciencedirect.com/}}; (iii) ACM DL\footnote{\url{https://dl.acm.org/}}; and (iv) IEEE Xplore\footnote{\url{https://ieeexplore.ieee.org/}}.

\textbf{Identification -- Search scope}. Papers explicitly addressing credibility assessment and credibility signals detection are quite scarce, therefore, for this type of papers we performed the search for the query terms in the whole body of the document. For individual categories of credibility signals, on the other hand, we searched for the given query term in the title, abstract and user-specified keywords\footnote{The search in author-specified keywords is available only at the ScienceDirect digital library.} only.

For ACM Digital Library (n = {\identificationACM}), IEEE Xplore (n = {\identificationIEEE}) and ScienceDirect (n = {\identificationScienceDirect}) we considered all returned papers as identified. Since ACL Anthology uses Google Search as a search engine, the search returned an extensive number of results, many of them being irrelevant. By a manual check, we found out that the relevant results appeared in first 10 pages or less, therefore for each search query we considered as identified papers that appear on first 10 pages of results (still resulting in a large number of {\identificationACL} identified papers).

\textbf{Identification -- Extended search}. Besides identification of the relevant research papers by the direct search, we applied also citation searching (i.e., search for additional potentially-relevant papers in references from/to the papers identified by the direct search). To some extent, we also checked additional papers published by the same author and at the same venue (e.g., in case of data challenges that are directly focused on the respective category of credibility signals). In this way, we identified additional papers (n = {\identificationExtendedSearch}), that partially comes from the pre-print servers, especially arXiv. Being a pre-print (a paper that have not underwent the peer-review yet), we carefully checked the quality of such specific cases to make sure that only high-quality papers are included in our survey.

\textbf{Screening}. After identifying all papers matching the search queries (n = {\identificationTotal}), we performed an initial screening of titles and abstracts directly within the user interfaces of the respective digital libraries. Based on this step, we excluded non-relevant papers (n = {\screeningRecordsExcluded}). Papers were excluded at this stage if they: (i) used the search keywords in a context that falls outside the scope of this survey as defined in Section~\ref{section:background-scope} (e.g., not related to online content credibility or NLP), or (ii) were not full, short, journal, or workshop papers (e.g., abstracts of invited talks or workshops).

From the remaining pool of potentially relevant papers (n = {\screeningRecordsSought}), the full-text PDF could not be retrieved for only a negligible number (n = {\screeningRecordsNotRetrieved}).

For all papers with an accessible full text (n = {\screeningRecordsAssessed}), a final eligibility assessment was performed to exclude any remaining irrelevant papers not identified during the initial screening. Based on an in-depth analysis of the full papers, exclusions were made for the following two reasons:
\begin{itemize}
    \item Non-NLP (n = {\screeningRecordsExcludedNonNLP} for direct search, n = {\screeningOtherReportsExcludedNonNLP} for extended search):  These papers explored credibility or credibility signals from theoretical, behavioral, human annotation, or other non-NLP perspectives (e.g., computer vision, social network analysis).
    \item Non-relevant (n = {\screeningRecordsExcludedNonRelevant} for direct search, n = {\screeningOtherReportsExcludedNonRelevant} for extended search): These papers mentioned the search terms but in contexts unrelated to credibility signals (e.g., “confidence indicator” in construction risk assessment). Such papers were excluded from further analysis.
\end{itemize}

\textbf{Included}.  Ultimately, we identified a set of relevant NLP papers (n = {\includedTotal}) that addressed either credibility assessment or one of selected credibility signal categories, with a primary focus on the NLP domain (i.e., papers that proposed automatic detection or aggregation methods for credibility signals, introduced relevant datasets, or developed systems for automatic credibility assessment).

\begin{figure}[!tbp]
  \centering
  \begin{minipage}[b]{0.40\textwidth}
    \includegraphics[width=\textwidth]{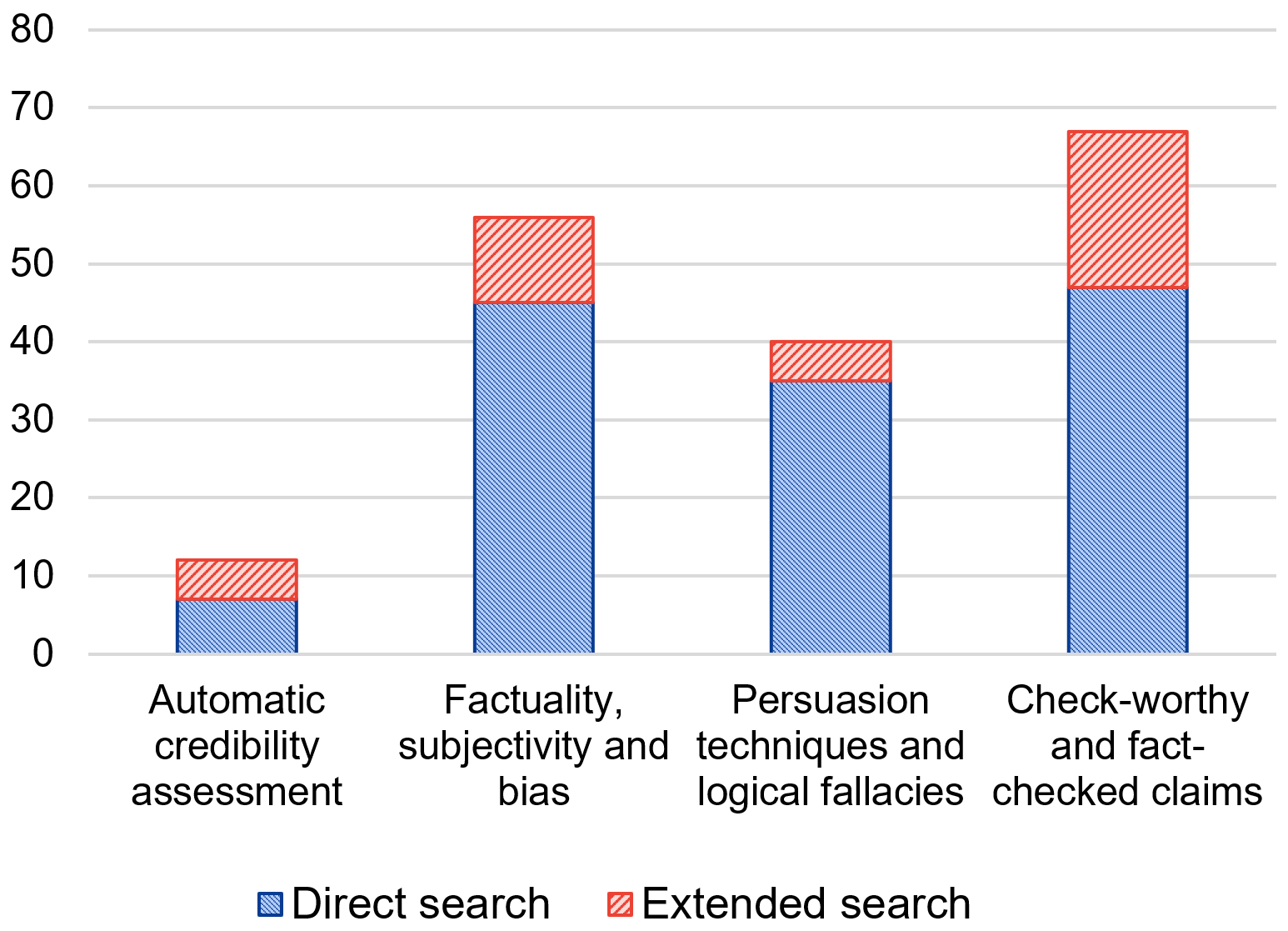}
    \caption{Number of included papers according the direct (by keywords) and extended (by citations) search.}
    \label{fig:stats-by-category-direct-extended-search}
  \end{minipage}
  \hfill
  \begin{minipage}[b]{0.57\textwidth}
    \includegraphics[width=\textwidth]{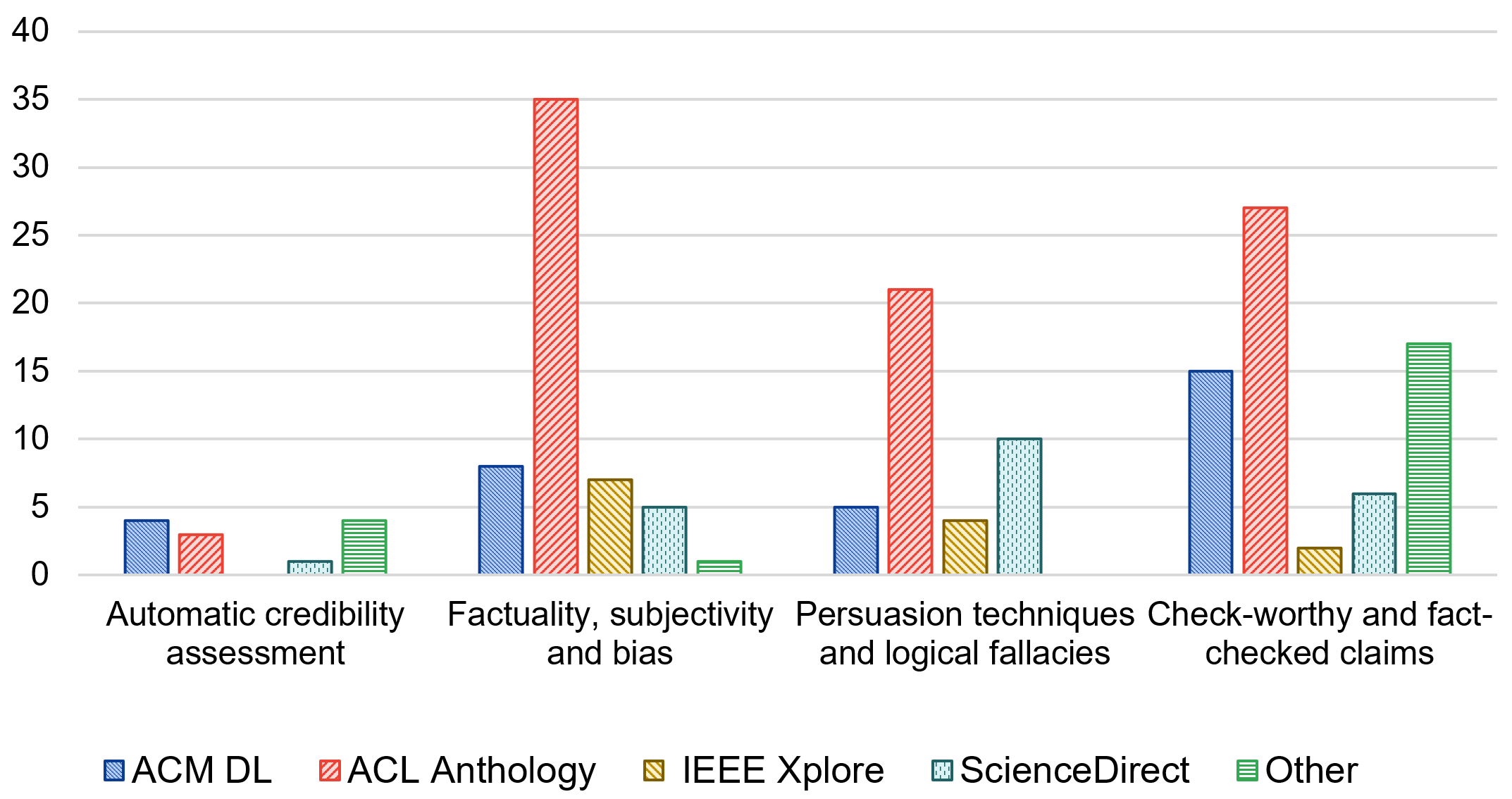}
    \caption{Number of included papers according to the digital library grouped by each credibility signal category. Other refers to additional repositories, such as arXiv.}
    \label{fig:stats-by-category-dl}
  \end{minipage}
\end{figure}

Among the included works, the smallest group consists of papers that explicitly address credibility assessment, whereas the most populated category is check-worthy and fact-checked claims (see Figure~\ref{fig:stats-by-category-direct-extended-search}). Due to the survey’s focus on NLP, the ACL Anthology emerges as the most prominent source for all three individual categories of credibility signals. In contrast, papers addressing credibility assessment are more evenly distributed across digital libraries (see Figure~\ref{fig:stats-by-category-dl}).

\begin{figure}[!tbp]
    \centering
    \includegraphics[width=0.6\linewidth]{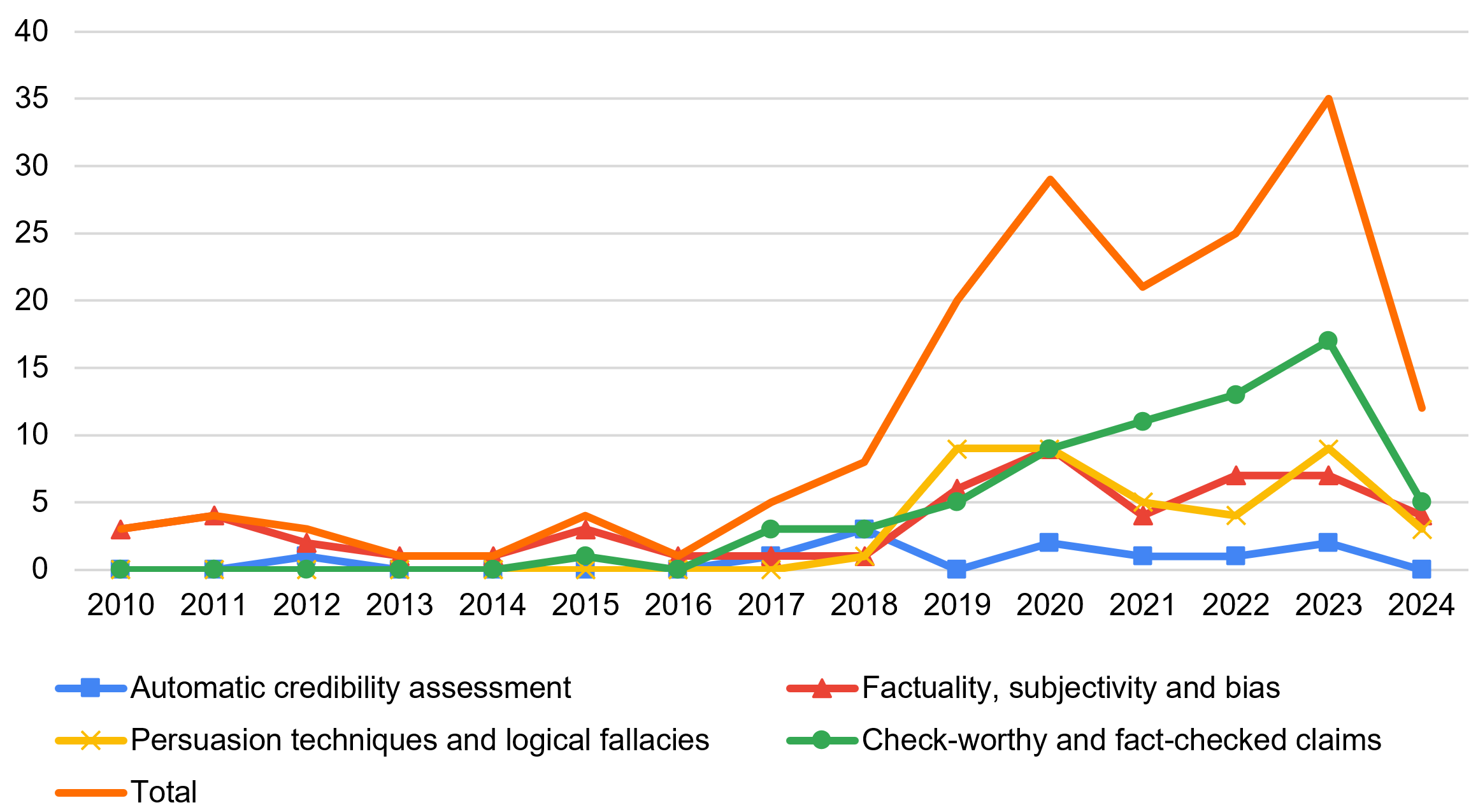}
    \caption{Evolution of included papers according to year they have been published in. Year 2024 includes papers published online before May 2024.}
    \label{fig:stats-by-year}
\end{figure}

In terms of publication trends, we observe an increase beginning in 2016, likely corresponding to heightened research attention following the U.S. presidential elections (see Figure~\ref{fig:stats-by-year}). The continued growth in subsequent years -- despite a temporary slowdown during the COVID-19 pandemic -- demonstrates sustained and growing interest in the topic addressed by this survey.

\subsection{Paper categorization and in-depth analysis}

Included papers were subsequently thoroughly analysed by the authors of this survey. Firstly, for each paper the basic metadata have been collected:

\begin{itemize}
    \item paper type (one of the following options: a full paper, a short paper, a journal paper, a workshop paper, other),
    \item venue (an abbreviation of a conference, a journal or a pre-print),
    \item year, when the paper was published in,
    \item main outcome (one of the following options: a method/model, a study, a dataset, a tool, a survey).
\end{itemize}

Moreover, we annotated several additional types of information that have been identified as the most useful for the analysis and description of the current state of the art, as well as for recognition of remaining open problems and potential future work, namely:

\begin{itemize}
    \item dataset used (if authors used an existing dataset, a name of such dataset, otherwise information about how own dataset was obtained),
    \item content/context signals (whether the paper tackles with content, context or both types of credibility signals, see Section~\ref{section:background-dimensions} for more information), 
    \item list of credibility signals (a specific list of signals addressed in the paper),
    \item automated detection (whether a paper attempts to perform also an automatic credibility assessment and/or automatic detection of credibility signals),
    \item models (if paper attempts to perform automated detection, what kind of models have been used),
    \item metrics (if paper attempts to perform automated detection, what kind of metrics have been used),
    \item human agreement (if paper introduces an annotated dataset, what kind of annotator mutual agreement was achieved, measured by standard metrics, e.g., by Cohen’s Kappa).
\end{itemize}

The full list of papers included in this survey, including all annotations is available as a supplementary material to this article in the ACM Digital Library\footnote{A full list of papers is available as supplementary material to this article in the ACM Digital Library and at \url{https://kinit.sk/public/acm-tist-credibility-assessment-survey.html}}. 

Following this analysis, we describe credibility assessment papers as well as each individual category of credibility signals from three perspectives (that are also utilized to structure the consequent categories): (i) datasets, (ii) methods and models, and (iii) tools and services. For each category we summarised the current state of the art and identified problems and challenges that are specifically present in such credibility signal category (such analysis is further elaborated in Section~\ref{section:challenges-and-open-problems}, with an orthogonal discussion across all included papers).

\section{Automatic credibility assessment}
\label{section:credibility-assessment}

This section provides an overview of research efforts that address automatic credibility assessment, encompassing both steps of the process: the detection and subsequent aggregation of credibility signals. In some cases, however, the features used as input to the credibility assessment algorithms are relatively shallow, relying primarily on surface-level linguistic characteristics (e.g., the number of unique words), which only approximate underlying credibility signals rather than explicitly capturing them. This group of approaches explicitly uses the terminology relevant to credibility (such as credibility assessment, credibility signal, credibility score), in contrast to approaches described in the following sections that address detection of a particular category of credibility signals, but typically do not denote them as such.

\subsection{Datasets}
\label{section:credibility-assessment-datasets}

Most of datasets used in the credibility assessment works are annotated at the content level (see Section \ref{section:background-dimensions} for the overview of levels of credibility assessment). Table~\ref{tab:credibility_assessment_datasets} provides an overview of the selected datasets created or adapted for the purpose of credibility assessment task.

\citet{10.1145/3184558.3188731} created a thoroughly annotated, however, only very small credibility-related dataset. Out of news articles, being the most shared on social media, 40 articles were selected (covering multiple topics: public health, climate science, diseases, vaccines). In total, 6 annotators were recruited to annotate the credibility signals. These indicators were adopted from the initial list of signals created by the \textit{Credibility Coalition} (see Section \ref{section:background-dimensions}). Out of them, 16 content and context indicators were selected to be annotated. Namely, content indicators included: title representativeness, clickbait title, quotes from outside experts, citation of organizations and studies, calibration of confidence, logical fallacies, tone, inference. Context indicators included: originality, fact-checked, representative citations, reputation of citations, number of ads, spammy ads, number of social calls, and finally placement of ads and social calls. Besides that, all articles were evaluated for an overall credibility by domain experts on a 5-point scale. In the following dataset analysis, two backward stepwise multiple regression models were utilized to measure the predictive value provided by both sets of indicators. For content-based signals, after model convergence, two variables remained: clickbait title and logical fallacies (slippery slope). This model was found to significantly predict credibility (F = 13.972, p < 0.001). For context-based signals, 6 variables remained: fact-checked–reported false, fact-checked–reported mixed results, number of social calls, number of mailing list calls, and placement of ads and social calls. Together, they were also found to significantly predict credibility (F = 12.986, p < 0.001). Despite a wide range of annotated credibility signals as well as overall credibility score, the practical value of the resulting dataset, which was created only as a proof of concept, is somehow limited by its size (N = 40) which is insufficient even in few-shot learning scenarios.

In the follow-up study, \citet{10.1145/3415164} gathered a dataset of over 4,000 credibility assessments taken from 2 crowd groups (journalism students and Upwork workers) and 2 expert groups (journalists and scientists). Similarly as in the previous case, a varied set of 50 news articles related to climate science were selected. News articles were annotated by both crowd and expert groups on a 5-point Likert scale. The in-depth analysis of the obtained annotations revealed differences in annotation not only between crowd and expert groups, but also within expert groups between journalists and scientists due to differing expert criteria that journalism versus science experts use. Following the observations, authors proposed directions how to better design crowdsourcing of content credibility.  

\citet{el-ballouli-etal-2017-cat} collected 17 million tweets in a period of two weeks that were written in Arabic and contained at least one hashtag. After preprocessing and grouping tweets by their hashtags, a topic-independent subset of 9,000 tweets was selected. Each of these tweets was annotated by three annotators on a binary scale (credible, non-credible), while the majority vote was used to determine the final label.

Due to a lack of sufficiently large credibility-annotated datasets, some authors (e.g., \cite{10.1016/j.ins.2019.12.040,podgurskiEvaluatingWebContent2022,leiteDetectingMisinformationLLMPredicted2023}) decided to use available false information (fake news) datasets, such as LIAR \cite{wang2017liar}, Weibo \cite{10.1016/j.ins.2019.12.040}, FakeNewsNet \cite{doi:10.1089/big.2020.0062}, FakeNewsAMT \cite{perez-rosas-etal-2018-automatic} or Celebrity dataset \cite{perez-rosas-etal-2018-automatic}. In such a case, various false information labels are mapped to credibility labels by taking an assumption that a false content is considered to be non-credible. \citet{QURESHI2023109028} also proceeded from the FakeNewsNet dataset, however, only to identify a potential set of tweets that were further manually annotated by 12 experts for credibility on a 5-point scale. The resulting dataset consists of 4,958 tweets, out of them some were discarded due to low annotators' agreement.

Another group of datasets used in the existing works relates to credibility annotated at a source (webpage) level. Microsoft Credibility \cite{schwarz2011augmenting} dataset consists of top 40 search results on 25 pre-defined queries on the topics of Health, Politics, Finance, Environmental Science, and Celebrity News, which were manually annotated for credibility on a 5-point scale. Another dataset, Content Credibility Corpus (C3) \cite{kakol2017understanding}, consists of 15,750 evaluations of 5,543 pages by more than 2,000 annotators recruited through Amazon Mechanical Turk. The textual credibility signals annotated in this dataset include readability, language quality, informativeness, completeness, and objectivity.

Finally, at the topic/event level, \citet{Mitra_Gilbert_2021} introduced a large scale dataset containing 60M tweets collected during a period of more than three months and grouped into 1,049 real-world events, each annotated by 30 human annotators on a 5-point scale.

\begin{table}[!h]
\caption{Selected datasets used in the works addressing automatic credibility assessment}
\label{tab:credibility_assessment_datasets}
\footnotesize

\resizebox{\columnwidth}{!}{
\begin{tabular}{|l|c|c|c|c|c|}
\hline
    \textbf{Dataset} & \textbf{Language(s)} & \textbf{\# Instances} & \textbf{Content type} & \textbf{Classes} & \textbf{Used by} \\ \hline
    
    \citet{10.1145/3184558.3188731} & English & 40 & News articles & \begin{tabular}[c]{@{}c@{}} 8 content-based signals, \\ 8 context-based signals, \\ overall credibility on a 5-point scale \end{tabular} & \cite{10.1145/3184558.3188731,podgurskiEvaluatingWebContent2022} \\ \hline

    \citet{10.1145/3415164} & English & 50 & News articles & \begin{tabular}[c]{@{}c@{}} overall credibility on a 5-point scale \end{tabular} & \cite{10.1145/3415164} \\ \hline
    
    \citet{el-ballouli-etal-2017-cat} & Arabic & 9,000 & Tweets & \begin{tabular}[c]{@{}c@{}} credible, non-credible \end{tabular} & \cite{el-ballouli-etal-2017-cat}\\\hline
     
    LIAR \cite{wang2017liar} & English & 12,800 & \begin{tabular}[c]{@{}c@{}} Short statements \\ from politifact.com \end{tabular} & \begin{tabular}[c]{@{}c@{}} pants on fire, false \\ barely-true, half-true \\ mostly-true, true \end{tabular} & \cite{10.1016/j.ins.2019.12.040} \\ \hline

    Weibo \cite{10.1016/j.ins.2019.12.040}  & Chinese & 18,000 & Microblogs & \begin{tabular}[c]{@{}c@{}} same 6 as in LIAR \end{tabular} & \cite{10.1016/j.ins.2019.12.040} \\ \hline

    FakeNewsNet \cite{doi:10.1089/big.2020.0062} & English & 23,196 & News articles, microblogs & \begin{tabular}[c]{@{}c@{}} true news, false news \end{tabular} & \cite{podgurskiEvaluatingWebContent2022,leiteDetectingMisinformationLLMPredicted2023} \\ \hline

    FakeNewsAMT \cite{perez-rosas-etal-2018-automatic} & English & 480 & Political news articles & \begin{tabular}[c]{@{}c@{}} fake, legitimate \end{tabular} & \cite{leiteDetectingMisinformationLLMPredicted2023} \\ \hline
    
    Celebrity \cite{perez-rosas-etal-2018-automatic} & English & 500 & Celebrity news articles & \begin{tabular}[c]{@{}c@{}} fake, legitimate \end{tabular} & \cite{leiteDetectingMisinformationLLMPredicted2023} \\ \hline

    \citet{QURESHI2023109028} & English & 4,958 & Microblogs & \begin{tabular}[c]{@{}c@{}} overall credibility on a 5-point scale \end{tabular} & \cite{QURESHI2023109028} \\ \hline

    Microsoft Credibility \cite{schwarz2011augmenting} & English & 1,000 & \begin{tabular}[c]{@{}c@{}} Websites (search results) \end{tabular} &  \begin{tabular}[c]{@{}c@{}} overall credibility on a 5-point scale \end{tabular} & \cite{esteves2018belittling,podgurskiEvaluatingWebContent2022} \\ \hline

    Content Credibility Corpus (C3) \cite{kakol2017understanding} & English & 5,691 & Websites & 25 signals grouped into 6 categories & \cite{esteves2018belittling,podgurskiEvaluatingWebContent2022} \\ \hline

    CREDBANK \cite{Mitra_Gilbert_2021} & English & 60M & Microblogs & \begin{tabular}[c]{@{}c@{}} credibility of 1,049 real-world events \\ annotated on a 5-point scale \end{tabular} & \\ \hline
\end{tabular}}
\end{table}

\subsection{Methods and models}

The methods and models used in the research works falling into this group typically employ a methodology consisting of three steps. Firstly, a set of credibility signals is selected. Secondly, various techniques are employed for their automatic detection. Finally, detected credibility signals are then used to predict an overall credibility which can be subsequently utilized for example as an input to information retrieval engine. In the following description, we provide an in-depth description of each of these three steps.

\textbf{Credibility signals selection}. In the earliest works, credibility signals were rather approximated by various (mostly shallow) linguistic features that can be easily detected by the machine, such as a number of words or exclamation marks. While being potentially helpful for traditional machine learning algorithms, people would not usually use such features for credibility assessment due to the absence of their direct interpretability. \citet{weerkamp2008credibility} adopted a subset of 11 such features for blog posts proposed in \cite{Rubin2006AssessingCO} while focusing on those that are text-based and can be easily detected automatically, such as \textit{capitalization}, \textit{shouting}, \textit{spelling} or \textit{post length}. Similarly, \citet{10.1145/2166966.2166998} opted for 12 numeric and 7 binary content-based credibility indicators for Twitter, such as \textit{a positive sentiment factor} or \textit{a number of mentions}. Approximately half of these indicators were adopted from the previous work by \citet{10.1145/1963405.1963500}.

\citet{el-ballouli-etal-2017-cat} selected 48 credibility signals, out of them 26 signals were content-based. Besides shallow linguistic features, as already included in the previous works (e.g., \textit{count of hashtags}, \textit{count of unique words}, \textit{count of exclamation/question marks}) the authors utilized also more advanced sentiment extraction to assess \textit{positive sentiment}, \textit{negative sentiment}, and \textit{objectivity}. A shift towards more complex credibility signals is also present in the work by \citet{QURESHI2023109028}. Authors initially complied a list of 33 potential credibility signals. By proceeding from the initial experiments, a final set of 11 content- and context-based signals were used. In the case of content-based signals, shallow features (such as \textit{a number of hashtags}, \textit{length of text}) were discarded in favor of more complex signals like, \textit{post hate}, \textit{informativeness}, or \textit{deception}.

For the purpose of website credibility evaluation, \citet{esteves2018belittling} selected 15 content-based indicators by proceeding from the taxonomy created by \citet{10.1007/978-3-642-36973-5_47}, such as \textit{authority} (authoritative keywords within the page HTML content), \textit{readability metrics}, \textit{text category} or \textit{sentiment}.

In contrast to works mentioned so far, the following works selected signals from the list created by the Credible Web Community Group (see Section \ref{section:background-dimensions}). At first, \citet{podgurskiEvaluatingWebContent2022} proposed a modular credibility assessment system of web pages using the 23 most frequent content- and context-based signals. The employed signals range from naive numerical and linguistic features, such as a \textit{number of external links}, \textit{linguistic statistics} (average text and word count, overall word count, etc.), \textit{spelling and grammar}, \textit{domain} endings and \textit{author information}, to more complex features, such as  sentiment, readability and presence of a clickbait title. Similarly, \citet{leiteDetectingMisinformationLLMPredicted2023} employed 18 credibility signals, including advanced ones that have not been considered in the previous works, such as \textit{call to action}, \textit{impoliteness}, \textit{sensationalism} or \textit{explicitly unverified claims}.

\textbf{Credibility signal detection}. Early shallow linguistic features were detected primarily by simple rule- and heuristic-based techniques \cite{weerkamp2008credibility,10.1145/2166966.2166998}, such as a predefined lexicon of positive and negative words, counting/searching for specific characters/patterns (e.g., capital letters, emoticons).

By proceeding from such superficial linguistic features to more advanced credibility signals, also their detection methods became more complex. For sentiment analysis various lexicon and rule-based sentiment analysis were employed, such as Vader for English \cite{esteves2018belittling,podgurskiEvaluatingWebContent2022}, or ArSenL for Arabic \cite{el-ballouli-etal-2017-cat}. For text category classification, well-known approaches employed in other NLP tasks were adopted, such binary multinomial Naïve Bayes (NB) classifiers or Latent Semantic Analysis (LSA) \cite{esteves2018belittling}. Various NLP libraries were used for automatic detection of additional credibility signals, such as LanguageTool library for detection of grammar \& spelling errors or TextBlob for subjectivity detection \cite{podgurskiEvaluatingWebContent2022}. Even further, \citet{QURESHI2023109028} detected the selected credibility signals by means of the classification methods proposed in the previous research works.

Surprisingly, adoption of LLMs for detection of credibility signals is only very rare despite a great potential of LLMs to target a wide range of various credibility signals categories. To this end, \citet{leiteDetectingMisinformationLLMPredicted2023} used 3 instruction-tuned LLMs (GPT-3.5-Turbo, Alpaca-LoRA-30B and OpenAssistant-LLaMa-30B) with a specific prompt designed for each credibility signals (e.g., ``Does the article make use of sensationalist claims?'').

\textbf{Credibility label/score prediction}. To estimate overall credibility from the credibility signals, \citet{weerkamp2008credibility} opted for simple weighting schemata. The resulting credibility score was subsequently incorporated into a blog post retrieval model. \citet{10.1145/2166966.2166998} proposed three computational models (based on a weighted combination, and on a probabilistic language-based approach) for assessing tweet's credibility, using not only content-based but also social and hybrid strategies. Experiments on the test set consisting of 1,023 instances (while using 10-fold cross-validation) revealed that social model was able to outperform content-based and hybrid model. This can be explained by a short textual content of tweets, as well as very shallow linguistic and possibly noisy credibility signals, such as \textit{a presence of exclamation mark}. \citet{podgurskiEvaluatingWebContent2022} combined individual signals' subscores with signal weights through a linear combination function. Signal weights were determined by authors empirically by considering prior research results, signal measurement accuracy, and experimental calculations on test data.

By advancing from simple weighting and heuristics, consequent works started to employ feature engineering with traditional machine learning approaches. \citet{el-ballouli-etal-2017-cat} trained a random forest decision tree classifier on the top of detected credibility signals -- each serving as a feature. To predict website credibility score, \citet{esteves2018belittling} used Gradient Boosting and AdaBoost classification algorithms on the top of Microsoft Credibility and Content Credibility Corpus (C3).

For the purpose of feature selection, \citet{QURESHI2023109028} utilized the Kolmogorov–Smirnov (KS) test to identify discriminating features. Subsequently, a wide range of machine learning algorithms (12 regression and 10 classification ones) were used to predict original 5-scale as well as simplified binary credibility label. Results revealed that context-bases signals, such as high user influence and medium-to-high spread count, are very indicative of credible tweets. Credible tweets are normally spread by topic experts (the topic of the tweet is in the top 3 topics for that user), typically do not contain deceptive information and are not posted by malicious accounts. In terms of content-based signals, credible tweets have high informativeness and tent to not contain hate speech or links to low-credible sources.

Analogically to other ML/NLP areas, also credibility assessment underwent a shift towards deep learning methods. In this direction, \citet{10.1016/j.ins.2019.12.040} introduced an ANSP model based on adversarial networks and multi-task learning to capture differential credibility features for information credibility evaluation. The input to the model consists of the concatenations of word embeddings (Word2Vec) and meta-data embeddings (various credibility signals provided by the datasets). Evaluation on English and Chinese datasets, LIAR and Weibo respectively, showed that the combination of all features performs best for both datasets, while content-based features were consistently outperformed by contextual ones. Among these, some signals are particularly powerful on their own, such as  the speaker, state info, party affiliation, or credit history.

Finally, \citet{leiteDetectingMisinformationLLMPredicted2023} used Prompted Weak Supervision (PWS) on the top of LLM-predicted credibility signals. This approach was compared with unsupervised and supervised baselines. In the supervised finetuning scenario, a RoBERTa-Base model was finetuned with the ground-truth labels. In the zero-shot scenario, an instruction-tuned LLM based on LLaMa2 was prompted without any signals or in-context examples. 
Experimental results on four datasets (FakeNewsNet - GossipCop, FakeNewsNet - PolitiFact, FakeNewsAMT, and Celebrity) showed that Prompted Weak Supervision outperformed the zero-shot baseline by $38.3\%$, and achieved $86.7\%$ of the performance of the state-of-the-art supervised baseline. Furthermore, in cross-domain settings, where the domain of the train set differs from the domain of the test set (e.g., Politics and Gossip), Prompted Weak Supervision outperformed the supervised baseline by $63\%$. Lastly, the authors study the association between credibility signals and veracity through (i) a statistical test, where $12$ out of the $19$ signals were shown to have an association with veracity, and (ii) an ablation study, in which signals were individually removed from the train set in order to inspect their contribution to the model's performance. This showed that the method's strength is in the combination of a wide range of signals rather than relying on a small set of signals that could be strong predictors of veracity. Through these analyses, the authors verified that some credibility signals are domain-specific. For example, signals such as \textit{Source Credibility} and \textit{Misleading about Content} improve performance mostly for the political domain, while others such as \textit{Expert Citation} and \textit{Call to Action} show benefits in entertainment news. In addition to the efficiency of this approach in terms of being independent of the costly annotation of long news articles with each signal, the main advantage of this method is generalisation capability. The authors found that their approach outperforms the fine-tuned methods on the out-of-domain test data. This result is particularly important in the constantly-evolving online environment, when new non-credible and false information emerge every day, making it impossible to have up-to-date annotated data.

\subsection{Tools and services}

While several services dedicated to credibility assessment exist, many of them are on a border with false information detection. From 2019, the Credibility Coalition maintain a \textit{CredCatalog}\footnote{\url{https://credibilitycoalition.org/credcatalog/}} -- a catalogue of initiatives that have a stated aim to improve information quality. Besides various organizations (like fact-checking or academic institutions), it provides overview of  relevant tools.

Besides that, there are commercial projects supporting end-users in evaluating the credibility of the online content directly in their browsers. NewsGuard\footnote{\url{https://www.newsguardtech.com/}} provides \textit{News Reliability Ratings} for news outlets based on nonpartisan journalistic criteria. Similarly, GroundNews\footnote{\url{https://ground.news/}} rate news stories according to their bias distribution (left-/right-leaning bias) or factuality (in the meaning of source reliability and general factual accuracy). The labelling is, however, done rather indirectly through sources writing about such news stories, instead of analysing a content of individual news articles themselves.

Another relevant tool is the \textit{Tanbih}~\cite{zhang-etal-2019-tanbih} -- a news aggregator that performs media-level and article-level analyses, aiming to help users better understand the content they consume. It organizes news articles into event-based clusters and generates media outlet profiles, which include indicators such as the general factuality (in the meaning of factual accuracy) of reporting, the degree of propagandistic content, hyper-partisanship, leading political ideology, overall framing, and the stance of the outlet toward various claims and topics.

Finally, an automatic analysis of several credibility signals (persuasion techniques, subjectivity, or machine-generated text among others) is available as a part of the \textit{Assistant} tool in the well-known \textit{Verification plugin}\footnote{\url{https://www.veraai.eu/category/verification-plugin}} developed and further enhanced as a part of EU-funded projects (\textit{InVID}, \textit{WeVerify} and \textit{vera.ai}). The Assistant tool allows users to provide a URL or a local file as input, from which the text is extracted and various text analysis are performed, credibility signals being a recent addition to them.

\subsection{Discussion}
\label{section:credibility-assessment-discussion}

\textbf{Limited research focus with absence of LLM-based solutions}. 
As already shown in Figure~\ref{fig:stats-by-category-direct-extended-search}, the number of works explicitly addressing automatic credibility assessment is considerably lower in comparison with other individual signal categories. This is in contrast to false information detection, which attracted a plethora of research (and also practitioners' and public) attention.

In parallel, our literature survey pointed out a lack of approaches utilizing LLMs. At the same time, LLMs provide a great opportunity to automatize detection of multiple credibility signals even in a zero-shot settings, as the work by \citet{leiteDetectingMisinformationLLMPredicted2023} clearly demonstrated. Even further improvement in the terms of accuracy can be achieved by employing in-context learning, or instruction-tuning LLMs. While such LLM-based approaches would inquire higher computational costs, existing Parameter-efficient Fine Tuning (PEFT) techniques \cite{xu2023parameterefficientfinetuningmethodspretrained} may be employed if necessary.

\textbf{Lack of multilingual datasets with annotated credibility signals}. Similarly to a limited research focus, also the situation with credibility-annotated datasets is falling behind false information research. There are very few datasets with manual (human) annotations at the level of individual credibility signals as well as overall credibility \cite{10.1145/3184558.3188731,10.1145/3415164}. Unfortunately, these datasets are very small, which prevents the use of language models even in few-shot settings. Despite the declared original intentions to extend them in future, to the best of our knowledge, no follow-up datasets have been created so far.

The remaining datasets are larger (containing hundreds or thousands of instances), nevertheless, annotated only at the level of overall credibility (see Table \ref{tab:credibility_assessment_datasets}). Absence of datasets providing annotations of multiple categories of credibility signals for the same set of instances, prevents multi-task training of models. Multi-task approaches hold considerable potential, particularly because many credibility signals inherently share underlying similarities that can enhance model performance, as already demonstrated by the results of Prompted Weak Supervision (PWS) in \cite{leiteDetectingMisinformationLLMPredicted2023}.

\textbf{Robustness of credibility signals}. For reliable credibility assessment, it is necessary that credibility signals and their relation to overall credibility remains the same in time, or at least is not influenced by significant domain and data drifts. To shed more light on the robustness of credibility signals, \citet{JI2023103210} evaluated how the signals and their prediction capability change over time. By utilizing the posts from the Weibo platform, two topic-specific datasets were created, one for climate change and second one for genetically modified organisms (GMO). The authors analysed how credibility signals evolve over time by splitting these datasets year-wise. They found that certain signals remain more stable over time, however, this stability is topic-dependent. For example, post sentiment performed better for the climate change dataset in most of the years compared with other features, while the post topic was the best predictor for the GMO dataset. In general, results showed that content-based features remained effective across time. On the other hand, contextual features, such as activeness and gender, were not correlated with veracity in the climate change dataset, while popularity showed no significant correlation with GMO misinformation in any given year from 2010 to 2020. In summary, content-based signals (especially sentiment and topic) seem to be more robust and less prone to the temporal data drift compared to context-based signals.
\section{Factuality, subjectivity and bias}
\label{section:factuality-subjectivity-bias}
Within false information research, the term {\em factuality} has been used to refer to a variety of related problems, with no unified ontology to define its levels. For example, \citet{nakov2021survey} equate factuality with factual accuracy (i.e., veracity) and propose a four-level ontology encompassing the claim, article, user, and source factuality levels. In this interpretation, factuality is seen not as as a credibility signal, but as the final verdict regarding truthfulness of information. 
In the context of large language models (LLMs), factuality is typically defined as a model's ability to generate outputs grounded in established real-world facts \cite{wang2023survey}. This aligns with the Cambridge English Dictionary’s definition of factual as “based on or containing facts”\footnote{\url{https://dictionary.cambridge.org/dictionary/english/factual}}.
From an NLP perspective, however, factuality is often approached differently. Rather than relying solely on external fact-checking, NLP literature tends to treat factuality as a contextual signal -- reflecting the degree of certainty a speaker conveys about the occurrence of events \cite{9310484, cao-etal-2021-uncertain, qian-etal-2019-document, li2024maven,rovera-etal-2025-modafact}. As identified through the systematic literature review conducted as part of this work, factuality in this view functions as a credibility signal rather than an overall veracity assessment, expressing how confidently an event is presented as having occurred or not occurred. 
This more fine-grained perspective follows earlier work by \citet{sauri2009factbank}, which defines factuality as the extent to which events are portrayed as corresponding to real-world situations, hypothetical scenarios, or uncertain interpretations.
Under this definition, the task of factuality detection can be used as the first step towards identifying check-worthy claims (discussed in Section~\ref{section:claims-and-veracity}), as only the events that the speaker is confident about typically form claims. In the rest of this section, we use the terms factuality and event factuality interchangeably. Where the meaning is not obvious from the context or could lead to ambiguity, we distinguish whether event factuality, factual accuracy or a different notion of factuality is meant there.

The literature on event factuality traditionally distinguishes between individual event factuality detection (EFD) \cite{9310484} and  document-level event factuality identification (DEFI) \cite{cao-etal-2021-uncertain} tasks. The main distinction between these two sub-tasks is in the scope of information concerned -- while the EFD task considers individual event mentions, DEFI aims to aggregate several mentions of a certain event within a document.

\textit{Subjectivity} represents an indicator of the overall subjectivity (or objectivity) of information. The task of identifying subjective information is modelled as either a binary or a more fine-grained problem that distinguishes between various degrees of subjectivity. Over the past years, there have been many shared tasks that focus on specific types of subjectivity. For example, \citet{piskorski-etal-2023-semeval} discriminate between objective reporting, opinionated news and satire. \citet{derczynski2017semeval} and \citet{gorrell2019semeval} organised the challenges for rumour detection. Irony \cite{derczynski2017semeval} and sarcasm \cite{gorrell2019semeval} is another common type of non-factual information that received close attention during the last years, with shared tasks and challenges dedicated to this problem.

Finally, \textit{bias} represents a similar but more complex phenomenon of imbalance in terms of opinions or facts. As highlighted by \cite{nakov2021survey}, there is no one single concept of bias among scholars. However, in many cases bias is seen as a systematic favouring of a certain ideology when covering information~\cite{waldman1998newspaper}. This can be expressed in deliberately withholding certain part of information that contradict the favourable point of view \cite{Smith2001FromPT} or vice versa, specifically searching for the facts that cover information from a certain political or ideological viewpoint \cite{Hassell2020ThereIN}. In some cases, even if the choice of information sources is unbiased, the presentation of information in those sources may be performed in a biased manner that highlights the importance of only certain facts. Some instances of biased presentation include framing \cite{piskorski-etal-2023-semeval, entman2007framing}, opinionated reporting style \cite{piskorski-etal-2023-semeval, soroka2012gatekeeping} and even visual clues \cite{Barrett2005BiasIN} used to distort the perception of information.

\subsection{Datasets}
\begin{table}[tbp]
\footnotesize
\centering

\caption{Selected datasets used for event factuality detection.}
\label{tab:factuality_data}
\resizebox{\columnwidth}{!}{
\begin{tabular}{|l|c|c|c|c|}
\hline
\textbf{Dataset} &
  \textbf{Language(s)} &
  \textbf{\# Instances} & \textbf{Content type} & \textbf{Classes} \\ \hline
\citet{qian-etal-2019-document} & English, Chinese & \begin{tabular}[c]{@{}c@{}} 1,948 (English) \\ 4,649 (Chinese) \end{tabular} & \begin{tabular}[c]{@{}c@{}}News articles from China Daily, \\Sina Bilingual News, and Sina News.\\ Sentence-level event factuality \end{tabular} & \begin{tabular}[c]{@{}c@{}} 
CT-: negated events\\
PS+: speculative events \\
PS-: speculative  negative events\\
Uu: events appear in question\\
CT+: factual events
\end{tabular} \\ \hline
\citet{qian2022document}&English, Chinese & \begin{tabular}[c]{@{}c@{}} 1,948 (English) \\ 4,649 (Chinese) \end{tabular} & \begin{tabular}[c]{@{}c@{}} Extension of \citet{qian-etal-2019-document}\\ with individual\\ and document-level event\\annotations\end{tabular} & \begin{tabular}[c]{@{}c@{}} 
CT-: negated events\\
PS+: speculative events \\
PS-: speculative  negative events\\
Uu: events appear in question\\
CT+: factual events
\end{tabular}  \\ \hline
\citet{li2024maven}&English&112,276 events& Documents and events &\begin{tabular}[c]{@{}c@{}} 
CT-: negated events\\
PS+: speculative events \\
PS-: speculative  negative events\\
Uu: events appear in question\\
CT+: factual events
\end{tabular} \\ \hline

\end{tabular}}
\end{table}

\begin{table}[t!]
\footnotesize
\centering
\caption{Selected datasets used for subjectivity detection.}
\label{tab:subjectivity_data}

\resizebox{\columnwidth}{!}{
\begin{tabular}{|l|c|c|c|c|}
\hline
\textbf{Dataset} &
  \textbf{Language(s)} &
  \textbf{\# Instances} & \textbf{Content type} & \textbf{Classes} \\ \hline
  \citet{piskorski-etal-2023-semeval} & \begin{tabular}[c]{@{}c@{}}English, German,\\French, Italian,\\Russian, Polish\end{tabular} & 1,592 news articles & \begin{tabular}[c]{@{}c@{}} News articles  \end{tabular} &\begin{tabular}[c]{@{}c@{}} Objective, Opinionated, Satire\end{tabular}   \\\hline
  \citet{SPINDE2023100264} & English & \begin{tabular}[c]{@{}c@{}} 2,800 news articles\\ 175,807 comments
and \\retweets referring \\to these articles\end{tabular} & News articles, tweets & \begin{tabular}[c]{@{}c@{}} Hateful vs Neutral\end{tabular} \\ \hline

\citet{BIYANI2014170} & English &  700 & \begin{tabular}[c]{@{}c@{}} Threads from two popular online \\forums,  Trip Advisor–New York\\ and Ubuntu. %The subjectivity \\ was manually annotated. 
\end{tabular} & Subjective vs Non-subjective \\ \hline
\citet{wiebe2005annotating} & English & \begin{tabular}[c]{@{}c@{}}10,657 sentences \\(535 documents)\end{tabular} & \begin{tabular}[c]{@{}c@{}} 187 different news sources \end{tabular} & \begin{tabular}[c]{@{}c@{}}  Objective\\
Expressive subjective \\
Direct subjective
\end{tabular} \\ \hline
\citet{spinde-etal-2021-neural-media} & English & 3,700 & \begin{tabular}[c]{@{}c@{}} Sentences collected from \\news organizations  with \\ different political leaning. \end{tabular} & \begin{tabular}[c]{@{}c@{}} Opinionated, Factual, or Mixed \end{tabular} \\ \hline
\citet{banea-etal-2010-multilingual}  & \begin{tabular}[c]{@{}c@{}}English, Arabic, \\French, German, \\Romanian, Spanish \end{tabular} & 9,700 in each language & \begin{tabular}[c]{@{}c@{}}This is an extension of \\MPQA dataset \cite{wiebe2005creating} created \\ by translating sentence-level \\data into other languages. \\All information is parallel.\end{tabular} & \begin{tabular}[c]{@{}c@{}} Objective vs Subjective\end{tabular} \\ \hline
\citet{atalla2011investigating} & German & 6,848 & \begin{tabular}[c]{@{}c@{}} Sentence-level annotation of \\news articles. Created to \\be compatible with MPQA.\end{tabular} & \begin{tabular}[c]{@{}c@{}} Objective vs Subjective\end{tabular} \\ \hline
\citet{wiebe2005creating} & Urdu & \begin{tabular}[c]{@{}c@{}} 500 articles: \\700 sentences annotated \\with emotion and 4,000 \\unbiased sentences \end{tabular} & News articles from BBC Urdu & \begin{tabular}[c]{@{}c@{}} Objective vs Subjective\end{tabular} \\ \hline
\citet{mourad2013subjectivity} & Arabic & 2,300 & \begin{tabular}[c]{@{}c@{}} Tweets published 2012, \\randomly sampled\end{tabular} & \begin{tabular}[c]{@{}c@{}} Neutral, Positive, Negative, \\ Both, Sarcastic \end{tabular} \\ \hline
\citet{jeronimo2020computing}& Portuguese & 450 words & \begin{tabular}[c]{@{}c@{}}Discourse markers for \\subjectivity in Portuguese\end{tabular} & \begin{tabular}[c]{@{}c@{}}Argumentation, Presupposition\\Modalization, Sentiment\\Valuation \end{tabular}\\ \hline
\citet{maks-vossen-2012-building}&Dutch&11,000--56,000 tokens & \begin{tabular}[c]{@{}c@{}} Lexicon for subjectivity in Dutch\\ based on Wikipedia articles \\and user comments \end{tabular} & \begin{tabular}[c]{@{}c@{}}Actor and
Speaker/Writer \\ subjectivity \end{tabular}\\ \hline
\citet{pang-lee-2004-sentimental}&English&1,000 & Movie reviews & Subjective vs Non-subjective\\ \hline
\end{tabular}}
\end{table}

\begin{table}[h!]
\footnotesize
\centering
\caption{Selected datasets used for bias detection.}
\label{tab:bias_data}

\resizebox{\columnwidth}{!}{
\begin{tabular}{|l|c|c|c|c|}
\hline
\textbf{Dataset} &
  \textbf{Language(s)} &
  \textbf{\# Instances} & \textbf{Content type} & \textbf{Classes} \\ \hline
  \citet{piskorski-etal-2023-semeval} & \begin{tabular}[c]{@{}c@{}}English, German,\\French, Italian,\\Russian, Polish\end{tabular} & 1,592 news articles & \begin{tabular}[c]{@{}c@{}} News articles  \end{tabular} &\begin{tabular}[c]{@{}c@{}} 14 presentation frames \end{tabular}   \\\hline
\citet{SPINDE2021102505} & English & 1,700 & \begin{tabular}[c]{@{}c@{}} Short statements\\ %span-annotated bias labels 
\end{tabular} & Biased vs non-biased \\\hline
\citet{spinde-etal-2021-neural-media} & English & 3,700 & \begin{tabular}[c]{@{}c@{}} Sentences collected from \\news organizations  with \\ different political leaning \end{tabular} & \begin{tabular}[c]{@{}c@{}}Biased vs non-biased \\ Word-level bias annotation. \end{tabular} \\ \hline
\citet{liu-etal-2019-detecting}&English& 2,990 &\begin{tabular}[c]{@{}c@{}}US news articles \\from 2018 annotated \\in terms of frames \\ based on headlines only\end{tabular} &\begin{tabular}[c]{@{}c@{}} \textbf{4 general
frames}: Politics;\\Public opinion; Society/Culture\\ Economic
consequences\\ \textbf{5 issue-specific frames:} Race/Ethnicity \\2nd Amendment (Gun Rights);\\ Gun control;
Mental health;\\ School/Public space safety\\
\end{tabular} \\ \hline
\citet{fan-etal-2019-plain} & English & 300 news articles & \begin{tabular}[c]{@{}c@{}} News articles \\from FOX, NYT and HPO \end{tabular} & \begin{tabular}[c]{@{}c@{}} Informational and Lexical Bias \\ (Sentence, token level) \end{tabular} \\ \hline
\citet{chen-etal-2020-analyzing} & English & 6,964 news articles & \begin{tabular}[c]{@{}c@{}} Articles from 41 publishers \\ Labels derived from AllSides \end{tabular} &\begin{tabular}[c]{@{}c@{}} Bias Detection \\ Unfairness Detection \\ (different levels of text granularity) \end{tabular}  \\ \hline
\citet{aksenov-etal-2021-fine} & German & 47,362 news articles & \begin{tabular}[c]{@{}c@{}} Articles from 36 publishers  \end{tabular} &\begin{tabular}[c]{@{}c@{}} Fine-grained Bias Detection \\ (different levels of text granularity) \end{tabular}  \\
\hline
\end{tabular}}
\end{table}
Tables~\ref{tab:factuality_data}, \ref{tab:subjectivity_data}, and ~\ref{tab:bias_data} represent a summary of the papers that introduce datasets annotated for detecting event factuality, subjectivity, and bias respectively. As can be seen, the prevalent majority of the datasets are only available in English.  Besides English, there are datasets in Urdu~\cite{mukund2010vector}, Arabic~\cite{mourad2013subjectivity} and German~\cite{atalla2011investigating, aksenov-etal-2021-fine}. Furthermore, multilingual subjectivity detection was a part of shared tasks at CheckThat! Labs of CLEF 2023~\cite{10.1007/978-3-031-28241-6_59} and CLEF 2024~\cite{10.1007/978-3-031-56069-9_62}, covering Arabic, German, English, Italian, and Turkish languages.

The ontology proposed by \citet{qian-etal-2019-document} is the most widely accepted classification for detecting event factuality, with subsequent extension of the dataset with document-level event factuality annotations \cite{qian2022document}. More recently, \citet{li2024maven} built the first large-scale annotated dataset based on this classification, by using LLM predictions with human judgments as a final step. 

MPQA opinion corpus~\cite{wiebe2005annotating} is a particularly widespread benchmarking dataset for subjectivity detection, appearing in the majority of studies covered by the systematic review~\cite{lin-etal-2011-sentence, wiegand2011convolution, wang2011cross, BIYANI2014170}. The dataset is manually annotated with three frames, \textit{objective}, \textit{expressive subjective} and \textit{direct subjective}. The dataset contains span-level annotations along with the annotation of the source (author, specific person, etc.) who expresses the subjective frame and the intensity of subjectivity. Direct subjective expressions are
typically more explicit than expressive subjective. Originally created in English for detecting a phrase-level subjectivity, it is widely used in multilingual tasks by utilising parallel translations into other languages, such as Arabic, French, German, Romanian and Spanish~\cite{banea-etal-2010-multilingual, mogadala2012language}. In addition, some of the datasets adopted the ontology of MPQA to create comparable corpora in other languages \cite{atalla2011investigating}.

In addition to the datasets for bias detection provided in Table~\ref{tab:bias_data}, \citet{10.1145/3539618.3591882} introduced Media Bias Identification Benchmark (MBIB) collection, which is the most comprehensive set of the benchmark corpora for media bias detection consisting of 22 datasets. The types of biases covered include \textit{hate speech}, \textit{lexical}, \textit{contextual}, \textit{linguistic}, \textit{gender}, \textit{cognitive}, \textit{racial} and \textit{political} biases. 3 out of 22 datasets, however, represent a more high-level annotation of information into \textit{fake} and \textit{trustworthy} rather than biases.

\subsection{Methods and models}
The methods discussed in this section can be categorized into three groups -- event factuality, subjectivity and bias/framing detection techniques. The methods in the first group can be further divided into two subcategories: (i) those targeting the factuality of individual event mentions and (ii) those assessing the document-level factuality of events. 

\textbf{Event factuality detection (EFD)}. The methods falling into this category aim to identify the author's level of certainty regarding the possibility of the individual mentioned event \cite{9310484, 7451542, 9892869, 9892209,lee-etal-2015-event}. The most common architectures for event factuality detection are LSTM and bi-LSTM models trained on BERT representations \cite{9310484, 9892869, 9892209}. A few studies also employ traditional machine learning approaches, such as SVM with LOSSO regression \cite{lee-etal-2015-event} and a combination of rule-based and maximum entropy methods \cite{7451542}. All the reviewed methods for event factuality detection are trained on either English \cite{7451542, lee-etal-2015-event} or Chinese \cite{9310484, 9892209}, or a combination of both languages \cite{9892869}.

\textbf{Document-level event factuality identification (DEFI).} Sentence-level event factuality often results in conflicts within a document, as different mentions of the same event may reflect varying degrees of factuality. Therefore, the methods in this sub-group  aim to conclude the overall event factuality in a document based on the various sentence-level factuality values within that document. 
\citet{cao-etal-2021-uncertain} propose an Uncertain Local-to-Global Network (ULGN) that makes use of two important characteristics of event factuality, \textit{local uncertainty} and \textit{global structure}. 
Similarly, \citet{qian-etal-2019-document} address the challenge of multiple event factuality values within a document by employing an LSTM model trained with both intra- and inter-sequence attention mechanisms to assess document-level event factuality. The model incorporates two types of input features, syntactic and semantic. Syntactic features are based on dependency paths from negative or
speculative references to the event, while semantic features are derived from the sentences containing the event.
Another approach to estimating document-level factuality is the Sentence-to-Document Inference Network (SDIN) proposed by \citet{zhang2023incorporating}. This architecture features a multilayer interaction network that aggregates individual event mentions into a global prediction. The last step employs \textit{gated aggregation} that uses a sigmoid function to generate a mask vector that captures the most critical semantic and factual features of the event mentions. The training process applies a multi-task learning approach, where individual and document-level event factuality prediction tasks share the same pretrained model and interaction network. As an input, the model receives all the event mentions encoded into BERT representations at the final hidden state of [CLS] token. This approach makes it possible to significantly outperform the models by \citet{cao-etal-2021-uncertain} and  \citet{zhang2023incorporating} described above on Chinese and English DLEF corpora \cite{qian-etal-2019-document}.
\citet{qian2022document} propose an approach called Document-level Event Factuality identification via Machine
Reading Comprehension Frameworks with Transfer Learning (DEFI-MRC-TL). The authors use BERT as a backbone model to train on a number of large-scale MRC
corpora and fine-tuned on the DEFI task. As a target dataset, Qian et al. construct their own MRC-style DEFI corpus called DLEFM by annotating both events and document-level event factuality. They achieve significantly higher results on their dataset than the models by Cao et al. \cite{cao-etal-2021-uncertain} and Zhang et al. \cite{zhang2023incorporating}.

\textbf{Subjectivity.} Subjectivity detection is a well-established NLP task that emerged long before the advent of deep learning and transformer models. 
As a result, the vast majority of the reviewed methods rely on traditional machine learning models, such as Multinomial
Naïve Bayes, Convolution Kernels, Support Vector Machines, Logistic Regression, Latent Dirichlet Allocation 
and Decision Trees over lexical, syntactic and semantic features \cite{BIYANI2014170, lin-etal-2011-sentence, cortis2022baseline, wiegand2011convolution, wang2011cross, banea-etal-2010-multilingual, atalla2011investigating, mukund2010vector, mourad2013subjectivity, mogadala2012language}. Rule- and lexicon-based approaches are also common for this task \cite{ANSAR2021100052, wiebe2004learning, hammer-etal-2014-sentiment, jeronimo2020computing, maks-vossen-2012-building}. The only transformer-based method identified through the systematic review is that by \citet{savinova2023analyzing}. The authors test the
efficiency of RoBERTa model in predicting subjectivity. They found a very high correlation with the human annotators and a significantly better performance than the existing rule-based and regression methods.

Some studies formulate the task of subjectivity detection as a part of the sentiment analysis task, where the ``neutral'' sentiment corresponds to the non-subjective class, and both ``positive'' and ``negative'' sentiments indicate subjective information \cite{CHATURVEDI201865}.  \citet{barbosa2010robust} explore traditional ML approached to perform the tweet sentiment prediction as a two-step approach. The authors first distinguish between subjective and non-subjective tweets, and then further perform ``positive'' and ``negative'' tweet prediction for subjective tweets. 

Finally, \citet{LU2024102203} address the task of sarcasm detection, which is often seen as a specialised type of subjective information. The authors propose a novel multimodal Fact-Sentiment Incongruity Combination Network (FSICN) approach that integrates the factual similarity and the sentiment information into the sarcasm detection task. The FSICN method incorporates a dynamic connection component that identifies the most relevant image-text pairs to detect fact incongruities between them.  

\textbf{Bias.} The methods used for the task of bias detection can be categorized based on the type of bias concerned. At a high level, the task can be approached as a binary classification of biased vs non-biased information. The methods falling into this group typically involve traditional machine learning models and lexicon-based methods \cite{SPINDE2021102505, AGGARWAL2020100025, 10.1109/JCDL.2019.00036, 8215582, 10.1145/3328526.3329582}. Over the last years, transformer models and LLMs became state-of-the-art approaches towards this task \cite{spinde-etal-2021-neural-media, lin-etal-2024-indivec, wessel2024beyond, liu-etal-2019-detecting, fan-etal-2019-plain, maab-etal-2023-effective, 10.1145/3529372.3530932, benson2024developing}.

More fine-grained bias analysis methods are predominantly focused on detecting the political leaning bias.  \citet{chen-etal-2020-analyzing} explore various levels of textual granularity (word, sentence, paragraph and discourse level) for political bias and unfairness detection. Their main method consists of a recursive neural network and uses GloVe embeddings \cite{pennington2014glove} as an input.  According to their analysis on the bias locations, last paragraph typically contains the most biased text segment, while all biased articles start with a neutral tone. \citet{baly-etal-2019-multi} model the task of political leaning bias detection jointly with the task of general trustworthiness detection. The authors use multi-task ordinal regression framework as the main model and consider 7 scales of political leaning. They found that each of the two task benefits from the joint training. \citet{sakketou2022factoid} look into the task of detecting user bias. They identify potential spreaders of biased information by applying Graph Attention Networks (GAT) over User2Vec embeddings. The latter is obtained through applying the SentenceBERT (SBERT) \cite{reimers-gurevych-2019-sentence} over the user's  history of posts. The majority of recent approaches for political bias detection are based on transfer learning using transformer models \cite{aksenov-etal-2021-fine, baly-etal-2020-detect, baly-etal-2020-written, baly-etal-2018-predicting, 10109099, 9377987}.

Another group of fine-grained bias detection methods concerns the idea of framing, where frames are seen as perspectives used to discuss the same topic. While some of the frames can represent political leaning, they can also be seen as discussion sub-topics. \citet{liu-etal-2019-detecting} analyse the dataset of news headlines on the topic of gun violence in the United States. Within this topic, they predict 9 discussion frames, such as mental health, politics, ethnicity, economical consequences, legal regulations, gun rights, public space safety and society. The authors found that BERT model significantly outperforms deep learning approaches, such as RNNs, LSTMs, Bi-LSTMs and Bi-GRU, for each of 9 frames.

Finally, \citet{10.1145/3539618.3591882} introduce the Media Bias Identification
Benchmark (MBIB) task of diverse bias detection, by unifying the existing datasets annotated with gender, political, cognitive, racial, textual and linguistic biases. They benchmark their dataset on  ConvBERT, Bart, RoBERTa-Twitter, ELECTRA, and GPT-2 models. The authors found that individual bias detection tasks benefit from different models. For example, ConvBERT  is better at predicting political, textual and cognitive bias, while ELECTRA model shows better average performance at racial and gender bias detection based on macro-average F1-scores.

\subsection{Tools and services}

The in-depth analysis of the research papers covered by this survey revealed several publicly available tools and services used by the NLP methods aimed at bias and factuality analysis. 

\textit{Media Bias/Fact Check (MBFC)} website\footnote{\url{https://mediabiasfactcheck.com/}} is a widely used service for media factuality (combining factual accuracy with other factors) and bias detection. It provides systematic human-centered annotations, spanning over 8 years and covering over 2,000 news websites. The annotators affiliated with International Fact-Checking Network (IFCN) assess factuality based on 4 factors: failed fact checks, sourcing, transparency, and one-sidedness/omission, each rated on a scale of 0 to 10. The final factuality score (also on a scale of 0 to 10) is a weighted average of these factors. This score is further mapped into a 6-level scale, indicating a ``very low'', ``low'', ``mixed'', ``mostly factual'', ``high'' and ``very high'' rankings. To detect bias, the annotators perform an aggregation of 4 factors: biased wording,
factual sourcing, story choice leaning (e.g., pro-liberal or pro-conservative), and political affiliation. Each of these categories is rated by experts on a scale of 0 to 10, from least to most biased. The final score is then calculated as a sum of the four respective scores and is mapped to the categories indicating a range of categories between extreme and moderate left and right bias and unbiased/centered reporting.
The original goal of the service was to educate the public on media factuality and bias and on deceptive news practices. In addition, the API makes the tool highly useful for NLP classification tasks~\cite{baly-etal-2018-predicting, baly-etal-2020-detect, 10.1007/978-3-031-42448-9_20, aires2020information, sales2021assessing, kangur2024checks}.

\textit{AllSides Media Bias Ratings}\footnote{\url{https://www.allsides.com/}} is a service similar to MBFC in terms of covering the political leaning bias based on a scale from right to left. The source coverage is also comparable to MBFC, with over 2,400 websited analysed. However, unlike MBFC, the judgment is not limited to a closed group of experts, but also includes ordinary people across the political spectrum trained to perform the analysis. This provides useful insights into how people with certain political leaning biases rate media news without knowing the source. Additionally, there is an option of a community feedback to agree or disagree with the provided rating. The labelled data from AllSides has been used in a number of classification tasks for political bias detection \cite{baly-etal-2020-detect, 9651872, li-goldwasser-2021-mean, 9651872, SPINDE2021102505, 10.1007/978-3-030-71305-8_17}.

\textit{Ad Fonted Media}\footnote{\url{https://adfontesmedia.com/}} provides an interactive media bias chart where media sources are annotated based on two different dimensions, \textit{reliability} and \textit{political bias}. The bias dimension consists of most extreme left/right, hyperpartisan left/right, strong left/right, skews left/right and middle or balanced bias. The reliability dimension consists of numerical scores that are mapped to 8 categories: fabricated news, misleading information, selective or incomplete story/propaganda, opinionated of highly varied reliability news, simple fact reporting, and high effort original fact reporting. This resource is also commonly used by NLP approaches to analyse media bias with various degrees of granularity \cite{chen-etal-2020-analyzing, jaradat2024detecting}. 

Finally, The University of Sheffield as a part of their \textit{GATE Cloud infrastructure} \cite{gate2002} provides public tools for multilingual news genre and framing detection based on its submission~\cite{wu-etal-2023-sheffieldveraai} to SemEval 2023 Shared Task 3~\cite{piskorski-etal-2023-semeval}. The models used in these tools achieved competitive results in the shared task, placing in top 3 for most of these languages. The genre detection tool\footnote{\url{https://cloud.gate.ac.uk/shopfront/displayItem/news-genre-classifier}} tackles the task of classifying the news into \textit{opinionated}, \textit{objective} and \textit{satire}. The framing detection tool\footnote{\url{https://cloud.gate.ac.uk/shopfront/displayItem/news-framing-classifier}} allows the analysis of the 9 principal frames used to present the information.

\subsection{Discussion}
As noted from the survey results, the tasks of factuality, subjectivity and bias detection cover various subtasks and task formulations. Among the subtasks, while subjectivity detection distinguishes between subjective and non-subjective content, certain formulations also perceive it as a sub-task of sentiment analysis. In turn, under certain formulations, sentiment detection tasks can be seen as a specific case of bias detection. As an example, distinction between neutral and positive/negative sentiment is a type of subjectivity detection task. In turn, positive and negative sentiment towards a certain topic can be seen as a biased representation of information.

\textbf{Lack of unified definitions and approaches.} One of the main challenges of this category of credibility signals is a high diversity in how researchers perceive and address factuality, subjectivity and bias, resulting in many definitions and ontologies. This problem strengthened by the 
inherent subjective nature of the subjectivity/bias is present already in dataset annotation causing also a low inter-annotator agreement~\cite{SPINDE2023100264}. Additionally, the problem of subjectivity detection is often seen as a sentiment detection \cite{mourad2013subjectivity} or a hate speech detection task \cite{LIN2022102872}. The problem of bias detection is sometimes seen as a specific case of fake news detection, where the biased presentation of facts is automatically seen as false information \cite{LIN2022102872}. The problem of factuality detection is sometimes seen as the degree of certainty of a speaker regarding the occurrence of events, e.g., \cite{9310484, cao-etal-2021-uncertain, rovera-etal-2025-modafact} and sometimes as factual accuracy (veracity)~\cite{nakov2021survey} or a mixture of veracity and other credibility signals (as is the case of the MBFC service). Finally, factuality, subjectivity and bias is a cumulative notion that is highly nuanced depending on how balanced the opinionated information is and how it is quoted and presented in text \cite{piskorski-etal-2023-semeval}.

\textbf{Data scarcity.} The biggest challenge in multilingual subjectivity and bias detection is data scarcity. Many English studies are US-centric and only few datasets are available in other languages. For this reason, the multilingual datasets are not well-representative for benchmarking models in global analyses and multicultural settings, since bias detection requires cultural context and background knowledge about a country's political spectrum. For instance, the concepts of right- and left-leaning media can be different across European or Middle East media.

\textbf{Focus on binary classification.} Despite the explanatory advantage of a fine-grained approach preferred by media professionals, the majority of the models are trained to perform the binary classification of texts into biased and non-biased. This can be, at least partially, a result of the scarcity of datasets available for a fine-grained classification (as illustrated in Table~\ref{tab:bias_data}). 

\textbf{Adoption of LLMs.} The use of large language models for detecting factuality, subjectivity, and bias is still in its early stages, largely due to two key limitations: the implicit biases embedded within LLMs themselves, and their tendency to hallucinate or generate factually incorrect information.

Recently, the task of subjectivity detection has been the focus of two large-scale studies examining it from the perspective of large language models. \citet{suwaileh2025thatiar} conducted a comprehensive benchmark of both mono- and multilingual models for subjectivity detection in Arabic sentences. Their findings show that LLMs outperform smaller multilingual fine-tuned models even in zero-shot settings. Performance further improves when LLMs are given few-shot examples. Similarly, \citet{shokri-etal-2024-subjectivity} evaluated transformer-based and large language models on subjectivity detection across three diverse datasets and domains. While fine-tuned models generally outperformed zero-shot LLMs, the latter still achieved performance close to that of the smaller fine-tuned models. The authors also explored various prompting strategies and chain-of-thought techniques, observing that in-context learning is not consistently robust, as its effectiveness is highly sensitive to the specific examples used in the prompt.

The task of bias detection has primarily been approached through the lens of framing analysis using LLMs. \citet{pastorino2024decoding} conducted the first benchmarking study of LLMs on framing detection, evaluating their performance in zero-shot, few-shot, and explainable prompting settings. The few-shot setting was further analysed using both in-domain and cross-domain examples. The authors found that prompting LLMs to generate explanations alongside predictions led to more robust and reliable results.
While some LLMs performed particularly well in few-shot settings, the models often perceive emotional style as framing, which poses a significant challenge for reliably applying LLMs to this task.  An additional challenge in applying LLMs to political bias detection is that LLMs themselves often exhibit inherent biases, which can shift in response to subtle changes in phrasing or context \cite{lunardi2024elusiveness, bang2024measuring}. Strategies such as prompt engineering and fine-tuning for debiasing \cite{lin2024investigating}, as well as increasing model transparency regarding bias-related behaviour \cite{mohanty2025fine}, may enhance the reliability of LLMs for future applications in this area.

\citet{mujahid2025profiling} explored the application of LLMs to assess political bias and factuality, where factuality is interpreted as factual accuracy (i.e., veracity) rather than as a contextual credibility signal. The authors propose two prompt-based approaches: a handcrafted method and a systematic profiling method. In the handcrafted approach, LLMs are presented with a series of questions designed to elicit judgments about a domain’s stance on specific political figures or topics, as well as assessments of the domain’s overall trustworthiness. The systematic approach involves using LLMs to profile editorial political bias across 16 policy areas. In both cases, the models were also prompted to provide justifications for their judgments. The outputs generated by the LLMs were then used as features to train traditional machine learning models and transformer-based models for predicting degrees of factuality (i.e., factual accuracy) and political bias. The study also compared these prompt-based methods to a zero-shot baseline, in which the LLM receives only the domain name, or a domain name plus a summary of five articles from the domain. Results show that the supervised prompt-based approaches significantly outperform the zero-shot baselines in both bias and factuality (i.e., factual accuracy) prediction tasks.

As mentioned above, factuality is also one of ubiquitous problems when evaluating the quality of LLMs. Research in this direction is related to how grounded LLMs' replies are to real-world facts rather than applying LLMs to estimate factuality of information. \citet{augenstein2024factuality} summarised major challenges of large language models in this directions, which include unreliable citations, hallucinations when grounding information, domain-specific unreliability, incoherent responses, factual inaccuracies and  difficulties performing deductive and inductive inferences. Additionally, factual accuracy of LLMs is affected by the knowledge gaps, which, depending on the information cut-off dates, can result in factually incorrect replies. These challenges become particularly dangerous when accompanied with confident tone and highly persuasive first-person replies. To address these challenges, the research community must consider a set of techniques, such as alignment of LLMs with certain values and safeguards, retrieval-based generation (RAG) to mitigate ungrounded replies, timely knowledge update and mitigation of hallucination. Finally, better evaluation metrics addressing the factuality (i.e., factual accuracy) of LLMs' outputs need to be introduced to ensure the quality of the generated replies. Certain benchmarks have already been introduced to perform the factuality assessment of LLMs and facilitate reliable and accurate replies \cite{iqbal2024openfactcheck, wang2024openfactcheck}.
\section{Persuasion techniques and logical fallacies}
\label{section:persuasion-and-fallacies}

Over the past decade, propagandistic efforts have been widely used on social media platforms to shape public opinion and drive engagement on a massive scale. To address this issue, several computational methods aimed at automatic detection of such a content have been proposed \cite{computational_propaganda_survey}. Within the scope of propagandistic content, persuasion techniques and logical fallacies aim to deliberately influence others' opinions using rhetorical and psychological mechanisms \cite{miller1939techniques}. Recently, the task of automatic detection of persuasion techniques has gained increased attention from the NLP research community, with several resources introduced over the last years.

\subsection{Datasets}
The majority of existing datasets are designed to support two related tasks: (i) classifying the input text as either containing persuasion techniques or not (binary classification), and (ii) extracting all specific persuasion techniques present in the input text (multi-label classification). While the binary task provides a simpler framework, it lacks the necessary granularity for detailed analysis of propagandistic content. The multi-label task is a more complex extension, requiring the model to identify individual persuasion techniques and thus enabling a more nuanced evaluation. Importantly, these tasks are implemented at different levels of granularity across datasets -- some annotate propaganda at the token-level (fine-grained), others at the sentence-level (coarse-grained), and some at the fragment-level (mid-grained), where continuous spans of text within a sentence are labelled. These distinctions significantly affect how models are trained and evaluated. For this reason, and due to its relevance for fine-grained analysis, we focus our analysis on the multi-label task. A summary of existing datasets and their key attributes, including annotation granularity, is presented in \autoref{tab:datasets_persuasion_techniques}.

\begin{table}[!tbp]
\footnotesize
\centering

\caption{Selected datasets used for detection of persuasion techniques.}
\label{tab:datasets_persuasion_techniques}

\resizebox{\columnwidth}{!}{
\begin{tabular}{|l|c|c|c|p{8cm}|}
\hline
\textbf{Dataset} &
  \textbf{Language(s)} &
  \textbf{\# Instances} &
  \textbf{Content type} &
  \textbf{Classes} \\ \hline
SemEval-2019 \cite{da-san-martino-etal-2019-findings} &
  English &
  7,485 &
  News articles &
  \textbf{18 persuasion techniques:} Appeal to authority, Appeal to fear/prejudice, Bandwagon, Black-and-white fallacy/dictatorship, Causal oversimplification, Doubt, Exaggeration or minimization, Flag-waving, Loaded language, Name calling or labelling, Obfuscation/intentional vagueness/confusion, Red herring, Reductio ad Hitlerum, Repetition, Slogans, Straw man, Thought-terminating cliché, Whataboutism \\ \hline
\citet{baisa-etal-2019-benchmark} &
  Czech &
  7,494 &
  News articles &
  \textbf{18 persuasion techniques:} Blaming, Labelling, Argumentation, Emotions, Demonizing, Relativizing, Fear mongering, Fabulation, Opinion, Location, Source, Russia, Expert, Attitude to a politician, Topic, Genre, Focus, Overall sentiment \\ \hline
\citet{lawson2020emailphising} &
  English &
  90 &
  Emails &
  \textbf{4 persuasion techniques:} Authority, Commitment/Consistency, Liking, Scarcity \\ \hline
PTC (SemEval-2020) \cite{da-san-martino-etal-2020-semeval} &
  English &
  8,981 &
  News articles &
  \textbf{14 persuasion techniques:} Appeal to authority, Appeal to fear/prejudice, Bandwagon/Reductio ad Hitlerum, Black-and-white fallacy, Causal oversimplification, Doubt, Exaggeration/minimization, Flag-waving, Loaded language, Name calling/labelling, Repetition, Slogans, Thought-terminating cliché, Whataboutism/Straw man/Red herring \\ \hline
SemEval-2021 \cite{dimitrov-etal-2021-semeval} &
  English &
  2,488 &
  Facebook posts &
  \textbf{22 persuasion techniques:} Appeal to authority, Appeal to (Strong) Emotions, Appeal to Fear or Prejudices, Bandwagon, Black-and-White Fallacy or Dictatorship, Causal Oversimplification, Flag-Waving Doubt, Exaggeration or Minimisation, Slogans, Glittering Generalities (Virtue), Loaded Language, Misrepresentation of Someone’s Position (Straw Man), Name Calling or Labelling, Repetition, Obfuscation/Intentional Vagueness/Confusion, Presenting Irrelevant Data (Red Herring), Reductio ad Hitlerum, Thought-Terminating Cliché, Smears, Transfer, Whataboutism \\ \hline
WANLP-2022 \cite{alam-etal-2022-overview} &
  Arabic &
  1,942 &
  Tweets &
  \textbf{20 persuasion techniques:} Appeal to authority, Appeal to fear/prejudices, Bandwagon, Black-and-white fallacy/dictatorship, Flag-waving, Doubt Causal oversimplification, Exaggeration/minimisation, Glittering generalities (virtue), Loaded language, Misrepresentation of someone’s position (straw man), Name calling or labelling, Repetition, Slogans Obfuscation/intentional vagueness/confusion, Presenting irrelevant data (red herring), Reductio ad Hitlerum, Smears, Thought-terminating cliché, Whataboutism \\ \hline
\citet{macagno2022argumentationprofiles} &
  \begin{tabular}[t]{@{}c@{}}English,\\ Italian,\\ Portuguese\end{tabular} &
  2,657 &
  Tweets &
  \textbf{9 persuasion techniques:} Straw man, False dichotomy, Ignoring qualifications, Question begging epithets, Post hoc ergo propter hoc, Hasty generalization, Slippery Slope, Persuasive definition, Quasi-definition \\ \hline
ArAIEval-2023 \cite{hasanain-etal-2023-araieval} &
  Arabic &
  5,919 &
  \begin{tabular}[t]{@{}c@{}}News articles\\ Tweets\end{tabular} &
  \textbf{23 persuasion techniques:} Appeal to Authority, Appeal to Fear/Prejudice, Appeal to Hypocrisy, Appeal to Popularity, Appeal to Time, Appeal to Values, Casting Doubt, Causal Oversimplification, Consequential Oversimplification, Conversation Killer, Exaggeration or Minimisation, False Dilemma or No Choice, Flag Waving, Guilt by Association, Loaded Language, Name Calling or Labelling, Red Herring, Repetition Obfuscation/intentional vagueness/confusion, Slogans Questioning the Reputation, Strawman, Whataboutism \\ \hline
SemEval-2023 \cite{piskorski-etal-2023-semeval} &
  \begin{tabular}[t]{@{}c@{}}English,\\ French,\\ German,\\ Georgian,\\ Greek,\\ Italian,\\ Polish,\\ Russian,\\ Spanish\end{tabular} &
  49,444 &
  News articles &
  \textbf{23 persuasion techniques:} Appeal to Authority, Appeal to Fear/Prejudice, Appeal to Hypocrisy, Appeal to Popularity, Appeal to Time, Appeal to Values, Casting Doubt, Causal Oversimplification, Consequential Oversimplification, Conversation Killer, Exaggeration or Minimisation, False Dilemma or No Choice, Flag Waving, Guilt by Association, Loaded Language, Name Calling or Labelling, Red Herring, Repetition Obfuscation/intentional vagueness/confusion, Questioning the Reputation, Slogans, Strawman, Whataboutism \\ \hline
\citet{almotairy2024arabpropaganda} &
  Arabic &
  2,100 &
  Tweets &
  \textbf{15 persuasion techniques:} Flag-waving, Smears, Name-calling, Loaded language, Exaggeration, Whataboutism, Glittering, Doubt, Causal oversimplification, Dictatorship, Appeal to fear, Slogan, Thought-terminating cliché, Appeal to authority, Reductio ad Hitlerum \\ \hline
\end{tabular}}
\end{table}

The NLP4IF-2019 shared task \cite{da-san-martino-etal-2019-findings} introduced a dataset for fine-grained propaganda detection consisting of news articles from $36$ propagandist and $12$ non-propagandist news outlets, and annotated with $18$ different propaganda techniques. Two subtasks were featured: sentence-level classification (binary task) and fragment-level classification (multi-label span identification task). 
Similarly, the SemEval-2020 shared task introduced the PTC-SemEval20 dataset \cite{da-san-martino-etal-2020-semeval}, which considered the same techniques used in the NLP4IF-2019 task, however, certain techniques were merged or removed due to their low frequency: \textit{Red Herring} and \textit{Straw man} were combined with \textit{Whataboutism}, and \textit{Bandwagon} was merged with \textit{Reductio ad Hitlerum}, while \textit{Obfuscation, Intentional vagueness, Confusion} was removed entirely. In SemEval-2021 \cite{dimitrov-etal-2021-semeval}, a dataset of multimodal persuasion techniques was introduced, containing Facebook posts with images and texts representing memes shared by users from 26 public groups. The visual modality provided additional context that was not present in the text modality alone. For instance, techniques such as \textit{Smears}, \textit{Doubt}, and \textit{Appeal to Fear/Prejudice} appeared more frequently when considering the image along with the text. 

Other datasets were introduced for non-English languages. \citet{baisa-etal-2019-benchmark} introduced a dataset for detection of persuasion techniques in the Czech language with around $7,000$ news articles. The techniques considered in their dataset differed substantially from others, with the presence of domain-specific techniques such as `Russia' (indicating that Russia was a topic discussed in the document). WANLP-2022 \cite{alam-etal-2022-overview}, ArAIEval-2023 \cite{hasanain-etal-2023-araieval}, and \citet{almotairy2024arabpropaganda} introduced datasets for persuasion detection in Arabic texts. The three Arabic datasets share several characteristics such as number of instances (between $2,000$ and $6,000$), content type (mainly tweets, with \citet{hasanain-etal-2023-araieval} additionally containing news articles), and label scheme (at least $10$ techniques are shared between the three datasets).

\citet{macagno2022argumentationprofiles} introduced the first multilingual dataset for detection of persuasion techniques, containing tweets in English, Italian, and Portuguese. However, the dataset is small ($<3,000$ instances in total) for the purpose of training deep learning models, and similarly to \cite{baisa-etal-2019-benchmark}, the set of persuasion techniques considered differs substantially from other datasets. SemEval-2023 \cite{piskorski-etal-2023-semeval} introduced a large-scale multilingual dataset covering $9$ languages (English, French, German, Georgian, Greek, Italian, Polish, Russian, and Spanish), with almost $50,000$ news articles labelled with $23$ different persuasion techniques, therefore being the largest dataset currently available in terms of number of instances, languages, and persuasion techniques. Three languages were considered ``surprise languages'' (Georgian, Greek, and Spanish), for which training sets were not available during the competition, thus encouraging systems to deal with out-of-domain data. Furthermore, the set of $23$ persuasion techniques were grouped into $6$ coarse-grained categories. For example, the techniques of \textit{Loaded Language}, \textit{Obfuscation, Intentional Vagueness, Confusion}, \textit{Exaggeration or Minimisation}, and \textit{Repetition}, were grouped into the umbrella of \textit{Manipulative Wording}.

\subsection{Methods and models}
The majority of methods and models developed for the task of automatic detection of persuasion techniques were introduced in the shared tasks discussed in the previous section. The joint effort of multiple different attempts at producing the best-scoring system allows to identify which methodological decisions are key to producing accurate models to detect persuasion. Table~\ref{tab:methods_persuasion} summarises the highest-scoring\footnote{Shared task systems that did not publish a description of their approach are not considered in our analysis.} systems across the shared tasks aimed at automatic detection of persuasion techniques.

\begin{table}[!htb]
\footnotesize
\caption{Top-scoring systems for automatic detection of persuasion techniques across different datasets.}
\label{tab:methods_persuasion}

\resizebox{\columnwidth}{!}{
\begin{tabular}{|l|l|l|l|l|}
\hline
\textbf{Dataset} &
  \textbf{System} &
  \textbf{Model} &
  \textbf{Approach} &
  \textbf{F1 Micro} \\ \hline
\multirow{2}{*}{SemEval-2019 \cite{da-san-martino-etal-2019-fine}} &
  newspeak \cite{yoosuf-yang-2019-fine} &
  BERT base uncased &
  \begin{tabular}[c]{@{}l@{}}- Token-level classification\\ with 20 classes: No PTs, one\\ of the 18 PTs, and an auxiliary\\ class to handle word-level tokenisation.\\ - Oversampling and class weighting.\end{tabular} &
  0.2488 \\ \cline{2-2} \cline{3-5} 
 &
  stalin \cite{ek-ghanimifard-2019-synthetic} &
  GROVER large \cite{grover_defending_neural_fakenews} &
  \begin{tabular}[c]{@{}l@{}}- Linear projection of contextual embeddings\\ - SMOTE oversampling\\ - BiLSTM classifier\end{tabular} &
  0.1453 \\ \hline
\multirow{2}{*}{SemEval-2020 \cite{da-san-martino-etal-2020-semeval}} &
  Hitachi \cite{morio-etal-2020-hitachi-semeval} &
  \begin{tabular}[c]{@{}l@{}}Ensemble \\ (BERT, GTP-2,\\ RoBERTa, XLM\\ XLM-RoBERTa, XLNet)\end{tabular} &
  \begin{tabular}[c]{@{}l@{}}- BIO encodings\\ - Contextual embeddings with POS tags and\\ named entities.\\ - Three training objectives: (i) BIO tag, \\ (ii) token-level, and (iii) sentence-level \\ classification.\\ - Two BiLSTMs, one for objectives  (i) and (ii),\\ and another for (iii).\\ - Class weighting\end{tabular} &
  0.5155 \\ \cline{2-2} \cline{3-5} 
 &
  ApplicaAI \cite{jurkiewicz-etal-2020-applicaai} &
  RoBERTa large &
  \begin{tabular}[c]{@{}l@{}}- Self-training using additional data\\ (500k sentences from OpenWebText).\\ - Added a continional random field (CRF) layer.\end{tabular} &
  0.4915 \\ \hline
\multirow{2}{*}{SemEval-2021 \cite{dimitrov-etal-2021-semeval}} &
  MinD \cite{tian-etal-2021-mind} &
  \begin{tabular}[c]{@{}l@{}} Ensemble \\ (BERT, RoBERTa,\\ XLNet, DeBERTa,\\ ALBERT)\end{tabular} &
  \begin{tabular}[c]{@{}l@{}}- Uses additional data from\\ \citet{da-san-martino-etal-2019-fine}.\\ - Model ensemble\\ - Custom rules for the\\ Repetition technique. \\ - Character-level n-grams \end{tabular} &
  0.593 \\ \cline{2-2} \cline{3-5} 
 &
  Volta \cite{gupta-sharma-2021-nlpiitr} &
  RoBERTa &
  - Used backtranslation as data augmentation &
  0.57 \\ \hline
\multirow{2}{*}{WANLP-2022 \cite{alam-etal-2022-overview}} &
  NGU CNLP \cite{hussein-etal-2022-ngu} &
  AraBERT &
  \begin{tabular}[c]{@{}l@{}}- Translated the PTC dataset \cite{da-san-martino-etal-2019-fine} to Arabic\\ and used as additional training data\\ - Stacking-based model ensemble\end{tabular} &
  0.649 \\ \cline{2-2} \cline{3-5} 
 &
  IITD \cite{mittal-nakov-2022-iitd} &
  XLM-RoBERTa large &
  \begin{tabular}[c]{@{}l@{}}- Simply fine-tuned the model using\\ the task dataset\end{tabular} &
  0.609 \\ \hline
\multirow{2}{*}{ArAIEval-2022 \cite{hasanain-etal-2023-araieval}} &
  UL \& UM6P \cite{lamsiyah-etal-2023-ul-um6p} &
  AraBERT-Twitter-v2 &
  \begin{tabular}[c]{@{}l@{}}- Used an asymmetric multi-label\\ loss objective \cite{assymetric_loss}\end{tabular} &
  0.5666 \\ \cline{2-2} \cline{3-5} 
 &
  rematchka \cite{abdel-salam-2023-rematchka} &
  AraBERT-v2 &
  \begin{tabular}[c]{@{}l@{}}- Class weighting\\ - Balanced data sampler\end{tabular} &
  0.5658 \\ \hline
\multirow{3}{*}{SemEval-2023 \cite{piskorski-etal-2023-semeval}} &
  \citet{sheffield_comparison_peft} &
  XLM-RoBERTa large &
  \begin{tabular}[c]{@{}l@{}}- Multilingual joint fine-tuning. \\ - LoRA \\ - Class weighting \end{tabular} &
  0.429 \\ \cline{2-2} \cline{3-5} 
 &
  KInITVeraAI \cite{hromadka-etal-2023-kinitveraai} &
  XLM-RoBERTa large &
  \begin{tabular}[c]{@{}l@{}}- Multilingual joint fine-tuning.\\ - Carefully chosen classification threshold for\\ each language (around 0.2).\end{tabular} &
  0.42 \\ \cline{2-2} \cline{3-5} 
 &
  Ampa \cite{pauli-etal-2023-teamampa} &
  XLM-RoBERTa large &
  \begin{tabular}[c]{@{}l@{}}- Oversampling\\ - Ensemble with models trained with one\\ and multiple languages.\end{tabular} &
  0.395 \\ \hline
\end{tabular}}
\end{table}

% Models
Transformer-based models were employed in all top-scoring systems for automatic detection of persuasion techniques due to their ability to capture nuanced contextual information \cite{vaswani-transformer}. In the NLP4AI-2019 shared task \cite{da-san-martino-etal-2019-findings}, $5$ out of the $6$ submissions included transformer-based models, while more recently in SemEval-2023 \cite{piskorski-etal-2023-semeval}, all $16$ submissions were comprised of transformer-based models. In earlier shared tasks, BERT was initially preferred over other architectures such as RoBERTa, ALBERT, and DeBERTa, which were adopted more frequently later, specially RoBERTa. In SemEval-2019 BERT was used in all transformer-based submissions. In SemEval-2020 and SemEval-2021, BERT was used in a total of $25$ systems, while RoBERTa is used in $14$ systems. Nevertheless, systems using RoBERTa achieved the top $2$ submissions in both shared tasks \cite{morio-etal-2020-hitachi-semeval, jurkiewicz-etal-2020-applicaai, tian-etal-2021-mind, gupta-etal-2021-volta}. For tasks in Arabic (WANLP-2022 and ArAIEval-2022), monolingual models (AraBERT and variations) outperformed multilingual models (e.g., XLM-RoBERTa and mBERT) \cite{hasanain-etal-2023-araieval}. In SemEval-2023 where $9$ languages were available, multilingual models (e.g., mBERT and XLM-RoBERTa) outperformed monolingual models trained with each language separately \cite{wu-etal-2023-sheffieldveraai}.
% This is likely due to the fact that English is generally a more resourceful language in the context of pretrained language models (e.g., the share of English data used to pretrain multilingual models is larger than other languages) (cite).
Unsurprisingly, the larger variations of the models generally outperformed the smaller versions. For reference, RoBERTa large has almost triple the size of RoBERTa base ($355$ and $125$ million parameters, respectively). Also, model ensembles combining different pretrained models were widely and effectively employed, although at the cost of requiring more computational resources \cite{morio-etal-2020-hitachi-semeval, tian-etal-2021-mind, gupta-etal-2021-volta, hossain-etal-2021-csecu-dsg, chernyavskiy-etal-2020-aschern}.

In addition to the contextual embeddings generated by transformer-based models, some works have experimented with supplementary input features such as part-of-speech (POS) tags \cite{morio-etal-2020-hitachi-semeval, khosla-etal-2020-ltiatcmu, roele-2021-wvoq}, named entity recognition encodings \cite{morio-etal-2020-hitachi-semeval, khosla-etal-2020-ltiatcmu}, word-level n-grams \cite{khosla-etal-2020-ltiatcmu, kim-bethard-2020-ttui, ghadery-etal-2021-liir}, character-level n-grams \cite{tian-etal-2021-mind, chernyavskiy-etal-2020-aschern}, and sentiment scores \cite{khosla-etal-2020-ltiatcmu, roele-2021-wvoq}. However, apart from the contextual embeddings, it is not clear if other input representation methods play a significant role in improving performance. In fact, apart from \citet{morio-etal-2020-hitachi-semeval} who used named entity recognition encodings, all other 1st and 2nd placing systems in shared tasks have not used other supplementary features apart from the contextual embeddings.

Arguably the most relevant methods employed in top-performing systems are aimed towards dealing with data skewness. Most datasets for this task suffer from class imbalance, meaning the majority of persuasion techniques are underrepresented, while a small set of persuasion techniques comprise the majority of the dataset. To deal with this issue, several different methods are employed, such as oversampling underrepresented techniques \cite{yoosuf-yang-2019-fine, ek-ghanimifard-2019-synthetic, pauli-etal-2023-teamampa}, scaling the contribution of the techniques to the loss function according to their proportion (i.e., class weighting) \cite{yoosuf-yang-2019-fine, morio-etal-2020-hitachi-semeval, abdel-salam-2023-rematchka, sheffield_comparison_peft}, and using supplementary data obtained with (i) data augmentation \cite{gupta-sharma-2021-nlpiitr}, (ii) semi-supervised methods \cite{jurkiewicz-etal-2020-applicaai}, or (iii) similar datasets \cite{tian-etal-2021-mind, abujaber-etal-2021-lecun, hussein-etal-2022-ngu, mittal-nakov-2022-iitd}.

\subsection{Tools and services}
In terms of production grade tools to detect persuasion techniques in texts, to the best of our knowledge, two services are currently available.

The \textit{Propaganda Persuasion Techniques Analyzer (PRTA)} \cite{da-san-martino-etal-2020-prta} is a discontinued tool that is no longer available online. It was designed to detect instances of propaganda in texts by highlighting the spans where specific techniques were used. It provided users with the ability to compare texts based on their use of propaganda techniques, offering detailed statistics on the prevalence of these techniques, both overall and over time. PRTA used a BERT model trained on the SemEval-2019 dataset, fine-tuned for both fragment-level and sentence-level classification tasks. To gather data, PRTA crawled a growing list of over $250$ RSS feeds, Twitter accounts, and websites, extracting text via the Newspaper3k library and performing deduplication using a hash function. The system then identified sentences containing propaganda, organised the articles into topics (such as COVID-19 or Brexit), and allowed users to compare the use of propaganda techniques across various media sources. The tool also allowed for filtering based on time intervals, keywords, or political orientation of the media. Although the user interface of the PRTA tool is no longer available, the underlying API for propaganda detection (along with other credibility signals) remains accessible\footnote{\url{https://apihub.tanbih.org/propaganda/}}.

The \textit{GATE Cloud} infrastructure \cite{gate2002} operated by the University of Sheffield  provides a service to detect $23$ different persuasion techniques across multiple languages using a multilingual BERT (mBERT) model\footnote{\url{https://cloud.gate.ac.uk/shopfront/displayItem/persuasion-classifier}}. The model was fine-tuned using data from SemEval-2023 in English, French, German, Italian, Polish, and Russian. Class weighting was applied during fine-tuning to account for the label skewness issue. The tool is capable of processing up to $1,200$ documents per day free of charge through its API, with an average processing rate of 2 documents per second. Researchers can request higher quotas if needed for larger-scale analysis.

\subsection{Discussion}

\textbf{Most commonly adopted persuasion techniques}. Table~\ref{tab:datasets_persuasion_techniques} shows that $7$ out of the $10$ datasets present a similar set of persuasion techniques \cite{da-san-martino-etal-2019-fine, da-san-martino-etal-2020-semeval, dimitrov-etal-2021-semeval, alam-etal-2022-overview, hasanain-etal-2023-araieval, piskorski-etal-2023-semeval, almotairy2024arabpropaganda}, while \cite{baisa-etal-2019-benchmark,lawson2020emailphising,macagno2022argumentationprofiles} largely diverge from them. Considering these seven similar datasets, we observe a set of $12$ persuasion techniques that are consistent among them: \textit{Appeal to authority}, \textit{Appeal to fear/prejudice}, \textit{Causal oversimplification}, \textit{Doubt}, \textit{Exaggeration or minimization}, \textit{Flag-waving}, \textit{Loaded language}, \textit{Name calling or labelling}, \textit{Slogans}, and \textit{Whataboutism} appear in all seven datasets, while \textit{Repetition} appears in six datasets (except \cite{almotairy2024arabpropaganda}), and \textit{Reductio ad Hitlerum} appears in five datasets (except \cite{hasanain-etal-2023-araieval, piskorski-etal-2023-semeval}). Combining existing datasets with overlapping persuasion techniques could be a promising research direction to develop more robust models and benchmarking resources. For example, \citet{tian-etal-2021-mind} achieved 1st place in SemEval-2021 Task 6 by leveraging the dataset from \cite{da-san-martino-etal-2019-fine} as additional training data.

\textbf{Label skewness.} A common characteristic inherent to this label scheme is the significant skewness in distribution of the persuasion techniques. For example, in SemEval-2023 \cite{piskorski-etal-2023-semeval} (the largest multilingual dataset, with $50,000$ instances), a small set of $6$ techniques represent $71.8\%$ of the entire dataset - \textit{Loaded
Language} ($18.5\%$), \textit{Name Calling-Labelling} ($23.7\%$), \textit{Casting Doubt} ($12.5\%$), \textit{Questioning the Reputation} ($7.6\%$), \textit{Appeal to Fear-Prejudice} ($4.8\%$), and \textit{Exageration/Minimisation} ($4.7\%$) - while the remaining $17$ techniques account for only $28.2\%$ of the dataset. The same trend can be seen in \cite{da-san-martino-etal-2019-fine}, with $6$ techniques representing $68.2\%$ of the dataset - \textit{Loaded language} ($34\%$), \textit{Name calling, labelling} ($17.3\%$), \textit{Repetition} ($10.2\%$), \textit{Exaggeration, minimization} ($7.6\%$), \textit{Doubt} ($7.5\%$), and \textit{Appeal to fear/prejudice} ($4.9\%$) - while the remaining $12$ techniques account for $31.2\%$ of the dataset. This sparse label distribution poses several challenges for training and evaluation. In particular, the overrepresentation of a few dominant techniques may lead models to bias their predictions toward these frequent categories, while underrepresented techniques suffer from insufficient training examples. This data imbalance increases the risk of overfitting to the few available samples of rare classes, potentially resulting in poor generalization.

\textbf{Evaluation.} Shared tasks on persuasion techniques detection typically employ the F1-Micro measure as the official evaluation metric \cite{dimitrov-etal-2021-semeval, da-san-martino-etal-2020-semeval, alam-etal-2022-overview, hasanain-etal-2023-araieval, piskorski-etal-2023-semeval}, however, the F1-Micro does not account for skewed class distributions, as opposed to F1-Macro\footnote{Generally F1-Macro is also reported as a secondary metric, but for the purpose of the competition, the F1-Micro determines which system is best.}. This evaluation strategy encourages models that excel at predicting a few densely distributed persuasion techniques. For example, in ArAIEval-2023 \cite{hasanain-etal-2023-araieval}, the baseline model (majority vote classifier that always predicts the most common persuasion technique), achieves an F1-Micro of $0.3599$, and an F1-Macro of $0.0279$, representing a difference of $92\%$ between the two scores. Similarly, the best submission by \citet{lamsiyah-etal-2023-ul-um6p} achieves F1-Micro and F1-Macro scores of $0.5666$ and $0.2156$, a difference of $62\%$. Similar figures are seen in other shared tasks such as SemEval-2023 \cite{piskorski-etal-2023-semeval}, with the average F1-Micro score for first place submissions across all languages resulting in $0.454$, and the F1-Macro in $0.211$, a difference of $53\%$. These figures highlight how current state-of-the-art models trained for this task still lack the ability to predict underrepresented persuasion techniques effectively.

\textbf{Multilinguality.} Despite efforts to create multilingual datasets such as SemEval-2023 \cite{piskorski-etal-2023-semeval} and \citet{macagno2022argumentationprofiles}, most datasets focus on English or Arabic. For languages without annotated datasets, machine-translation and zero-shot classification approaches may be viable alternatives (if some level of noise introduced by machine translation is acceptable). Nevertheless, joint multilingual training often result in more accurate models. For instance, \citet{sheffield_comparison_peft} experimented training monolingual models with translated data from SemEval-2023, which improved performance only for English, but not for other languages. Therefore, multilingual datasets with wider variety of languages are still required to produce state-of-the-art models and evaluation benchmarks.

\textbf{Adoption of LLMs.}
Recent research has increasingly explored the capabilities of large language models in generating persuasive and deceptive content, particularly in the context of disinformation and propaganda \cite{persuasive_simchon_2024, persuasive_llm_2024, rogiers_persuasion_2024, xu_earth_2024, pauli_measuring_2024}. LLMs can manipulate existing factual content into misleading or false narratives at scale, often outperforming human-written disinformation in believability and fluency \cite{matz2024potential, chen2023can} and can even personalize the content to specific target groups~\cite{zugecova2024evaluationllmvulnerabilitiesmisused}. However, while the generation side has been widely studied, significantly less attention has been paid to using LLMs for detecting or countering persuasive techniques. Bridging this gap by harnessing LLMs for persuasion detection presents a promising but underexplored research direction.
\section{Check-worthy and fact-checked claims}
\label{section:claims-and-veracity}

Given the impossibility to verify the veracity of every single piece of information posted online, two credibility signals and corresponding tasks are particularly crucial in supporting fact-checkers and other actors involved in the fight against disinformation and misinformation \cite{guo2022survey}: \textit{check-worthiness detection} and the \textit{retrieval of previously fact-checked claims}. The former is aimed at identifying which claims are check-worthy because of their relevance and interest to the general public \cite{10.1145/2806416.2806652}. The second, which ideally takes place after the first one, is meant to ease claim verification by checking whether the given claim was already verified before by searching in existing repositories of fact-checked claims~\cite{shaar-etal-2020-known}, both monolingual and multilingual. 

While these tasks are usually studied in separation as a part of the fact-checking pipelines, their outputs can be and also are used (usually in combination) as credibility signals to indicate whether the check-worthy (central) claims contained in a piece of content have been fact-checked and if so, what was the given veracity value or values. In fact, they are recognized as such both in the credibility signals list created by the \textit{Credible Web Community Group} (signals `article has a central claim' and `fact-check status of a claim'; see ~\cite{multipleauthorsCredibilitySignalsUnofficial}) as well as in the related works~\cite{10.1145/3184558.3188731}. The check-worthiness of a claim can be considered a content-based signal, while fact-check status of a claim is a context-based one, since it requires additional external sources to be detected. On the other hand, fact-checking itself is not considered a credibility signal per se -- its aim is to directly ascertain the veracity value of a piece of information, while the former are only signals that the information might be check-worthy (and thus more attention is needed before the users make credibility judgment) or that it contains a claim that have already been fact-checked (but it does not by itself analyse the stance towards that claim, only its presence).

It is also worth noting that these signals differ from factuality signal as it is understood in the surveyed works as discussed in Section~\ref{section:factuality-subjectivity-bias}).

\subsection{Datasets}

Check-worthy claim detection is a popular task within the NLP community thanks to the series of CheckThat! shared tasks organised at CLEF, which led to the release of related datasets. Indeed, check-worthy claim detection is the only subtask that has been proposed at all seven CheckThat! editions \cite{DBLP:conf/clef/AtanasovaMBESZK18,DBLP:conf/clef/AtanasovaNKMM19,DBLP:conf/clef/Barron-CedenoEN20,DBLP:conf/clef/NakovMEBMSAHHMH21,DBLP:conf/clef/NakovBMAMCKZLSM22,DBLP:conf/clef/AlamBCSHHLMMZN23,10.1007/978-3-031-56069-9_62}. Through those editions, datasets for training check-worthy claim detection models have been constantly extended to cover additional languages, starting from English and Arabic at CheckThat! 2018 \cite{DBLP:conf/clef/AtanasovaMBESZK18} to Arabic, Dutch, English and Spanish at CheckThat! 2024 \cite{10.1007/978-3-031-56069-9_62} and even multimodal and multi-genre data in the 2023 edition \cite{DBLP:conf/clef/AlamBCSHHLMMZN23}.  Beside CheckThat!, additional datasets have been created and made available for research, as shown in Table \ref{tab:checkworth}. The  sources covered by such datasets are typically social media, news and political debates, i.e. three relevant areas where the presence of disinformation may have detrimental effects on a large audience. Particular relevance was given to COVID-related content during the pandemic \cite{HADJAMEUR2021232,10.1145/3543873.3587643}, when the continuous flow of misleading information could negatively affect public health. 
Although English is the most represented language, datasets for check-worthiness detection have also been developed in several other languages. This can be attributed to the fact that public relevance is both time- and geographically bounded, with claims typically referring to events specific to individual countries or regions.

\begin{table}[!t]
\caption{Selected datasets used for check-worthiness detection. Note that the datasets used in the different CLEF CheckThat! editions often overlap.}
\label{tab:check_worthiness_datasets}
\footnotesize

\resizebox{\columnwidth}{!}{
\begin{tabular}{|l|c|c|c|c|c|}
\hline
    \textbf{Dataset} & \textbf{Language(s)} & \textbf{\# Instances} & \textbf{Content type} & \textbf{Classes} \\ \hline
      TR-Claim19 \cite{kartal-kutlu-2020-trclaim} & Turkish & 2,287 & Tweets &   
   \begin{tabular}{@{}c@{}}Check-worthiness \\ 26 rationale categories\end{tabular}\\
   \hline
   
   CW-USPD-2016 \cite{gencheva-etal-2017-context} & English & 5,415 & Political debates & 
   Check-worthiness \\ 
   \hline

   Dhar and Das \cite{dhar-das-2021-leveraging}& Bengali, Hindi & 2,402 & Political news, Twitter &   
   Check-worthiness \\ 
 \hline

    MM-Claims \cite{cheema-etal-2022-mm}& English & 3,400 & tweets (image + text)  &   
   \begin{tabular}{@{}c@{}}Claim detection, \\check-worthiness,\\ visual relevance\end{tabular} \\
\hline

    Sheikhi et al. \cite{sheikhi-etal-2023-automated}& Norwegian & 4,885 & News  &   
   Check-worthiness \\
\hline 

      AraCOVID19-MFH \cite{HADJAMEUR2021232} & Arabic & 10,828 & COVID-related tweets & 
      \begin{tabular}{@{}c@{}}Check-worthiness, \\factual, hateful \end{tabular}\\

 \hline

      Faramarzi et al.  \cite{10.1145/3543873.3587643} & English & 7,017 & COVID-related tweets & 
      \begin{tabular}{@{}c@{}}Check-worthiness, \\claim extraction \end{tabular}\\ 
 \hline
      ClaimBuster dataset  \cite{DBLP:conf/icwsm/ArslanHLT20} & English & 22,281 & Presidential debates & 
      \begin{tabular}{@{}c@{}}Check-worthiness, \\factual \end{tabular}\\

 \hline

 \begin{tabular}{@{}l@{}}CLEF-2018 CheckThat! \\Lab Task 1 \cite{DBLP:conf/clef/AtanasovaMBESZK18}\end{tabular}& English, Arabic & 17,300 & Political debates  &   
   Check-worthiness \\
\hline
 \begin{tabular}{@{}l@{}}CLEF-2019 CheckThat! \\Lab Task 1 \cite{DBLP:conf/clef/AtanasovaNKMM19}\end{tabular}& English & 24,000 & Debates, speeches, press conferences  &   
   Check-worthiness \\

\hline
 \begin{tabular}{@{}l@{}}CLEF-2020 CheckThat! \\Lab Task 1 \cite{DBLP:conf/clef/Barron-CedenoEN20}\end{tabular}& English, Arabic & 962 (EN) 7,500 (A),  & Tweets, political debates, speeches  &   
   Check-worthiness \\

   \hline
 \begin{tabular}{@{}l@{}}CLEF-2021 CheckThat! \\Lab Task 1 \cite{DBLP:conf/clef/NakovMEBMSAHHMH21}\end{tabular}& \begin{tabular}{@{}c@{}}Arabic, Bulgarian, English, \\Spanish, Turkish \end{tabular}& \begin{tabular}{@{}c@{}}18,014 (tweets), \\50,123 (sentences) \end{tabular}& Debates, speeches, tweets  &   
   Check-worthiness \\

   \hline
 \begin{tabular}{@{}l@{}}CLEF-2022 CheckThat! \\Lab Task 1 \cite{DBLP:conf/clef/NakovBMAMCKZLSM22}\end{tabular}& \begin{tabular}{@{}c@{}}Arabic, Bulgarian, Dutch, \\English, Spanish, Turkish\end{tabular}& 30,363& Tweets  &   
   \begin{tabular}{@{}c@{}}Check-worthiness, \\verifiable, harmful \end{tabular}\\

   \hline
 \begin{tabular}{@{}l@{}}CLEF-2023 CheckThat! \\Lab Task 1 \cite{DBLP:conf/clef/AlamBCSHHLMMZN23}\end{tabular}& Arabic, English, Spanish & 70,806 & Tweets, political debates, speeches  &   
    \begin{tabular}{@{}c@{}}Multimodal and multigenre \\check-worthiness\end{tabular} \\
 \hline
 
  \begin{tabular}{@{}l@{}}CLEF-2024 CheckThat! \\Lab Task 1 \cite{10.1007/978-3-031-56069-9_62}\end{tabular}& \begin{tabular}{@{}c@{}}Arabic, Dutch, \\English, Spanish\end{tabular} & 64,700 & Tweets, political debates, speeches   &   
   Check-worthiness \\

 \hline
\end{tabular}}
\label{tab:checkworth}
\end{table}

Concerning the retrieval of previously fact-checked claims, there are several datasets containing verified (fact-checked) claims collected from professional fact-checking organizations, such as X-Fact~\cite{gupta-srikumar-2021-x}, MultiFC~\cite{augenstein-etal-2019-multifc} or ClaimsKG~\cite{claims-kg-2019}. Being usually collected for the task of automatic fact-checking, these are not directly usable for the task of retrieval of previously fact-checked claims by themselves, as they lack the input claims (e.g., social media posts) and the pairs of the input and the verified claims. The first datasets designed specifically for the task appeared in 2020 in the works by Shaar et al.~\cite{shaar-etal-2020-known} and Vo and Lee~\cite{vo-lee-2020-facts} who independently prepared datasets based on both Snopes and PolitiFact. The former became the basis for a series of CheckThat! Lab shared tasks (Task 2) organised at CLEF in 2020~\cite{DBLP:conf/clef/Barron-CedenoEN20}, 2021~\cite{shaar_overview_2021} and most recently 2022~\cite{nakov_overview_2022}. The tasks gradually expanded the original dataset in size and also by adding an additional language (Arabic). The political debates part of the dataset was additionally expanded in~\cite{shaar-etal-2022-assisting}.

The datasets for retrieval of previously fact-checked claims are usually collected by using one of the following approaches: (i) looking into the fact-checking articles for links to the original content making the claim that is being verified, e.g.,~\cite{shaar-etal-2020-known, DBLP:conf/clef/Barron-CedenoEN20, shaar_overview_2021, nakov_overview_2022, pikuliak-etal-2023-multilingual, singh2024-mmtweets} or (ii) searching for (social media) content (such as discussion threads) that contains URL of or semantic links to the fact-checking articles, e.g.~\cite{vo-lee-2020-facts, hardalov-etal-2022-crowdchecked, nielsen-mcconville-2022-mumin}. The former typically achieves high precision at the cost of lower number of pairs and possible missing connections (i.e., many false negatives), while the latter can maximise recall at the cost of introducing noise unless manual checking is applied. Typical example of a high noise are CrowdChecked~\cite{hardalov-etal-2022-crowdchecked} or MuMiN~\cite{nielsen-mcconville-2022-mumin} which both contain a large number of social media posts which distinguishes them from other datasets.

While many available datasets focus on English only, there are several newer ones supporting other languages (e.g., Arabic~\cite{shaar_overview_2021, nakov_overview_2022} or Spanish~\cite{martin_2022_facter_check}) or even a range of languages~\cite{kazemi-etal-2021-claim, pikuliak-etal-2023-multilingual, nielsen-mcconville-2022-mumin, singh2024-mmtweets}. The most notable among these are MultiClaim~\cite{pikuliak-etal-2023-multilingual} and MMTweets~\cite{singh2024-mmtweets} due to the number of included languages, high precision of identified pairs and their amount as well as due to the fact that they both introduced a task of crosslingual retrieval, in which input claims are in different language than that of the verified claims.

A full list of selected relevant datasets is reported in Table \ref{tab:prev_fc_retrieval_datasets}. Besides these, it is also worth mentioning two datasets focusing on COVID-related claims, namely CoAID~\cite{cui2020-coaid} and MM-COVID~\cite{li2020mmcovid}; however, these focus on a different task of fake news/disinformation detection. A dataset of COVID-related tweets and fact-checks is also presented in~\cite{jiang-etal-2023-categorising, haouari-etal-2021-arcov19}. In this case, although the dataset was introduced for a different task, it could also be useful for previously fact-checked claim retrieval.

\begin{table}[t]
\caption{Selected datasets used for retrieval of previously fact-checked claims.}
\label{tab:prev_fc_retrieval_datasets}
\footnotesize

\resizebox{\columnwidth}{!}{
\begin{tabular}{|p{3.7cm}|p{3cm}|c|c|c|p{3.3cm}|}
\hline
    \multirow{2}{3.7cm}{\textbf{Dataset}} & \multirow{2}{3cm}{\textbf{Language(s)}} & \multicolumn{3}{|c|}{\textbf{\# Instances}} & \multirow{2}{3.3cm}{\textbf{Content type}} \\ \cline{3-5}
    & & \textbf{\# Input claims} & \textbf{\# Verified claims} & \textbf{\# Pairs} &  \\ \hline
    
    That is a Known Lie -- Snopes~\cite{shaar-etal-2020-known} & English & 1,000 & 10,396 & 1,000 & social media posts (Twitter) \\ \hline
    That is a Known Lie -- PolitiFact~\cite{shaar-etal-2020-known} & English & ~768 & 16,636 & 768 & political debates \\ \hline
    Snopes (\citet{vo-lee-2020-facts}) & English &  11,167 & 1,703 & 11,202 & social media posts (Twitter) \\ \hline
    PolitiFact (\citet{vo-lee-2020-facts}) & English & 2,026 & 467 & 2,037 & social media posts (Twitter) \\ \hline
    \citet{kazemi-etal-2021-claim} & English, Hindi, Bengali, Malayalam, Tamil & NA & NA &  2,343 & instant messages (WhatsApp) \\ \hline
    CLEF-2022 CheckThat! Lab Task 2A~\cite{nakov_overview_2022} & English, Arabic & 2,518 & 44,214 & 2,699 & social media posts (Twitter) \\ \hline
    CLEF-2022 CheckThat! Lab Task 2B~\cite{nakov_overview_2022} & English & 752 & 20,771 & 869 & political debates and speeches \\ \hline
    CrowdChecked~\cite{hardalov-etal-2022-crowdchecked} & English & 316,564 & 10,340 & 332,660 & social media posts (Twitter) \\ \hline
    MuMiN~\cite{nielsen-mcconville-2022-mumin} & 41 languages & 21,565,018 & 12,914 & NA & social media posts (Twitter) \\ \hline
    NLI19-SP (FacTeR-Check)~\cite{martin_2022_facter_check} & Spanish & ~40,000 & 61 & NA & social media posts (Twitter) \\ \hline
    MultiClaim~\cite{pikuliak-etal-2023-multilingual} & posts in 27 languages, fact-checks in 39 languages &  28,092 & 205,751 & 31,305 & social media posts (Twitter, Facebook, Instagram) \\ \hline
    MMTweets~\cite{singh2024-mmtweets} & posts in 4 languages, fact-checks in 11 languages &  1,600 & 30,452 & 4,258 & social media posts (Twitter) \\ \hline
\end{tabular}}
\end{table}

\subsection{Methods and models}

To propose methods for check-worthy claim detection, existing works first define how \textit{check-worthiness} is operationalized. In a recent survey, \citet{PANCHENDRARAJAN2024100066} identify two main aspects making a claim check-worthy: its \textit{verifiability} and its \textit{priority}. The first item refers to the possibility of determining the veracity of a claim, which can be likely supported by evidence. A claim is verifiable when it contains a factual statement that can be checked, which means that personal opinions or events presented as uncertain are excluded. Second, given that verifying all statements about the world is impossible, it is important to prioritize claims which are considered timely, interesting to the general public and whose verification might have a broader impact \cite{DAS2023103219}. In this latter aspect it differs from a related, but a distinct task of \textit{claim detection} which solely aims to identify what constitutes a claim in a text (either using a binary classification or by identification of spans)~\cite{mittal-etal-2023-lost, wuehrl-etal-2023-entity, gupta-etal-2021-lesa}.
%Check-worthy claim detection has been modelled as the first step of a potential fact-checking pipeline \cite{guo2022survey}. 

Given this need for prioritization, check-worthiness detection has been cast as a ranking problem since the first editions of the CheckThat! Lab within the CLEF Evaluation initiative\footnote{\url{https://www.clef-initiative.eu/}}, which made the task well-known within the NLP community, although some related works had already been presented before  \cite{10.1145/2806416.2806652}. In particular, given a political debate, the first CheckThat! task was aimed  at predicting which claims should be prioritized for fact checking. This is reflected in the evaluation approach proposed for the CheckThat! series, which has become the \textit{de facto} standard in check-worthiness detection: systems should output the list of input claims ranked by check-worthiness, which is usually evaluated using \textit{Mean Average Precision (MAP)}, reciprocal rank and $P@k$ for $k$ $\in$ $\{1, 3, 5, 10, 20, 30\}$. More recently, check-worthiness has been, however, evaluated using F1 as a binary classification task rather than ranking \cite{DBLP:conf/clef/AlamBCSHHLMMZN23,10.1007/978-3-031-56069-9_62}.

Regarding methods, in the CheckThat! editions up to 2023 state-of-the-art results for check-worthy claim detection were reached by methods that rely on fine-tuned transformer-based methods such as BERT, RoBERTa, DistilBERT, \cite{DBLP:conf/clef/Williams0T21,frick2023fraunhofer,DBLP:conf/clef/SawinskiWKSLSA23} and language-specific variants, \cite{DBLP:conf/clef/AlamBCSHHLMMZN23}, often combined with data manipulation or model ensembling strategies.  For example, the best results on the English portion of the CheckThat 2022 dataset were achieved by a RoBERTa model that leveraged a back-translation-driven data augmentation process \cite{DBLP:conf/clef/Savchev22}. 
The 2024 edition, instead, has seen an increased interest in using also generative LLMs for the task such as LLama 2 and 3, GPT3.4 and 5, Mixtral and Mistral \cite{10.1007/978-3-031-56069-9_62}. For instance, the best performing system on the English language dataset relied on fine-tuned Llama2 7b on the original training data, using prompts generated
by ChatGPT \cite{DBLP:conf/clef/LiPZ24}.   
Nevertheless, despite the notable progress in data and methods for check-worthy claim detection, there is currently a lack of an extended coverage across languages and topics.

Concerning previously fact-checked claims, the task is also formulated as a retrieval one, although the name of the task may vary across the literature; fact-checking URL recommendation~\cite{vo_lee_2018_guardians}, detection of previously fact-checked claims~\cite{shaar-etal-2020-known}, verified claim retrieval~\cite{DBLP:conf/clef/Barron-CedenoEN20}, searching for fact-checked information~\cite{vo-lee-2020-facts}, claim matching~\cite{kazemi-etal-2021-claim} or retrieval of previously fact-checked claims~\cite{pikuliak-etal-2023-multilingual} have all been previously used to denote it. 

Being a retrieval task, the existing methods apply one or a series of (re-)rankers and \textit{mean average precision (MAP)}, \textit{mean reciprocal rank (MRR)}, $P@k$ or a $Success@k$ ($Hit@k$) are used as evaluation metrics. In case of a series of rankers, the works tend to use a baseline ranker that is easy to compute and has a good recall and then one or more subsequent rerankers that work over a progressively smaller subset of results retrieved by a previous ranker in the pipeline, see, e.g.,~\cite{shaar-etal-2020-known, hardalov-etal-2022-crowdchecked}. The rerankers' task is to improve precision by moving the relevant results to the top; since they are working with a smaller set of results, they can be more computationally demanding. Alternatively, the rankers could be used in combination as an ensemble to improve the precision at the cost of higher computational demands, but this is not observed in the surveyed works. In most of the works, BM25~\cite{bm25} or similar information retrieval algorithms are used as a baseline. Various neural text embedding models are used as either sole rankers (e.g., in~\cite{pikuliak-etal-2023-multilingual}), rerankers (e.g., in~\cite{shaar-etal-2020-known}) or as ensembles~\cite{martin_2022_facter_check}, especially sentence transformers~\cite{reimers-gurevych-2019-sentence}, which use Siamese networks to pre-train text representations, usually on various sentence similarity datasets. 

The approaches also use several other techniques to improve the retrieval performance, such as text embedding models fine-tuning~\cite{pikuliak-etal-2023-multilingual, kazemi-etal-2021-claim}, distance supervision to work with noisy data~\cite{hardalov-etal-2022-crowdchecked}, key sentences extraction~\cite{sheng-etal-2021-article}, extended context of the input and verified claims (especially for political debates~\cite{shaar-etal-2022-role}), extraction of text from images~\cite{vo-lee-2020-facts, pikuliak-etal-2023-multilingual}, using multimodal representation combining text and images~\cite{vo-lee-2020-facts}, or query (input claim) modification or rewriting to be more easily matched with the fact-checked claims~\cite{bhatnagar-etal-2022-harnessing, kazemi-etal-2023-query, sundriyal-etal-2023-chaos}. The approaches may further differ by the use of loss, selection of negative examples and other (hyper-)parameters when fine-tuning the neural models. All surveyed approaches (including solutions submitted to the CheckThat! Lab challenge~\cite{DBLP:conf/clef/Barron-CedenoEN20, shaar_overview_2021, nakov_overview_2022}) rely on smaller languages models, such as BERT, XLMRoBERTa, etc. One exception is the work of~\citet{sundriyal-etal-2023-chaos}, where the authors employ LLMs to normalize the claims for the purpose of query rewriting for the retrieval task; however, they do not perform experiments on the retrieval task, thus focusing only on the first step (claim detection) in the pipeline.

\subsection{Tools and services}

Given that automatizing check-worthiness detection can greatly support and speed up  fact-checking activities, a number of systems has been already developed for the task, some of which are based on insights and databases actually used by fact-checking organisations.

\textit{ClaimBuster} \cite{10.1145/3097983.3098131} was the first end-to-end system for computer-assisted fact-checking trained on a human-labeled dataset of check-worthy factual claims from the U.S. general election debate transcripts. The first component of the pipeline is a detector of check-worthy factual claims which, given a sentence, first labels it as being `non-factual', `factual and unimportant' or `factual and check-worthy'. In case of the latter, a ranking score is assigned based on SVM decision function. \citet{10.1145/3132847.3133150} present \textit{Tathya}, a tool focusing only on check-worthiness detection, which compared to ClaimBuster can yield a significant performance improvement, particularly on recall.  It is based on a multi-classifier system using features such as topics, entity history and PoS tuples.  

\textit{ClaimRank} \cite{jaradat-etal-2018-claimrank} performs check-worthy claim detection and supports English and Arabic texts. Its strength is that it was trained on actual annotations from nine reputable fact-checking organizations, therefore mimicking their real strategy for claim selection. The ranking is based on a number of lexical, structural and semantic features, used to train a neural network with two hidden layers as proposed by \citet{gencheva-etal-2017-context}. Another system, focusing specifically on tweets in Arabic, is \textit{Tahaqqaq} \cite{10.1145/3539618.3591815}, which includes the possibility to identify check-worthy claims, estimate the user credibility in terms of spreading fake news, and find authoritative accounts.

\textit{dEFEND} \cite{10.1145/3357384.3357862} is another end-to-end system that, given a link to a post or a news, detects check-worthy sentences by assigning them a score, with the goal to distinguish between check-worthy factual claims from subjective ones. The system displays also the propagation network of the text as well as an analysis of the news comments and the textual evidence supporting the classifier decision. Another similar end-to-end platform, providing evidence snippets to credibility classification and check-worthiness decisions, is \textit{BRENDA} \cite{10.1145/3397271.3401396}, which provides also the possibility to collect users' feedback about wrong predictions. The tool is released as Google Chrome extension.

Among the few systems dealing with languages other than English, \textit{FactRank} \cite{BERENDT2021100113} was the first system able to process check-worthy texts in Dutch. The classification algorithm was developed iteratively, combining expert fact-checker input, a codebook to support reliable human labelling, and active-learning. Check-worthiness classification performance is comparable to results obtained on English with ClaimBuster.   

Concerning retrieval of previously fact-checked claims, it is supported by some of the end-to-end verification systems mentioned above, namely ClaimBuster~\cite{10.1145/3097983.3098131}, Tahaqqaq~\cite{10.1145/3539618.3591815} or BRENDA~\cite{10.1145/3397271.3401396}. Besides these, \textit{Google Fact-Check Explorer}\footnote{\url{https://toolbox.google.com/factcheck/explorer}} is often used to perform the task since it indexes a large corpus of fact-checks. Other specialized tools include \textit{Fact-Check Finder}\footnote{\url{https://fact-check-finder.kinit.sk}} built on models developed in~\cite{pikuliak-etal-2023-multilingual}.

\subsection{Discussion}

\textbf{Multilinguality}. We observe in both check-worthiness detection as well as in retrieval of previously fact-checked claims stronger shift towards multilinguality. This is, on one hand, reflected in newer datasets containing also languages other than English (either multilingual ones or focused on a specific language), on the other by the more prevalent use of multilingual models. Since the amount of data in other languages is often limited, approaches for transfer learning~\cite{kazemi-etal-2021-claim, lucas-etal-2022-detecting}, low-resource fine-tuning~\cite{cekinel-etal-2024-cross} or adapter fusion~\cite{schlicht_2023_multilingual} are explored. However, using translation to English in combination with an English language model can still sometimes outperform a multilingual approach, as was observed, e.g., in the previously fact-checked claim retrieval task~\cite{pikuliak-etal-2023-multilingual}. This is likely to change in the future with the employment and/or development of larger and better balanced multilingual models.

\textbf{Multimodality}. Although most available datasets are mostly textual, if they contain links to the original content where the claim was made, it is sometime possible to get to other modalities, such as images, videos or audio, which can be contained in a piece of content (e.g., a social media post). These can be important, because in many cases it is there where the actual claim is being made or the claim requires multiple modalities to be properly interpreted. Multimodal content can also be perceived as more credible by users, has a higher engagement and is increasingly easier to produce~\cite{akhtar-etal-2023-multimodal}. Thus, multimodal approaches continue to grow in both prevalence and importance. At the moment, most approaches transcribe the modality to text by either using OCR or image description approaches~\cite{mansour_2022_did, pikuliak-etal-2023-multilingual}, however, there are already some approaches that process the other modalities directly, be it images~\cite{vo-lee-2020-facts} or speech~\cite{ivanov_2024_detecting}. The future advances will likely lie in advancements of the latter category of approaches.

\textbf{Availability of datasets}. Although there are available resources for both check-worthiness detection and retrieval of previously fact-checked claims, both tasks have their own (sometimes overlapping) sets of challenges. In case of check-worthiness detection, it is relatively easy to collect check-worthy claims -- these are all claims verified by fact-checkers. However, collecting non-check-worthy ones is much more challenging. In case of retrieval of previously fact-checked claims, the challenge lies in collecting input claims (e.g., in the form of social media posts) and in identification of pairs between the input and the fact-checked claims. As discussed above, existing methods either lead to too strict matching with many unidentified (false negative) pairs or to too much noise in the data. Another issue for both tasks is that many datasets were previously built using Twitter. If only the IDs of the tweets have been published, it is now very expensive for researchers to use them due to the X's current API limitations and pricing, thus making their use impractical or completely unfeasible.

\textbf{Combination of check-worthy claim detection and retrieval of previously fact-checked claims}. As can be seen from the surveyed works, most of them approach the tasks in separation. This is reasonable from the scientific perspective, but more end-to-end (combined) approaches capable of first detecting a check-worthy claim and then retrieving previously fact-checked claims are needed for practitioners to use.

\textbf{Adoption of LLMs}. Although we observe some approaches using LLMs in check-worthiness detection~\cite{10.1007/978-3-031-56069-9_62}, their potential for the retrieval of fact-checked claims begins only now to be more systematically explored. Most recently, LLMs have been used and evaluated as text embedding models as well as rerankers of retrieved fact-checked claims in multilingual and crosslingual settings in~\cite{ramponi2025multilingualvscrosslingualretrieval}, outperforming the fine-tuned smaller models. They have been also employed in zero- and few-shot settings as binary classifiers filtering out irrelevant retrieved results using a range of prompting strategies~\cite{vykopal2025llms, pisarevskaya-zubiaga-2025-zero}; these recent works showed that while useful also in this setup, no single prompting strategy proved as the best overall and the performance of the current LLMs is lower for low-resource languages compared to high-resource ones. The use of LLMs (as text embedders, filters or rerankers) have been further explored in system papers submitted to the recent SemEval-2025 Task 7 as summarised in~\cite{peng2025semeval2025task7multilingual}. Finally, the benefit of LLMs for both check-worthiness detection or retrieval of previously fact-checked claims lies in input claim normalization~\cite{sundriyal-etal-2023-chaos}, using retrieval augmented generation~\cite{huang-etal-2022-concrete}, or in providing summaries of the check-worthy content or of the retrieved fact-checks~\cite{vykopal2025generativeaidrivenclaimretrievalcapable}.
\section{Additional credibility signals}
\label{section:additional-signals}

\subsection{Text quality}

\textit{Text quality} is a broad category of credibility signals measuring text's linguistic accuracy, such as readability, grammatical correctness, or spelling mistakes. It is strongly related to perceived credibility, since high-quality, more professional, content is often seen as more trustworthy. Research on statin-related websites found that more readable and accurate information significantly improves users' perceptions of credibility \cite{info:doi/10.2196/42849}. A similar study by \citet{killicaritahowstudentscred} highlighted that professionalism in text, such as proper grammar and clear structure, plays a crucial role in how credibility is judged. 

The similar relation to credibility can be observed for a low-quality content. \citet{virtualSalt} showed that poor grammar or frequent spelling errors can be a signal of lack of credibility, prompting readers to question the reliability of the information presented. \citet{10.3389/fpsyg.2022.940903} investigated how various editorial elements such as superlatives, clickbaits, boldface and poor grammar affect the quality and  credibility of online messages. Thus, the quality of the text directly enhances trust in it.

Various NLP techniques have been developed to rate text quality and, by extension, its credibility. \citet{Mosquera_Moreda_2021} evaluated text by extracting features like contractions, slang, misspellings, emoticons and readability (using the Readability Index). They also measured information content through entropy and emotional tone using emotional distance and, finally, applied the Expectation-Maximization (EM) algorithm to cluster texts by informality levels. The tool obtained a F1 score of 60.6\%. In \citet{10.5555/1613715.1613742}, the proposed model combines lexical, syntactic, and discourse features to predict human judgments of text readability. It evaluates vocabulary difficulty using unigram language models, syntactic complexity by measuring parse tree height, and discourse relations through annotated markers. Additionally, entity coherence is examined by analysing semantic continuity between sentences. When all features are used to feed a linear regression, the accuracy on readability results in 88.88\%.
The model proposed by \citet{mesgar-strube-2018-neural} uses neural networks to evaluate the quality of the text.  It captures semantic connections between adjacent sentences by representing sentences as vectors and identifying the most similar states between them. The model uses Recurrent Neural Networks to account for word context within sentences and a Convolutional Neural Network to identify patterns of coherence across the text. This allows the analysis of sentence-to-sentence transitions to predict the readability with a 97.77\% accuracy.

For the purpose of text quality classification, datasets of various content types originating in a different sources have been used so far. To name a few, \citet{Mosquera_Moreda_2021} used a dataset of 50 Yahoo! Answers posts, rated by 6 people in 4 informality level. \citet{10.5555/1613715.1613742} utilized 30 Wall Street Journal articles from the Penn Treebank, rated by college students for readability; and \citet{mesgar-strube-2018-neural} used a dataset of 105 texts human labeled from the British National Corpus and Wikipedia.

\subsection{References and citations}

When speaking of credibility, elements such as \textit{references}, \textit{citations} and partially also \textit{quotes} can influence the confidence of the user in the content and naturally represent important credibility signals \cite{10.1145/3428658.3431077,10.1145/3308560.3316460,10.1145/3184558.3188731,Baier2022}. It should be noted that the role and influence of references or citations is different to the influence of the quotes although they can occasionally have a similar role \cite{10.1145/3184558.3188731,Baier2022}. 

References and citations are traditionally analysed in the context of scientific articles, where they give support by acknowledging prior work and providing a knowledge basis, and thus enhancing their reliability and supporting the validity of their findings \cite{10.1145/3184558.3188731,Baier2022}. Commonly, explicit references make the sources more transparent, incrementing credibility perception. Quotes from outside experts further enhance the credibility of scientific articles by offering validation and expert perspectives \cite{10.1145/3428658.3431077,10.1145/3184558.3188731}. Analogously, the role of references and citations can be extrapolated to other domains and online content, such as journalism and newspapers, social media content and marketing \cite{10.1145/3184558.3188731,Baier2022}.

In \citet{Baier2022}, it is discussed how the explicitness of references and the depth of assurance provided by citations influence the perceived credibility of the information with experimental results showing that news articles with explicit citations are perceived as more credible by readers. Similarly, \citet{10.5555/3200334.3200343} highlights the importance of citations in presenting diverse perspectives. Besides overall credibility, references, citations and quotes also can be used in network analysis to detect biases \cite{10.1145/3428658.3431077}. \citet{10.1145/3308560.3316460} used references and citations in link analysis and bias detection, while quotes are discussed in regards to the effectiveness for bias detection.

The datasets used in these studies are obtained from multiple domains, including journalism, social media, and scientific content. In \citet{Baier2022}, the dataset comprises articles with varying levels of citation explicitness to experimentally assess reader perceptions of credibility. \citet{10.5555/3200334.3200343} aggregated news articles from diverse sources and examined their references and citation patterns. Similarly, works \cite{10.1145/3428658.3431077,10.1145/3308560.3316460} derived the datasets from media outlets, focusing on link analysis and citation patterns to detect bias and assess credibility. These datasets reflect a broad range of content types, enhancing the generalizability of the findings across different informational contexts.

\subsection{Clickbaits and title representativeness}

The term \textit{clickbait} refers to content designed to raise curiosity and attract users to click on links, often by using sensationalized or misleading headlines. The goal is typically to increase web traffic and advertising revenue \cite{ZHANG2020100095}. Clickbait headlines often blur the lines between fact and fiction, contributing to the spread of fake news online \cite{10.1145/3449183}.

To detect clickbait, various NLP methods can be employed, focusing on analysing lexical and semantic features of headlines. For instance, frequent use of sensationalist language, unresolved pronouns, and forward-referencing structures can indicate clickbait \cite{ZHANG2020100095}. Traditional machine learning models, such as Support Vector Machines (SVM) and Naïve Bayes classifiers, are often used to identify these features by assigning probabilities to words and phrases, which are then used to classify headlines as clickbait or non-clickbait \cite{10.1145/2823465.2823467}. In this direction, \citet{ZHANG2020100095} focused on Chinese social media and found significant regional variations in clickbait prevalence. Their studies highlight the effectiveness of SVM and Naïve Bayes classifiers, achieving an F1-measure of 0.834. Combining these approaches with other models like Long Short-Term Memory (LSTM) networks can further enhance classification accuracy, with the SVM model achieving 98.53\% accuracy \cite{WINARTO2023282}.

\textit{Title representativeness}, which measures how accurately a title reflects the content of the article, is another crucial aspect. \citet{10.1145/3449183} examined the credibility of news when provided with different combinations of title, image, and source bias. Their study found that combining these elements yielded the best accuracy for automated detectors 0.83, underscoring the importance of integrating multiple meta-data elements for improved accuracy in detecting misleading content.

The datasets used in these studies vary in their scope and origin. For example, one of them focused on Chinese social media platforms like WeChat, where they analysed regional clickbait patterns using a dataset of social media posts \cite{ZHANG2020100095}; whereas the study by \citet{WINARTO2023282} leveraged YouTube titles in their study, creating a large, labeled dataset of video headlines to train their machine learning models. Additionally, \citet{10.1145/3449183} collected news articles with varying degrees of title representativeness, image content, and source bias, analysing how these elements impacted credibility perceptions. Together, these datasets provide a diverse foundation for training and evaluating models aimed at detecting clickbait across different platforms and content types.

\subsection{Originality and content reuse}

\textit{Originality and content reuse} have a direct impact on the credibility of information. When the content is original, it reflects the author's unique insights, analysis, or research, which strengthens its authenticity and trustworthiness. However, when the content is reused, whether through plagiarism, replication without attribution, or even subtle forms of paraphrasing, it can undermine the perceived credibility of the information. Plagiarism is defined as taking intellectual property from another and passing it off as one's own without citation \cite{bin-habtoor2012survey}.

In the context of NLP, plagiarism detection has become more difficult due to the amount of texts available both in traditional print publications and now online. The challenge lies in detecting not only direct copying but also other forms of plagiarism, such as paraphrasing, translations or using hired writers to produce content.
The exploration of originality and credibility signals in text reuse involves a variety of approaches combining NLP techniques with advanced machine learning methods. One prominent strategy is leveraging sentence segmentation, tokenization, and syntactic parsing to dissect the structure of text \cite{chong2010using}. The proposed framework integrates these NLP techniques with trigram similarity measures and dependency relations matching, showing a detection accuracy of 70.53\% for plagiarized short excerpts of text. This proves how traditional linguistic tools can be enhanced by statistical models to detect subtle patterns in content reuse. Similarly, trigram similarity alongside language model probability and longest common subsequence methods to deal with plagiarism detection can be adopted \cite{10112546}. Use of a Naïve Bayes classifier resulted in the same accuracy rate of 70.53\%, underscoring the importance of pairing conventional text analysis techniques with machine learning algorithms to capture nuanced instances of content duplication. 

Other works advanced further by incorporating deeper layers of analysis through methods like Latent Semantic Analysis (LSA) and Latent Dirichlet Allocation (LDA) \cite{elngar2021plagiarism}. The proposed solution employs lexical, semantic, and syntactic analysis to detect similarities, achieving an accuracy of 89\%. By focusing on semantic relationships between words and phrases, it makes it particularly effective at discovering more sophisticated forms of content reuse.

Beyond general plagiarism detection, more specialized approaches concentrate themselves on verifying the authenticity of authorship. \citet{10.1145/3543895.3543928} developed an Authorship Verification model aimed at identifying hired plagiarism, where someone other than the credited author produces the work. By analysing stylistic features, such as writing patterns and comparing them with the claimed author's profile, their approach achieved an accuracy of 85\%. This sheds light on the possibilities of stylistic analysis as a signal of originality, distinguishing between genuine and outsourced authorship. 

The datasets used in these studies on plagiarism detection can be found from a variety of sources and text types, reflecting the different challenges posed by plagiarism. For example, \citet{bin-habtoor2012survey} focus on general plagiarism detection systems, analysing publicly available datasets from academic publications. \citet{chong2010using} utilize a dataset of academic articles to explore plagiarism through techniques like trigram similarity and dependency relations, applying them to short passages to measure detection accuracy. \citet{10.1145/3543895.3543928} use a dataset of student writing to train their Authorship Verification model, particularly aimed at detecting plagiarism in commissioned work. \citet{elngar2021plagiarism} apply their algorithm to a dataset comprising academic and non-academic texts, with an emphasis on lexical, semantic, and syntactic analysis. Lastly, \citet{10112546} use a dataset of online content, including articles and essays, to experiment with automatic plagiarism detection using advanced NLP techniques, combining feature extraction with machine learning models. These datasets represent a range of content, allowing for the robust evaluation of plagiarism detection methods across different textual formats.

\subsection{Offensive language}

The use of \textit{offensive language} (and its related phenomena, such as hate speech, abusive or toxic language; see, e.g.,~\cite{Papcunova2023} for definitions), particularly in online discourse, often correlates with the low credibility of the content, containing false information, prejudices, biases, and what in general is considered as toxic language. For example, \citet{botella2024exploring} show that there is a direct association between the credibility of media sources and the presence of hate in online comments, while \citet{bourgeade_2023_RH_SM} show how fake stories are used to spread hate against immigrants by analysing racial hoaxes on Twitter. Also some of the most widely used datasets for check-worthy claim detection already include information about the presence of hateful \cite{HADJAMEUR2021232} or harmful \cite{DBLP:conf/clef/NakovBMAMCKZLSM22} content, showing the importance of adopting a multi-faceted view on the problem of misleading information. Indeed, the use of abusive or toxic language, has also been included in the W3C list of credibility signals \cite{multipleauthorsCredibilitySignalsUnofficial}.
%Nevertheless, research on credibility signals and offensive language has traditionally treated the two phenomena as distinct. 

Offensive language detection has been extensively investigated within the NLP community for at least a decade, starting from English \cite{waseem2016hateful} and then comprising more and more languages (for example the Hate Speech Datasets repository\footnote{\url{https://hatespeechdata.com/}} that lists more than one hundred datasets in 25 languages). 
The creation of resources, which led in turn to the development of several approaches to offensive language detection, has been fostered by the shared tasks on hate or offensive language detection organised throughout the years at SemEval and other data challenges~\cite{DBLP:conf/sepln/FersiniRA18,DBLP:conf/evalita/BoscoDPST18,overviewGermeval,zampieri-etal-2019-semeval,zampieri-etal-2020-semeval,DBLP:conf/semeval/PavlopoulosSLA21}. Among the different approaches implemented for offensive language detection, transformer-based architectures based on RoBERTa \cite{DBLP:journals/corr/abs-1907-11692} and its multilingual variants have proven to be very effective for the task when fine-tuned on offensive data, with F1 scores $>$ 0.90 \cite{zampieri-etal-2020-semeval}. Nevertheless, the problem of online content moderation is far from being solved given the amount of toxic content still circulating on social media, which suggests that research on the topic still needs to address several understudied aspects of the phenomenon. Current works have identified some research directions worth studying, such as the problem of human label variation when creating datasets for offensive language detection \cite{leonardelli2021agreeing}, the role of annotators' biases \cite{sap2022annotators}, the presence of spurious correlations in existing datasets \cite{wiegand2019detection,ramponi2022features} and the lack of robustness when classifying data from different online platforms and domains \cite{salminen2020developing}. 
Concerning generative LLMs, recent experiments showed that classifying offensive language using zero- and few-shot learning LLMs yields considerably lower results in comparison to smaller models fine-tuned on the entire training set \cite{edwards-camacho-collados-2024-language}. Generative LLMs have been alternatively used to create synthetic datasets for hate speech detection, with the goal to address problems such as data decay and privacy concerns related to social media \cite{casula-etal-2024-dont}. Similarly, LLMs have also been employed to generate functional tests for a fine-grained evaluation of hate speech detection \cite{jin-etal-2024-gpt}. However, the fact that major generative LLMs have been adjusted through so-called \textit{alignment} to avoid generating hateful content may represent an obstacle for future research in this direction.

\subsection{Machine-generated text}

The task of \textit{machine-generated text} detection (also called synthetic or neural text detection or authorship attribution~\cite{uchendu-2023-attribution}) has already been researched since about 2018 when the arrival of GPT-1 and transformer architecture have enabled generation of reasonably coherent texts\cite{uchendu-2023-attribution}. However, it has been only with the arrival of generative LLMs that the task gained upon practical importance due to increased fidelity and quality of the machine-generated texts in English~\cite{uchendu_turingbench_2021}, but also in many other languages~\cite{macko-etal-2023-multitude, wang-etal-2024-m4gt, macko-etal-2025-multisocial} and due to low costs of such generation. Since then, there has been recognised potential of the machine-generated texts to be misused for influence operations~\cite{goldstein_generative_2023}, disinformation~\cite{buchanan_truth_2021, vykopal-etal-2024-disinformation, lucas-etal-2023-fighting, zugecova2024evaluationllmvulnerabilitiesmisused}, spam or unethical authorship~\cite{crothers_machine_2022}. Although there have been voices tempering down fears of massive misuse of LLMs for disinformation generation~\cite{simon2023misinformation}, actual field studies are missing and the rapid advancement of the new models makes the misuse of LLMs to generate and/or amplify disinformation increasingly easier. Nevertheless, when considering using machine-generated text detection as a credibility signal, it has to be noted that there are also many benign and legitimate uses (e.g., machine translation, use of LLMs to improve stylistics or grammar, etc.), so the context of such a use needs to be taken into account as well. 

Due to increased interest in the topic, there have been lately several new datasets released extending the task to a range of languages and domains~\cite{macko-etal-2023-multitude, wang-etal-2024-m4gt, macko-etal-2025-multisocial}, often supported by specialized data challenges and tasks, such as SemEval-2024 Task 8~\cite{wang-etal-2024-semeval-2024}. As to the detection methods and models, these range from statistical, e.g., Binoculars~\cite{hans2024spottingllmsbinocularszeroshot}, Fast-DetectGPT~\cite{bao2024fastdetectgptefficientzeroshotdetection}, etc. to fine-tuned language models, such as Longformer~\cite{beltagy2020longformerlongdocumenttransformer, li-etal-2024-mage}, RoBERTa~\cite{solaiman2019releasestrategiessocialimpacts} or MDeBERTa~\cite{macko-etal-2023-multitude}, but also increasingly including fine-tuned LLMs or their combinations~\cite{spiegel-macko-2024-kinit}.

Despite the undeniable progress in the task, there are several remaining challenges, such as robustness of detectors to the out of distribution data and adversarial attacks (such as authorship obfuscation~\cite{macko2024authorshipobfuscationmultilingualmachinegenerated}), their interpretability and explainability or detection of not purely generated texts~\cite{jawahar-etal-2020-automatic, uchendu-2023-attribution, wu2024surveyllmgeneratedtextdetection}.

\subsection{Non-NLP research on credibility signals}
\label{section:additional-signals-non-nlp}

While the goal of this survey is to give an overview of the credibility assessment and credibility signals from the perspective of Natural Language Processing (NLP), additional credibility signals that can be extracted from ancillary elements are not of lesser relevance. These non-NLP credibility signals can range from the multimedia attached or linked to the textual information \cite{10.1145/3579536,10.1145/3397271.3401221,10.5555/3200334.3200343}, to other contextual cues, such as the impact of an author \cite{10.1145/2492517.2492574,10.1145/3555637,ZHAO20161}, or even the acceptance and feedback provided by the engaged public \cite{8620917,10035643,LIN2016264}. For example, it can be expected that a fake video would unlikely be a part of credible content; likewise, a biased author with an untrustworthy historic record could lead to the same outcome.

Several taxonomies can be used to categorize non-NLP credibility signals, for example, a distinction can be made on the signals coming from different content modalities, such as complementary videos or images, or contextual entities, like author or source information \cite{10.1145/3292500.3330709,10.1145/3308560.3316460,10.1145/3428658.3431077}. A general classification of non-NLP credibility signals is summarised in Table \ref{tab:non-nl-classification} with corresponding signals' examples and references. It is noticeable that the credibility signals are from a large variety of sources and formats, and some of them have been presenting good research results recently \cite{7010810, 10.1145/3292500.3330709, 10.1145/1056808.1057012}.

\begin{table}[!tbp]
\footnotesize
\centering

\setlist{itemsep=-0.1em}
\setlist[itemize]{leftmargin=*}

\caption{Overview of complementary non-NLP signals}
\label{tab:non-nl-classification}

\begin{tabular}{|p{3cm}|p{8cm}|p{2cm}|}
\hline
\textbf{Category} & \textbf{Examples of non-NLP credibility signals} & \textbf{References} \\
\hline
Image-based signals & 
Image quality (resolution), Relevance of images to the rest of the content, Source attribution, Image authenticity (fake or generated), Presence of alternative text metadata
&
\cite{10.1145/3184558.3188731, OGRADY200658, 10.1145/3449183}
\\
\hline
Video-based signals & 
Video quality (high definition, video production), Relevance of video to the rest of the content, Source attribution, Video authenticity, Presence of transcriptions and subtitles
&
\cite{10.1145/3184558.3188731,OGRADY200658}
\\
\hline
Audio-based signals & 
Audio quality, Relevance of audio to the rest of the content, Source attribution, Audio authenticity, Presence of transcriptions
&
\cite{7010810}
\\
\hline
Interactive and embedded content signals & 
Quality and usability of interactive elements (interactive frameworks, maps, utility elements or forms), Trustworthiness of embedded elements (e.g. YouTube, newspapers, software providers)
&
\cite{10237152}
\\
\hline
Source signals & 
Domain authority (reputation of the website, domain authority scores), Publisher reputation (historical credibility of publisher/organization),  Website technology and design (quality, usability, loading speed, SEO implementation, consistency in design, branding/marketing), Traffic statistics (website analytics)
&
\cite{10.1145/3449183,10.1145/3292500.3330709}
\\
\hline
Ownership and transparency & 
Author information (availability, detail depth, verifiability of author),  Editorial policies (regulations and moderation/revision processes of medium), Funding and sponsorship disclosure
&
\cite{OGRADY200658}
\\
\hline
External references & 
External references or reviews from renowned sites, External fact-checking from trustworthy sites
&
\cite{10035643,6204953,10.1145/3555637,10.1145/3292500.3330709}
\\
\hline
User interaction & 
Comments and discussion or readers' profiles, User reviews and engagement, Social media activity (e.g. virality, influencers)
&
\cite{10035643,6204953,9752090,10.1145/3579536,LIN2016264,JI2023103210,ZHAO20161}
\\
\hline
Security of website & 
HTTPS protocol, Legal policies and service terms, Compliance with standards (GDPR or local regulations), Cybersecurity practices
&
\cite{OGRADY200658}
\\
\hline
Content presentation & 
Consistency in design and structure, Visual and audio aesthetics
&
\cite{10.1145/3449183}
\\
\hline
\end{tabular}
\end{table}

Studies have shown that combining textual analysis with multimedia analysis leads to significant improvements in credibility assessment as well as misinformation detection~\cite{10.1145/3673236,alam2022surveymultimodaldisinformationdetection,hangloo2022combating,10.1145/2492517.2492574,10.1145/3555637}. This combination is typically achieved through ensemble models or architectures that unify the outputs of various sub-models focused on different media types. A common method involves using transformer-based models such as BERT or GPT for text analysis alongside convolutional neural networks (CNNs) for image or video analysis. These models work in tandem, the text models identifying linguistic signals of credibility while the CNNs assess visual features like image authenticity, quality, and relevance. Additional layers of analysis might focus on audio quality, transcription accuracy, or even the relationship between the text and accompanying media. For example, a multimodal model might analyse a tweet’s text for sentiment and coherence, while also assessing the attached image’s metadata and authenticity. If the image shows signs of manipulation or does not align with the textual claim, the credibility for that piece of content would be lowered. Another approach is the usage of propagation graphs \cite{9752090,10237152,10.1145/3579536}, where the social activity of the sharing and propagation of a piece of news in a social network is the subject of study. Finally, some works combine both the user and community engagement together with the actual content (including using NLP techniques to process the text) to detect the trustworthiness of the piece of news in particular \cite{8620917,LIN2016264,ZHAO20161}.

At the same time, studies in multimodal credibility assessment consistently show that while non-textual cues (e.g., image authenticity, video manipulation detection) significantly enhance the detection process, they are most effective when paired with deep text analysis. For instance, a recent work combines user engagement patterns with textual signals to identify low credibility content \cite{QURESHI2023109028}. Another promising approach is the integration of social media metadata (such as virality metrics and user behaviour) alongside textual content to detect the spread of such low-credible content \cite{10.1145/3579536, 10.1145/3449183}.
\section{Challenges and open problems}
\label{section:challenges-and-open-problems}

Following the analysed research works as well as our own experience in the area of credibility assessment and credibility signal detection, we follow up with an overarching and orthogonal (to credibility assessment and categories of credibility signals described so far) discussion of challenges and open problems.  

\subsection{Research fragmentation and need for integration}

The overarching conclusion of this literature survey is the high degree of fragmentation across the field of credibility assessment research. This issue became apparent throughout the entire survey process -- beginning with the analysis of existing taxonomies/lists of credibility signals and our unified categorization (Section \ref{section:background}), continuing with the review of research on automatic credibility assessment (Section \ref{section:credibility-assessment}), and extending to approaches for detecting individual credibility signals (Sections \ref{section:factuality-subjectivity-bias}–\ref{section:additional-signals}).

This fragmentation has several important consequences. First, there is a lack of standardized taxonomies for credibility signals. The most prominent attempt so far -- an extensive list of approximately 200 crowdsourced credibility signals proposed by the W3C Credible Web Community Group~\cite{multipleauthorsCredibilitySignalsUnofficial} -- remains incomplete and not fully categorized, limiting its practical utility for guiding research and tool development.

Second, there is a significant disconnection within the NLP research community itself. Research on automatic credibility assessment is often conducted in isolation from work on the detection of individual credibility signals. As a result, even highly advanced methods for detecting certain signals (e.g., persuasion techniques) are rarely integrated into broader credibility assessment solutions. Consequently, little is known about how such signals may contribute to predicting overall content credibility in practice.

Third, researchers working on individual credibility signals frequently appear unaware of related efforts in neighbouring categories, despite these signals often being identified under the shared conceptual umbrella of credibility. Stronger interconnection between these lines of research could unlock mutual benefits -- for instance, the creation of shared, curated datasets annotated for both individual signals and overall credibility (also building on the first efforts in this direction by \cite{10.1145/3184558.3188731,10.1145/3415164}, as discussed in Section \ref{section:credibility-assessment-discussion}). This would not only increase annotation quality and sample size but also enable the study of correlations between different signals and support the development of multitask learning models -- techniques currently underutilized due to a lack of such integrated resources.

Interestingly, although research efforts remain isolated in their own silos built around the single NLP task, some end-user-oriented tools already incorporate multiple credibility signals in practice. Notable examples include the Tanbih system\footnote{\url{https://tanbih.org/}}~\cite{zhang-etal-2019-tanbih} and the Verification Plugin\footnote{\url{https://www.veraai.eu/category/verification-plugin}}.

\begin{figure}[!tbp]
    \centering
    \includegraphics[width=\linewidth]{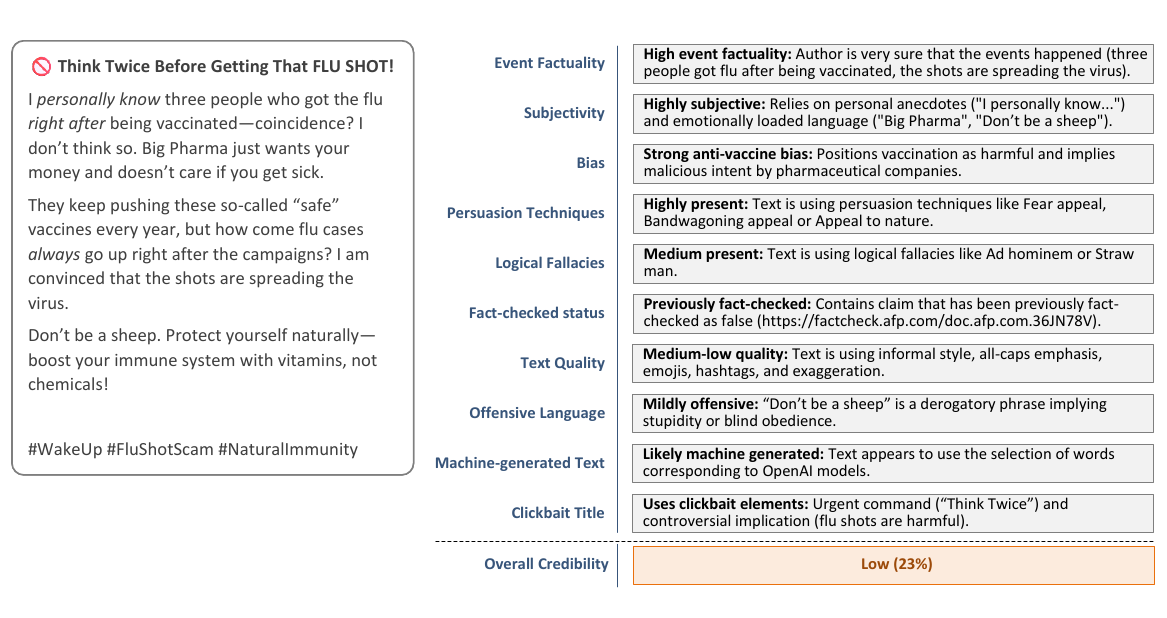}
    \caption{An illustrative example of diverse credibility signals determining the social media post as being of a low credibility. Each signal is associated with predicted values and a short explanation. The overall credibility label and score reflects the credibility predicted by individual signals.}
    \label{fig:example-low-credibility}
\end{figure}

\begin{figure}[!tbp]
    \centering
    \includegraphics[width=\linewidth]{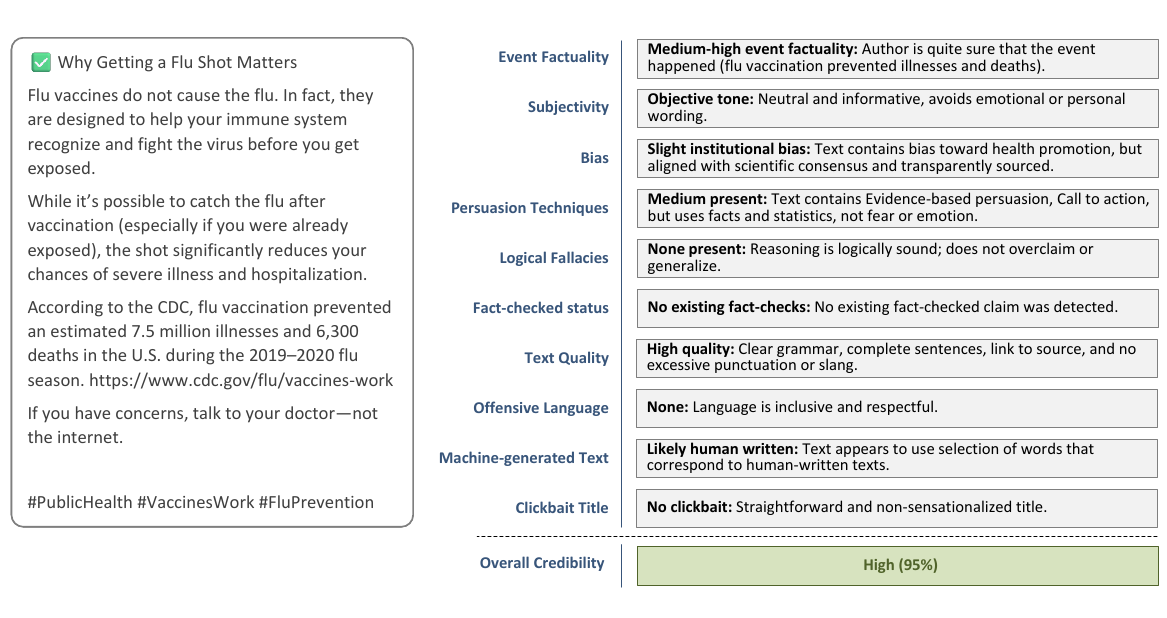}
    \caption{An illustrative example of diverse credibility signals determining the social media post as highly credible. Each signal is associated with predicted values and a short explanation. The overall credibility label and score reflects the credibility predicted by individual signals.}
    \label{fig:example-high-credibility}
\end{figure}

To illustrate the potential benefits of detecting and aggregating multiple credibility signals (representing individual categories systematically covered by this survey) into a unified credibility label/score, we provide two illustrative examples: one depicting a low-credibility social media post (Figure~\ref{fig:example-low-credibility}) and another representing a high-credibility example (Figure~\ref{fig:example-high-credibility}). These examples clearly showcase an untapped potential of synergy between detection of diverse advanced credibility signals, their explanations, as well as aggregation. To achieve such desired state, we advocate for significantly stronger integration of current research efforts.

\subsection{Adoption and potential of generative LLMs}
\label{section:challenges-and-open-problems-llms}

% context on recent instruction-tuned LLMs - zero-shot capabilities, RAG, reasoning.
Generative Large Language Models (LLMs) have demonstrated substantial improvements in complex tasks that require reasoning abilities \cite{qiao-etal-2023-reasoning}. \citet{brown-llm-few-shot-2020} showed that pretrained LLMs are capable of few-shot learning, meaning they can learn to perform new tasks with only a few training examples. Similarly, \citet{petroni-etal-2019-language} highlighted the strong ability of LLMs to recall relational knowledge acquired during pretraining to perform various tasks without further annotated labels or human supervision (i.e., zero-shot learning). Additionally, several key advancements, such as Retrieval-Augmented Generation (RAG) \cite{rag_neurips_2020}, Reinforcement Learning with Human Feedback (RLHF) \cite{rlhf}, and robust prompting techniques \cite{chain-of-thought, qiao-etal-2023-reasoning}, have further enhanced the capabilities of LLMs. 
% These advancements enable them to
In this context, recent generative LLMs
operate as dialogue systems, where the model is prompted by the user with instructions to perform specific tasks.

As we have discussed in the sections on individual credibility signals (see Sections \ref{section:factuality-subjectivity-bias}, \ref{section:persuasion-and-fallacies}, \ref{section:claims-and-veracity}, but also \ref{section:additional-signals}), the uptake of LLMs differs across the signals. In some cases, first such uses appeared only in late 2024 and beginning of 2025, such as in the case of previously fact-checked claims (Section~\ref{section:claims-and-veracity}). Nevertheless, their prevalence gradually increases. However, approaches using LLMs for credibility assessment (see Section~\ref{section:credibility-assessment}) are still rare with one such notable exception being the work of~\citet{leiteDetectingMisinformationLLMPredicted2023}. Consequently, there are still several unexplored or underexplored opportunities how such models could address challenges related to the automatic detection of credibility signals.

One of the key advantages of using prompting with LLMs is the flexibility in adapting a single foundational model to handle multiple subtasks associated with credibility assessment. With a carefully designed framework, LLMs can be guided to focus on different aspects of content analysis, such as detecting persuasion techniques, evaluating the veracity of claims, identifying potential bias, and recognizing patterns of misinformation. This flexibility reduces the need to develop and fine-tune separate models for each task, allowing practitioners to use the same model across various credibility-related tasks. In fact, a promising research direction is to explore multi-task learning, as in verifying if the capacity of performing certain credibility-assessment tasks can aid in other related ones (e.g., persuasion and bias).

Moreover, the capability of learning with zero/few examples is particularly valuable for tasks where domain-specific data is scarce or constantly changing, as is the case with credibility assessment. An enormous amount of human effort is required to curate high-quality annotated datasets for the different subtasks involved in assessing credibility. Specially since labelling most of credibility signals (such as biased content) often demand the expertise of domain specialists such as fact-checkers, journalists and social scientists. Adding to this challenge, credibility indicators can be highly context-dependent, varying across cultural and temporal dimensions. In this context, LLMs offer a more scalable approach by drawing on vast amounts of unsupervised pretraining data, and by adjusting to specific end-tasks through careful prompting strategies, which require far less human effort than manual data labelling. As an example, in \citet{leiteDetectingMisinformationLLMPredicted2023}, a generative LLM was employed to predict $19$ different credibility signals present in textual content without using any training data (i.e., in a zero-shot setting).

Finally, the generative capabilities of large language models can be leveraged to produce more explainable and interpretable\footnote{Here, the concept of interpretability differs from explainability, which is often used in the field of machine learning to refer to specific methods to analyse how intermediate states of the model lead to certain outcomes \cite{burkart2021survey}.} outputs, which is crucial for subject-matter experts that may leverage the model’s predictions. Instead of providing only binary or scalar outputs (e.g., true/false, misinformation/non-misinformation, credible/non-credible) as in usual classification tasks, generative LLMs can produce detailed explanations or summaries that can highlight the reasoning behind their predictions. This transparency allows human experts to critically assess the model’s outputs, cross-check them with external information, and ensure that (i) incorrect predictions (in this context, known as \textit{model hallucinations} \cite{llm-halluciation-survey-2023}) are properly mitigated, and (ii) any credibility assessment aligns with the context of the content being investigated. This property of interpretable outputs can significantly increase trust in model-driven decisions and reduce the likelihood of over-reliance on machine predictions, ensuring that humans remain in the loop for final judgments.

We acknowledge that LLM adoption (despite providing a lot of potential) is also accompanied with several challenges. Fine-tuning as well as deployment of LLMs require a considerably higher computing power which directly translates to higher costs. Moreover, learning techniques commonly used in limited labelled data scenarios (prompting, in-context learning, fine-tuning, meta-learning, or few-shot learning) are known to be sensitive to various randomness factors \cite{10.1145/3691339}, what can result in undesired performance instability. Beside randomness factors, also systematic choices, such as the format of the prompt or how many (in-context) samples are used, have significant effect on the overall performance and the stability of these approaches~\cite{pecher2024sensitivitylearninglimitedlabelled,voronov-etal-2024-mind}. Nevertheless, ongoing research in the area of LLMs is already providing suitable solutions to such problems, such as Parameter-efficient Fine Tuning (PEFT) techniques \cite{xu2023parameterefficientfinetuningmethodspretrained}, which have been demonstrated to perform well also on credibility signal detection tasks \cite{sheffield_comparison_peft}; or instability mitigation techniques, such as ensembling, noise regularisation and model interpolation \cite{pecher2024fightingrandomnessrandomnessmitigating}. At the same time, we would like to stress that employing LLMs does not necessarily provide benefit in all cases. When a sufficient amount of labelled samples is available (100-1000 depending on the specific task and model), fine-tuned smaller language models can outperform larger general ones \cite{pecher2024comparingspecialisedsmallgeneral}.

Finally, as highlighted in Section~\ref{section:factuality-subjectivity-bias}, implicit biases present in various LLMs can make them unreliable in when assessing external biases or other credibility signals \cite{lunardi2024elusiveness, bang2024measuring}. Efficient debiasing strategies, evaluation metrics for bias detection and alignment strategies must be implemented before such models can be applied to real-world tasks \cite{lin2024investigating, mohanty2025fine}. Factual correctness of LLMs is also vulnerable to slight contextual shifts and hallucinations \cite{augenstein2024factuality}, and improvements in source attribution, domain-specific adaptation and reasoning capabilities are cornerstone tasks in this direction that need to be addressed.

\subsection{Dataset availability and multilinguality}
\label{section:challenges-and-open-problems-datasets}

Dataset availability heavily differs across various categories of credibility signals. Firstly, we observed a significant lack of large-enough datasets providing overall (expertly-determined) annotation of credibility that can be used for training and evaluating solutions on credibility assessment task (Section~\ref{section:credibility-assessment-datasets}). Furthermore, datasets containing annotations for overall credibility and (at least some) credibility signals at the same time are even more scarce. As a result, many researchers opted to use (more available) fake news datasets as a replacement. At this place, we would like to  highlight again that fake news annotations cannot reliably replace credibility annotations (non-credible content is not necessarily only false content and vice versa, since credibility is rather a parallel and complementary dimension to veracity; see Section \ref{section:background}).

On the other hand, the situation with dataset availability for individual categories of credibility signals is much better. This is especially thanks to data challenges (particularly SemEval tasks and CLEF CheckThat! Labs as evidenced in Sections~\ref{section:factuality-subjectivity-bias}, \ref{section:persuasion-and-fallacies}, or \ref{section:claims-and-veracity}) in which either new datasets were introduced or existing datasets were extended (with more data, new languages, or additional types of annotations). Unfortunately, in some cases (e.g., persuasion techniques dataset by \citet{piskorski-etal-2023-semeval} introduced in Section~\ref{section:persuasion-and-fallacies}), the datasets from data challenges are not published completely and a hidden/testing set is not shared with researchers not even when the competition is over (and thus only the training/validation sets remain available for further research).

Especially for international news or the investigation of global claims, journalists and fact-checkers need to verify the credibility of information by cross-checking sources in different languages. Furthermore, emerging disinformation in one country could be spread to other countries, especially when its topic is global (e.g., pandemic, wars, international relations). Therefore, it is important to implement credibility analysis tools that support multiple languages. However, due to the scarcity of multilingual datasets, trained models can exhibit biases towards some languages and cultures and hence can underperform on content in low-resourced languages (as is for example the case for fact-checked claims retrieval as evidenced in~\cite{vykopal2025llms}; see Section~\ref{section:claims-and-veracity}).

In Table~\ref{tab:summary}, we present a summarised overview of language-specific datasets available for different categories of credibility signals. As in many other areas of NLP, English remains the dominant language, with the majority of available resources focused on English-language content. Nevertheless, notable datasets also exist for other widely spoken languages, particularly Arabic and Spanish. In contrast, there is a clear lack of annotated resources for many low- and mid-resource languages, such as Czech or Tamil. Among the covered categories, fact-checked claims demonstrate the broadest language coverage. This can be attributed to the global presence of fact-checking organizations, which produce multilingual artifacts that serve as the foundation for constructing such diverse and multilingual datasets. On the other hand, categories such as factuality and bias show limited availability of resources beyond English, highlighting a significant gap in the development of multilingual tools for credibility signals detection.

\begin{table}[]
\caption{Summary of language-specific resources per individual categories of credibility signals. In a case of \cite{pikuliak-etal-2023-multilingual, nielsen-mcconville-2022-mumin} datasets -- which are highly multilingual, comprising 42 and 39 languages respectively -- we directly report those languages that have an overlap with other datasets, while the number of remaining languages is provided in the \textit{Additional languages} row. For a full list, please, refer to the original papers.}
\label{tab:summary}
\resizebox{\columnwidth}{!}{
\begin{tabular}{|l|llllll|}
\hline
\multicolumn{1}{|c|}{\multirow{2}{*}{\textbf{Language}}} & \multicolumn{6}{c|}{\textbf{Credibility Signals}}                                                                                                                                                                                                                                                                                                                                                                                                  \\ \cline{2-7} 
\multicolumn{1}{|c|}{}                                   & \multicolumn{1}{c|}{\textbf{Subjectivity}}                               & \multicolumn{1}{c|}{\textbf{Factuality}} & \multicolumn{1}{c|}{\textbf{Bias}} & \multicolumn{1}{c|}{\textbf{\begin{tabular}[c]{@{}c@{}}Persuasion \\ techniques\end{tabular}}} & \multicolumn{1}{c|}{\textbf{\begin{tabular}[c]{@{}c@{}}Check-\\ worthiness\end{tabular}}} & \multicolumn{1}{c|}{\textbf{\begin{tabular}[c]{@{}c@{}}Fact-checked claims\end{tabular}}} \\ \hline
English                                                      & \multicolumn{1}{l|}{\begin{tabular}[c]{@{}l@{}}\cite{piskorski-etal-2023-semeval, SPINDE2023100264, BIYANI2014170}\\
\cite{spinde-etal-2021-neural-media, banea-etal-2010-multilingual,pang-lee-2004-sentimental}\\
\cite{wiebe2005annotating}\end{tabular}}                      & \multicolumn{1}{l|}{\cite{qian2022document, qian-etal-2019-document, li2024maven}}                    & \multicolumn{1}{l|}{\begin{tabular}[c]{@{}l@{}}\cite{piskorski-etal-2023-semeval, SPINDE2021102505, spinde-etal-2021-neural-media}\\
\cite{liu-etal-2019-detecting, fan-etal-2019-plain, chen-etal-2020-analyzing}\end{tabular}}                      & \multicolumn{1}{l|}{\begin{tabular}[c]{@{}l@{}}\cite{da-san-martino-etal-2019-findings, piskorski-etal-2023-semeval, lawson2020emailphising}\\ \cite{da-san-martino-etal-2020-semeval, dimitrov-etal-2021-semeval, macagno2022argumentationprofiles}\end{tabular}}                                                                          &    \multicolumn{1}{l|}{\begin{tabular}[c]{@{}l@{}}\cite{gencheva-etal-2017-context, cheema-etal-2022-mm, 10.1145/3543873.3587643}\\ \cite{DBLP:conf/icwsm/ArslanHLT20, DBLP:conf/clef/AtanasovaMBESZK18, DBLP:conf/clef/AtanasovaNKMM19}\\ \cite{DBLP:conf/clef/Barron-CedenoEN20, DBLP:conf/clef/NakovMEBMSAHHMH21, DBLP:conf/clef/NakovBMAMCKZLSM22}\\ 
\cite{DBLP:conf/clef/AlamBCSHHLMMZN23, 10.1007/978-3-031-56069-9_62}\end{tabular}} &  \multicolumn{1}{l|}{\begin{tabular}[c]{@{}l@{}}\cite{shaar-etal-2020-known, nakov_overview_2022}\\ \cite{vo-lee-2020-facts, kazemi-etal-2021-claim}\\ \cite{ hardalov-etal-2022-crowdchecked, singh2024-mmtweets, nielsen-mcconville-2022-mumin,pikuliak-etal-2023-multilingual}\end{tabular}}                                         \\ \hline
Chinese                                                       & \multicolumn{1}{l|}{--}                      & \multicolumn{1}{l|}{\cite{qian2022document, qian-etal-2019-document}}                    & \multicolumn{1}{l|}{--}                      & \multicolumn{1}{l|}{--}                                                                          & \multicolumn{1}{l|}{--}&      \multicolumn{1}{l|}{\cite{nielsen-mcconville-2022-mumin,pikuliak-etal-2023-multilingual}}                                         \\ \hline
German                                                       & \multicolumn{1}{l|}{\cite{atalla2011investigating, piskorski-etal-2023-semeval,banea-etal-2010-multilingual}}                      & \multicolumn{1}{l|}{--}                    & \multicolumn{1}{l|}{\cite{piskorski-etal-2023-semeval, aksenov-etal-2021-fine}}                      & \multicolumn{1}{l|}{\cite{piskorski-etal-2023-semeval}}                                                                          &  \multicolumn{1}{l|}{--} &   \multicolumn{1}{l|}{\cite{nielsen-mcconville-2022-mumin,pikuliak-etal-2023-multilingual}}                                          \\ \hline
Urdu       & \multicolumn{1}{l|}{\cite{wiebe2005creating}}                      & \multicolumn{1}{l|}{--}                    & \multicolumn{1}{l|}{--}                      & \multicolumn{1}{l|}{--}                                                                          &   \multicolumn{1}{l|}{--}  &  \multicolumn{1}{l|}{\cite{nielsen-mcconville-2022-mumin,pikuliak-etal-2023-multilingual}}                                          \\ \hline
 Arabic                                                        & \multicolumn{1}{l|}{\cite{mourad2013subjectivity, banea-etal-2010-multilingual}}                      & \multicolumn{1}{l|}{--}                    & \multicolumn{1}{l|}{--}                      & \multicolumn{1}{l|}{\cite{alam-etal-2022-overview, hasanain-etal-2023-araieval, almotairy2024arabpropaganda}}                                                                          &  
 \multicolumn{1}{l|}{\begin{tabular}[c]{@{}l@{}}\cite{HADJAMEUR2021232, DBLP:conf/clef/AtanasovaMBESZK18, DBLP:conf/clef/Barron-CedenoEN20}\\ \cite{DBLP:conf/clef/NakovMEBMSAHHMH21, DBLP:conf/clef/NakovBMAMCKZLSM22, DBLP:conf/clef/AlamBCSHHLMMZN23}\\
 \cite{10.1007/978-3-031-56069-9_62}\end{tabular}} & \multicolumn{1}{l|}{\cite{nakov_overview_2022,nielsen-mcconville-2022-mumin,pikuliak-etal-2023-multilingual}} \\ \hline
French                                                         & \multicolumn{1}{l|}{\cite{piskorski-etal-2023-semeval, banea-etal-2010-multilingual}}                      & \multicolumn{1}{l|}{--}                    & \multicolumn{1}{l|}{\cite{piskorski-etal-2023-semeval}}                      & \multicolumn{1}{l|}{\cite{piskorski-etal-2023-semeval}}                                                                          &    \multicolumn{1}{l|}{--} &  \multicolumn{1}{l|}{\cite{nielsen-mcconville-2022-mumin,pikuliak-etal-2023-multilingual}}                                           \\ \hline

Polish                                                         & \multicolumn{1}{l|}{\cite{piskorski-etal-2023-semeval}}                      & \multicolumn{1}{l|}{--}                    & \multicolumn{1}{l|}{\cite{piskorski-etal-2023-semeval}}                      & \multicolumn{1}{l|}{\cite{piskorski-etal-2023-semeval}}                                                                          &     \multicolumn{1}{l|}{--} &  \multicolumn{1}{l|}{\cite{nielsen-mcconville-2022-mumin,pikuliak-etal-2023-multilingual}}                                          \\ \hline

Italian                                                         & \multicolumn{1}{l|}{\cite{piskorski-etal-2023-semeval}}                      & \multicolumn{1}{l|}{--}                    & \multicolumn{1}{l|}{\cite{piskorski-etal-2023-semeval}}                      & \multicolumn{1}{l|}{\cite{piskorski-etal-2023-semeval, macagno2022argumentationprofiles}}                                                                          &                \multicolumn{1}{l|}{--} & \multicolumn{1}{l|}{\cite{nielsen-mcconville-2022-mumin,pikuliak-etal-2023-multilingual}}                                \\ \hline

Russian                                                         & \multicolumn{1}{l|}{\cite{piskorski-etal-2023-semeval}}                      & \multicolumn{1}{l|}{--}                    & \multicolumn{1}{l|}{\cite{piskorski-etal-2023-semeval}}                      & \multicolumn{1}{l|}{\cite{piskorski-etal-2023-semeval}}                                                                          &       \multicolumn{1}{l|}{--} &  \multicolumn{1}{l|}{\cite{nielsen-mcconville-2022-mumin,pikuliak-etal-2023-multilingual}}                                        \\ \hline

Spanish                                                         & \multicolumn{1}{l|}{\cite{banea-etal-2010-multilingual}}                      & \multicolumn{1}{l|}{--}                    & \multicolumn{1}{l|}{--}                      & \multicolumn{1}{l|}{--}                                                                          &     \multicolumn{1}{l|}{\begin{tabular}[c]{@{}l@{}}\cite{DBLP:conf/clef/NakovMEBMSAHHMH21, DBLP:conf/clef/NakovBMAMCKZLSM22, DBLP:conf/clef/AlamBCSHHLMMZN23}\\ \cite{10.1007/978-3-031-56069-9_62}\end{tabular}} &   \multicolumn{1}{l|}{\cite{martin_2022_facter_check, singh2024-mmtweets, nielsen-mcconville-2022-mumin,pikuliak-etal-2023-multilingual}}                                        \\ \hline

Romanian                                                         & \multicolumn{1}{l|}{\cite{banea-etal-2010-multilingual}}                      & \multicolumn{1}{l|}{--}                    & \multicolumn{1}{l|}{--}                      & \multicolumn{1}{l|}{--}                                                                          &    \multicolumn{1}{l|}{--} &  \multicolumn{1}{l|}{\cite{pikuliak-etal-2023-multilingual}}                                         \\ \hline

Portuguese                                                         & \multicolumn{1}{l|}{\cite{jeronimo2020computing}}                      & \multicolumn{1}{l|}{--}                    & \multicolumn{1}{l|}{--}                      & \multicolumn{1}{l|}{\cite{macagno2022argumentationprofiles}}                                                                          &               \multicolumn{1}{l|}{--} &  \multicolumn{1}{l|}{\cite{singh2024-mmtweets,nielsen-mcconville-2022-mumin,pikuliak-etal-2023-multilingual}}                              \\ \hline
Dutch                                                         & \multicolumn{1}{l|}{\cite{maks-vossen-2012-building}}                      & \multicolumn{1}{l|}{--}                    & \multicolumn{1}{l|}{--}                      & \multicolumn{1}{l|}{--}                                                                          &   \multicolumn{1}{l|}{\cite{DBLP:conf/clef/NakovBMAMCKZLSM22, 10.1007/978-3-031-56069-9_62}} &  \multicolumn{1}{l|}{\cite{nielsen-mcconville-2022-mumin,pikuliak-etal-2023-multilingual}}                                           \\ \hline

Czech                                                         & \multicolumn{1}{l|}{--}                      & \multicolumn{1}{l|}{--}                    & \multicolumn{1}{l|}{--}                      & \multicolumn{1}{l|}{\cite{baisa-etal-2019-benchmark}}                                                                          &        \multicolumn{1}{l|}{--} &    \multicolumn{1}{l|}{\cite{nielsen-mcconville-2022-mumin,pikuliak-etal-2023-multilingual}}                                    \\ \hline

Turkish                                                         & \multicolumn{1}{l|}{--}                      & \multicolumn{1}{l|}{--}                    & \multicolumn{1}{l|}{--}                      & \multicolumn{1}{l|}{--}                                                                          &        \multicolumn{1}{l|}{\cite{kartal-kutlu-2020-trclaim, DBLP:conf/clef/NakovMEBMSAHHMH21, DBLP:conf/clef/NakovBMAMCKZLSM22}} &  \multicolumn{1}{l|}{\cite{nielsen-mcconville-2022-mumin,pikuliak-etal-2023-multilingual}}                                      \\ \hline

Hindi                                                         & \multicolumn{1}{l|}{--}                      & \multicolumn{1}{l|}{--}                    & \multicolumn{1}{l|}{--}                      & \multicolumn{1}{l|}{--}                                                                          &        \multicolumn{1}{l|}{\cite{dhar-das-2021-leveraging}} &   \multicolumn{1}{l|}{\cite{kazemi-etal-2021-claim, singh2024-mmtweets, nielsen-mcconville-2022-mumin, pikuliak-etal-2023-multilingual}}                                     \\ \hline

Bengali                                                         & \multicolumn{1}{l|}{--}                      & \multicolumn{1}{l|}{--}                    & \multicolumn{1}{l|}{--}                      & \multicolumn{1}{l|}{--}                                                                          &        \multicolumn{1}{l|}{\cite{dhar-das-2021-leveraging}} &    \multicolumn{1}{l|}{\cite{kazemi-etal-2021-claim,nielsen-mcconville-2022-mumin,pikuliak-etal-2023-multilingual}}                                    \\ \hline

Norwegian                                                         & \multicolumn{1}{l|}{--}                      & \multicolumn{1}{l|}{--}                    & \multicolumn{1}{l|}{--}                      & \multicolumn{1}{l|}{--}                                                                          &        \multicolumn{1}{l|}{\cite{sheikhi-etal-2023-automated}} &  \multicolumn{1}{l|}{\cite{nielsen-mcconville-2022-mumin,pikuliak-etal-2023-multilingual}}                                      \\ \hline

Bulgarian                                                         & \multicolumn{1}{l|}{--}                      & \multicolumn{1}{l|}{--}                    & \multicolumn{1}{l|}{--}                      & \multicolumn{1}{l|}{--}                                                                          &        \multicolumn{1}{l|}{\cite{DBLP:conf/clef/NakovMEBMSAHHMH21, DBLP:conf/clef/NakovBMAMCKZLSM22}} &  \multicolumn{1}{l|}{\cite{pikuliak-etal-2023-multilingual}}                                      \\ \hline

Tamil                                                         & \multicolumn{1}{l|}{--}                      & \multicolumn{1}{l|}{--}                    & \multicolumn{1}{l|}{--}                      & \multicolumn{1}{l|}{--}                                                                          &        \multicolumn{1}{l|}{--} &  \multicolumn{1}{l|}{\cite{kazemi-etal-2021-claim,nielsen-mcconville-2022-mumin,pikuliak-etal-2023-multilingual}}                                      \\ \hline

Malayalam                                                         & \multicolumn{1}{l|}{--}                      & \multicolumn{1}{l|}{--}                    & \multicolumn{1}{l|}{--}                      & \multicolumn{1}{l|}{--}                                                                          &        \multicolumn{1}{l|}{--} &  \multicolumn{1}{l|}{\cite{kazemi-etal-2021-claim,nielsen-mcconville-2022-mumin,pikuliak-etal-2023-multilingual}}                                      \\ \hline
\hline

Additional languages & \multicolumn{1}{l|}{--}                      & \multicolumn{1}{l|}{--}                    & \multicolumn{1}{l|}{--}                      & \multicolumn{1}{l|}{--}                                                                          &        \multicolumn{1}{l|}{--} &   \multicolumn{1}{l|}{ + 22 \cite{nielsen-mcconville-2022-mumin}, + 18 \cite{pikuliak-etal-2023-multilingual}}    \\ 

\hline

\end{tabular}}
\end{table}

To overcome the scarcity of multilingual datasets, global collaborations could be initiated for creating multilingual resources. In this direction, we can already observe a positive trend within data challenges. Many of them introduced multilingual datasets, commonly considering some languages as surprise ones (i.e., languages that are present in a test set, but missing in a train/validation set). Such approach motivates participants to develop multilingual solutions that are capable to make a prediction in a zero-shot setting (considering a language a predicted content is written in).

Besides datasets availability and multilinguality, a quality of annotations remain another challenge. Annotation of overall credibility as well as individual credibility signals is many times highly subjective (as also shown in \cite{10.1145/3415164}), especially in cases such as persuasion techniques where presence of a signal and borders between various signals can be blurred. The situation is getting even more challenging in multilingual settings, where typically native speakers are needed to annotate data. Firstly, acquiring human experts fluent in several languages is challenging itself. Secondly, organizing and consolidating annotation process (including post-annotation verification) is a complex task. Considering also our own experience working with such datasets, the provided labels cannot be easily verified and many times we identified (inevitable) incorrect labels.

Last but not least, credibility assessment and automatic detection of credibility signals naturally happen in very dynamically evolving (online) environment. New topics and global events constantly emerge, causing significant data and concept drifts. In some cases, such drifts can cause that the existing datasets may become obsolete and non-representative. Secondly, the list of credibility signals itself evolves. We can take a machine-generated text as an illustrative example. This kind of signal become highly relevant only recently with generative LLMs becoming easily available for a large end user base (and unfortunately also bad actors). Such new/redefined signals thus naturally result into a demand for new datasets. Finally, the dynamics of this area also lies in its adversarial character. Bad actors (e.g., the ones who are spreading propaganda or disinformation) will always try to get their content undetected as low-credible one by employing various obfuscation techniques. This must be taken into consideration when introducing new datasets. By continuing with an illustrative example of machine-generated text, there are already datasets (e.g., \cite{macko2024authorshipobfuscationmultilingualmachinegenerated}) providing besides machine-generated texts also their alternative versions after applying several authorship obfuscation techniques, which allows to train and evaluate more robust detection models.

\subsection{Ethical and legal issues}

Credibility assessment of online (primarily social media) content is from its nature an area that must address several ethical and legal issues. 

First, such ethical and legal issues are especially prominent when the researched outcomes get deployed and used in practice by end users. The challenge for tool makers is to be fully transparent about the limitations of their tools, to provide guidelines to avoid misleading their users (e.g., by false positives and hallucinations) and to support human control (in line with the human-in-the-loop approach), for more details see Section \ref{section:challenges-and-open-problems-deployment}.

Second, similarly to other related research areas (e.g., false information detection), credibility assessment methods may be potentially \textit{misused by bad actors} in order to create content that appears to be more credible. Also by following open-research spirit, publishing credibility assessment systems can theoretically result into misusing such system in the adversarial manner to tune disinformation/propaganda generation systems and allowing them to stay undetected. This kind of potential threat is, however, an analogical issue to the security domain and the principle of security by obscurity. Nothing prevents bad actors to develop their own credibility assessments systems and apply then in adversarial training scenario. Moreover, positive outcomes of credibility assessment research are more tangible, with many practical tools (as also showed across this survey) already put into the hands of media professionals or general public.

Besides potential misuse, additional ethical considerations must be addressed thoroughly during the research activities. First of all, training various classification/detection systems is inherently a subject of potential \textit{biases}. Such biases can come directly from the datasets used for the training purposes (in terms of data selection, data pre-/post-processing, or data annotation itself; see Section \ref{section:challenges-and-open-problems-datasets}) or can be introduced during training the models, especially when fine-tuning pre-trained LLMs that have incorporated biases by themselves (including biases between high- and low-resource languages; see Section \ref{section:challenges-and-open-problems-llms}). Secondly, the authors should always clearly formulate \textit{intended use and failure cases} of the trained models/deployed tools (e.g., by means of model cards). In this way, we may prevent media professionals to over-rely on the predicted (potentially incorrect) values. Fortunately, the two-step approach to credibility assessment (i.e., detect more granular credibility signals and then aggregate them) makes the whole process more transparent. Last but not least, explainability and interpretability of the models' predictions plays a crucial role in this area, since credibility assessment must be credible itself, otherwise it would not provide expected level of trust to its end users.

Besides ethical issues, the research in this area must address also several legal issues. Many of them are shared with other research works on social media.
Firstly, as we have also witnessed recently, social media platforms can change their \textit{data processing policies} and restrict access to their data, which may delay progress in research and development of credibility tools. Additionally, social media data must be anonymized to protect \textit{user privacy} before being used as training data or for the model inference. Media organizations also impose limitations due to \textit{copyright laws}, with some not permitting their content to be used in AI tool development. LLMs, especially closed LLMs such as ChatGPT and GPT-4, lack transparency regarding pre-training data and LLMs can memorize content in their pretraining data~\cite{karamolegkou2023copyright,mueller2024llms}. Therefore, anonymization and removal of copyrighted content is crucial to credibility tools, even when they serve as foundational or backbone models. Paradigms such as unlearning~\cite{chen-yang-2023-unlearn} or LLM editing could be potential research directions to tackle these issues.

\subsection{Practical deployment}
\label{section:challenges-and-open-problems-deployment}

Deploying credibility assessment and credibility signal detection models in real-world settings presents significant challenges. For media professionals, editorial guidelines in most newsrooms and fact-checking organizations recommend treating AI-based credibility assessment or fact-checking tools as exploratory aids rather than definitive sources of evidence. Fact-checkers are generally advised to use AI-generated outputs as complementary cues -- only after accumulating sufficient independent evidence pointing to the falsity of the content in question. A similar critical challenge concerns the general public, who often lacks the expertise to critically assess AI-detected credibility signals \cite{lu2022effects}. This poses serious risks -- if such indicators are incorrect or poorly communicated, users may be misled, potentially reinforcing belief in or further disseminating false information. These practical concerns stem from multiple underlying (not only purely technological) challenges. 

First, \textit{explainability, interpretability, and transparent communication} of model confidence and limitations are essential components of any credibility assessment system intended for real-world use for both types of end users -- media professionals as well as the general public. Many classifiers (e.g., on subjectivity or persuasion techniques) are providing clues that the editor can proofread, maintaining editorial control on the analysis. At the same time, the epistemic shift of the AI-generated content is that humans are now struggling to determine if the content is synthetic or not, and therefore tend to rely more on automated tools, on which exercising editorial control is much more complicated. How can an editor trust an AI-based detector that has an unexplained and non-negligible known rate of false positives? How can she take the reputation risk of writing that the content is synthetic, especially in the context of political life, elections, and public debate, if it proves to be authentic later on? To this end, end users must be able to understand the strengths and weaknesses of models, and model providers must disclose the constraints and limits of the tools they provide (including information on the provenance of the datasets and the training of the models).

The previous research showed that appropriate visual explanations foster end users' trust in AI-predicted classification labels \cite{10.1145/3377325.3377480}. In this direction, \citet{PRZYBYLA2021102653} proposed a set of interactive visualizations designed to explain the rationale behind automatic credibility assessments, with the aim of increasing users’ confidence in the underlying methods. In a user study involving 14 participants, the authors found that the stylometric classifier was perceived as more interpretable than the neural classifier, although the latter achieved higher predictive performance. Participants were also significantly more accurate and more confident in their credibility judgments after interacting with the visual explanations. These findings underscore that providing meaningful explanations and interpretability -- beyond a simple black-box credibility label or score -- remains an open challenge. Moreover, the study highlights that addressing this issue is non-trivial, particularly given that state-of-the-art models such as deep neural networks often suffer from limited inherent explainability, despite their high performance.

Another critical challenge lies in the communication gap between technical practitioners and end users, particularly media professionals and the general public. These groups often operate using distinct terminologies and conceptual frameworks, which can hinder the interpretation of AI-generated outputs or the understanding of AI system limitations. Furthermore, many existing datasets and benchmarks have been developed primarily by computational linguists or computer scientists, without active involvement of end users. This can lead to divergent perspectives regarding what constitutes credible content or when a credibility signal is present. As our survey shows, certain credibility signal detection tasks, such as bias detection (Section \ref{section:factuality-subjectivity-bias}) and check-worthiness assessment (Section \ref{section:claims-and-veracity}), are inherently subjective, which may further exacerbate these discrepancies. Consequently, model predictions may align poorly with end users' expectations or only succeed on simple unambiguous examples.

Second, detection models making use of credibility signals should \textit{minimize false positive rates} as much as possible for integration in media organisations and journalism toolkits. However, based on both literature and our own practical experience, we observe that model performance often deteriorates when transitioning from offline evaluations to real-world applications. This discrepancy largely stems from the out-of-distribution (OOD) nature of deployment data, which may differ from training and testing datasets in various aspects, such as topic, format or language. These shifts highlight the need for more robust and generalizable models. However, achieving such robustness is complicated by the already-discussed limited availability of diverse datasets (Section \ref{section:challenges-and-open-problems-datasets}) and the dynamic nature of online environments. 

Furthermore, deployed methods and tools for automatic credibility assessment are inherently exposed to potential \textit{adversarial threats}. Malicious actors who intentionally disseminate non-credible information may attempt to evade detection by slightly altering the content -- creating adversarial examples that exploit vulnerabilities in detection models and trigger incorrect classifications \cite{Przybyła_Shvets_Saggion_2024}. To be effective in real-world applications, automatic classifiers must therefore be robust to such adversarial manipulations. Recent studies have demonstrated that LLMs, when combined with diversity incentives techniques, can be effectively leveraged for data augmentation \cite{cegin-etal-2024-effects}. By generating diverse paraphrases and content variations, such approaches not only improve model generalization but also increase resilience against potential adversarial attacks -- including those generated using similar LLM-based strategies.

From a technical deployment perspective, \textit{resource constraints} present additional barriers. While from the research perspective, open-source models are desirable for their transparency and reproducibility, hosting and running multiple models can be financially burdensome for media organizations or academic institutions. Researchers and practitioners often face a trade-off between model size and deployment feasibility: larger language models generally yield higher classification performance but require costly GPU infrastructure to support real-time inference. This necessitates compromises between computational cost and acceptable performance degradation. Techniques such as model distillation, which transfer knowledge from large models into smaller more efficient versions, may alleviate some of these challenges, but often at the cost of reduced performance.

\subsection{Multimodal approaches}

While text is still one of the most common methods to spread information, in practice, it is commonly combined with other media types, like images, videos and audios. The demand for multimodal approaches also grows with a continuous shift of social media platforms towards short multimedia formats, such as short videos, reels or slideshows. While Natural Language Processing (NLP) allows to detect linguistic patterns that may signal low credibility content (as shown throughout this survey), it can overlook cues that are often found in such non-textual elements. For example, low-credibility content can back-up its misleading text with an altered or fully generated image or video to enhance its perceived credibility. 

As previously discussed in Section \ref{section:additional-signals-non-nlp}, there are promising results on multimodal approaches that involve processing multiple data streams simultaneously, offering a richer understanding of the content and its credibility. The main potential of these methods lies in their ability to overcome limitations inherent in text-only analysis. To this end, they combine traditional NLP methods with analysis of visual, auditive, contextual or interactive content. Even in such multimodal systems, textual analysis still remains an important component as text provides rich semantic information, often containing the explicit claims or arguments being made. Without appropriate text analysis, it would be difficult to discern the specific intent behind multimedia content.

In the future, there is potential to enhance multimodal models by integrating additional modalities, such as biometric signals (e.g., eye-tracking data \cite{10.1145/3397271.3401221}), interactive content quality, or real-time user engagement metrics. Research on multimodal credibility assessment is therefore still ongoing and provides several promising directions for future research.
\section{Conclusions}

With the rapid development of generative AI, the potential misuse of LLMs for hybrid operations has become one of the most frequently cited risks \cite{goldstein_generative_2023,buchanan_truth_2021}. Recent research has highlighted the particularly high capacity of LLMs to generate (personalized) disinformation across both global and local narratives \cite{vykopal-etal-2024-disinformation,zugecova2024evaluationllmvulnerabilitiesmisused}. In this context, the large-scale capability to automatically assess the credibility of online social media content becomes even more crucial as it was in the pre-LLM era.

Potential negative impact on society and democratic values was recognized by the research community and previously led to emergence of many false information detection approaches. Unfortunately, automatic credibility assessment and the detection of credibility signals have not yet received the same level of attention. At this point, we emphasize that credibility assessment, while closely related to false information detection, provides complementary insights. In particular, false information that appears credible can have far more detrimental effects than content that is easily recognized as non-credible.

Thanks to inherent two-step nature of credibility assessment (to detect credibility signals and then to aggregate them into a credibility label/score), credibility assessment encompasses a high level of explainability and wide opportunities for application in research as well as in practice. Detected high-credible content can be promoted by recommender systems or search engines, or highlighted in a user interface of social media platforms, while low-credible content can be accompanied with a low-credibility warning. With a breakdown of predicted credibility score to individual credibility indicators, end users (either media professionals or the general public) are able to explore the evidence leading to the credibility assessment or manually evaluate individual detected credibility signals and make a final assessment by themselves. Such level of explainability and pro-active human involvement in the decision process is vital and unfortunately lacks in many current false information detection approaches that commonly result into a single predicted value (commonly a binary one) with challenging or even impossible explanation caused by a black-box nature of the employed techniques (e.g., deep learning approaches).

Despite these considerable advantages of credibility assessment, the current state-of-the-art research works suffer from multiple drawbacks. Most crucially, \textbf{credibility assessment field can be characterized as highly fragmented}. On one side, there are credibility assessment approaches that automatically detect credibility signals and aggregate them to make a final prediction about the content credibility. The utilized credibility signals are, however, mostly shallow linguistic ones (such as a number of hashtags), their automatic detection relies mostly on outdated and simple methods (like rule- or heuristic-based techniques), and also prediction utilizes mostly basic weighting schemata. There are only very few approaches that are in line with the current state of the art (deep learning, LLMs), such as \cite{leiteDetectingMisinformationLLMPredicted2023}; or building upon the previous research results to detect more complex credibility signals, such as \cite{QURESHI2023109028}. 

On the other side, there are automatic approaches detecting various categories of credibility signals, like event factuality, biases, persuasion techniques, or previously fact-checked claims. The prevalence of state-of-the-art techniques (including LLMs and various fine-tuning approaches, including PEFTs) is much higher in these works. However, \textbf{such approaches remain isolated from credibility assessment}, many times even not mentioning that their prediction can be considered as one of more advanced and more reliable credibility signals.

To address this undesired gap between research works, lack of interconnection of research results, as well as hindering application of the outcomes in practice, we conducted this systematic survey on automatic credibility assessment and detection of credibility signals from the NLP perspective. By collecting and describing {\includedTotal} research papers, we not only systematically summarised the current state of the research in (currently fragmented) research areas, but also identified shared challenges and potential for future research. Our thorough analyses and discussions aim to point to an interesting avenues for future research -- particularly, we would like to highlight three most promising research streams:

\begin{itemize}
    \item \textbf{Adoption of advanced multilingual and multimodal LLMs}. The emerging multilingual and multimodal LLMs offer substantial -- yet still largely untapped -- potential for advancing credibility assessment. Their capabilities can significantly improve both detection performance and language coverage. Moreover, these models open new avenues for deploying credibility assessment tools to support both media professionals and the general public. Such tools can serve both purposes -- to detect non-credible content as well as to identify credible one that is worth further reading and sharing in social media environment.
    \item \textbf{Development of multilingual and multi-category benchmark datasets}. The current lack of standardized, multilingual benchmark datasets hampers direct comparison between methods and slows progress in the field. We advocate for future work to focus on unifying existing -- but often fragmented -- efforts into a shared benchmark that would include diverse credibility signals annotated over the same content. Such a resource would enable novel research opportunities -- such as multi-task learning or novel credibility assessment algorithms -- and at the same time, would potentially reduce current redundant efforts in dataset creation. Especially existing highly multilingual datasets used for previously fact-checked claim retrieval could serve as a foundation for building such a benchmark.
    \item \textbf{Addressing practical deployment challenges}. Finally, bridging the gap between academic research and real-world deployment requires greater attention to practical considerations -- ranging from ethical and legal implications to the robustness of models against out-of-distribution and even adversarial inputs. Additionally, issues of computational efficiency and explainability must be addressed to ensure credibility assessment systems are both effective, trustful and reliable. Despite these challenges, existing tools already demonstrate the positive real-world impact such systems can have, both in supporting media professionals and enhancing the information ecosystem for everyday users.
\end{itemize}

\begin{acks}
This work was partially supported by the European Union under the Horizon Europe projects: \textit{vera.ai} (GA No. \href{https://doi.org/10.3030/101070093}{101070093}), \textit{AI-CODE} (GA No. \href{https://cordis.europa.eu/project/id/101135437}{101135437}), and by \textit{AI4TRUST} (GA No. \href{https://cordis.europa.eu/project/id/101070190}{101070190});
by the UK’s innovation agency (Innovate UK) grant 10039055; by EU NextGenerationEU through the Recovery and Resilience Plan for Slovakia under the project No. 09I03-03-V03-00020.
\end{acks}

\bibliographystyle{ACM-Reference-Format}
\bibliography{bibliography}

%\appendix

\end{document}